\documentclass[10pt,twocolumn,letterpaper]{article}

\usepackage{iccv}
\usepackage{times}
\usepackage{epsfig}
\usepackage{graphicx}
\usepackage{amsmath}
\usepackage{amssymb}

\usepackage{color}
\usepackage{booktabs}
\usepackage{multirow}
\usepackage{subfigure}
\usepackage{colortbl}
\usepackage{tabularx}

\newcommand{\figref}[1]{Fig. \ref{#1}}
\newcommand{\tabref}[1]{Table \ref{#1}}

\definecolor{hscolor}{rgb}{1,0,0}

\usepackage[pagebackref=true,breaklinks=true,letterpaper=true,colorlinks,bookmarks=false]{hyperref}

\iccvfinalcopy % *** Uncomment this line for the final submission

\def\iccvPaperID{7776} % *** Enter the ICCV Paper ID here

\ificcvfinal\fi

\begin{document}

%%%%%%%%% TITLE
\renewcommand\footnotemark{}
\title{Adaptive confidence thresholding for monocular depth estimation\thanks{This work was supported by the Institute of Information $\&$ Communications Technology Planning $\&$ Evaluation (IITP) grant funded by the Korean government (MSIT) (No. 2020-0-00056) and the Mid-Career Researcher Program through the NRF of Korea (NRF-2021R1A2C2011624). S. Kim$^4$ was supported in part by the MSIT under the ICT Creative Consilience Program (IITP-2021-2020-0-01819).
}}

\author{
	%Authors
	% All authors must be in the same font size and format.
	Hyesong Choi\textsuperscript{\rm 1}$^{*}$\thanks{$^{*}$ Equal contribution. $^{\dagger}$ Corresponding author: dbmin@ewha.ac.kr}, Hunsang Lee\textsuperscript{\rm 2}$^{*}$, Sunkyung Kim\textsuperscript{\rm 1}, Sunok Kim\textsuperscript{\rm 3},\\Seungryong Kim\textsuperscript{\rm 4}, Kwanghoon Sohn\textsuperscript{\rm 2}, Dongbo Min\textsuperscript{\rm 1}$^{\dagger}$\\
	\textsuperscript{\rm 1}Ewha W. University,
	\textsuperscript{\rm 2}Yonsei University,
	\textsuperscript{\rm 3}Korea Aerospace University,
	\textsuperscript{\rm 4}Korea University\\
	%\tt\small $\{$hyesongchoi2010,skk0951324$\}$@gmail.com,
	%\tt\small $\{$hslee91,khsohn$\}$@yonsei.ac.kr,\\
	%\tt\small sunok.kim@kau.ac.kr,
	%\tt\small seungryong$\_$kim@korea.ac.kr,
	%\tt\small dbmin@ewha.ac.kr
	\\
}

\maketitle
\ificcvfinal\thispagestyle{empty}\fi

%%%%%%%%% ABSTRACT
\begin{abstract}
   %Self-supervised monocular depth estimation relies on the reconstruction loss to leverage the self-supervision from a pair of stereo images or monocular video sequences.
   Self-supervised monocular depth estimation has become an appealing solution to the lack of ground truth labels, but its reconstruction loss often produces over-smoothed results across object boundaries and is incapable of handling occlusion explicitly.
   In this paper, we propose a new approach to leverage \emph{pseudo} ground truth depth maps of stereo images generated from self-supervised stereo matching methods.
   The confidence map of the pseudo ground truth depth map is estimated to mitigate performance degeneration by inaccurate pseudo depth maps.
   To cope with the prediction error of the confidence map itself, we also leverage the threshold network that learns the threshold dynamically conditioned on the pseudo depth maps.
   %   The confidence map is thresholded via a differentiable soft-thresholding operator using this truncation boundary $\tau$.
   The pseudo depth labels filtered out by the \emph{thresholded} confidence map are used to supervise the monocular depth network.
   Furthermore, we propose the probabilistic framework that refines the monocular depth map with the help of its uncertainty map through the pixel-adaptive convolution (PAC) layer.
   Experimental results demonstrate superior performance to state-of-the-art monocular depth estimation methods.
   Lastly, we exhibit that the proposed threshold learning can also be used to improve the performance of existing confidence estimation approaches.
\end{abstract}

%%%%%%%%% BODY TEXT
\section{Introduction}

Monocular depth estimation, which predicts a dense depth map from a single image, plays an important role in various fields such as scene understanding and autonomous driving. %to recognize the environment and predict the state of the target.
%This task is heavily underconstrained since a single image may be produced from an infinite number of distinct 3D scenes.
%%%Like other computer vision tasks, the monocular depth estimation has been evolved through deep learning.
Early works \cite{eigen2014depth, li2015depth, cao2017estimating} are based on supervised learning in which the performance depends on a huge amount of training data with ground truth depth labels.
%To capture accurate depth information for ground truth, auxiliary depth sensors are essential.
%However, a prevalently used 3D laser scanner to obtain depth cue such as Lidar, has the disadvantage of having lower resolution and lower sparsity (e.g. $\leq{6\%}$ for KITTI dataset\cite{geiger2013vision}).

Since establishing such a large-scale training data is very costly and labour-intensive, recent approaches rely on the self-supervised learning regime \cite{garg2016unsupervised, godard2017unsupervised, luo2018single, godard2019digging, poggi2020uncertainty}.
Instead of using ground truth labels for training the network, they attempt to leverage the self-supervision from a pair of stereo images or monocular video sequences,
under the assumption that the geometric structure of a scene can be encoded with the reconstruction loss based on pixel-wise intensity similarities \cite{garg2016unsupervised}.
This loss function seems to be an appealing alternative to the lack of large-scale ground truth labels, but it often leads to blurry results around depth boundaries and does not consider occluded pixels \cite{godard2017unsupervised}.
%Additionally, the reconstruction loss using pixel-wise similarities does not consider structure-aware information in the monocular depth estimation.
%However, due to the limitation of reconstruction based loss function, they often result unsatisfactory depth map with blurry depth boundaries, and are conducted without respect to occluded areas.

Instead of relying on the self-supervised reconstruction loss across stereo images, Cho \textit{et al.}~\cite{cho2019large} attempted to train the monocular depth estimation network through \emph{pseudo} depth labels of the stereo images generated from pre-trained stereo matching network~\cite{pang2017cascade}. %However, \cite{cho2019large} used ground truth data when training stereo network, despite the fact that the lack of ground truth is serious problem in depth estimation. To prevent this issue in advance, we used self-supervised stereo matching \cite{watson2020learning} to produce a pseudo ground truth.
To mitigate performance degeneration by inaccurate pseudo depth labels, they leverage stereo confidence maps ($\in[0,1]$) indicating the reliability of the pseudo depth labels.
The confidence map is truncated with a threshold \cite{cho2019large, tonioni2019unsupervised} so that depth values with low confidence are excluded.
However, a fixed threshold for all training dataset still has the risk of inaccurate pseudo depth values being used in the network training \cite{cho2019large}.
The method of \cite{tonioni2019unsupervised} attempted to address this issue by learning the threshold with an additional regularization term, but the performance gain is rather limited due to its hard thresholding and the implicit constraint by the regularization term.
%the lack of an explicit supervision for the threshold learning.

%As shown in Fig. \ref{Overall_method}, it consists of the confidence network $M_C$ and the threshold network $M_T$. To cope with the prediction error of the confidence map estimated from $M_C$, we first infer the threshold $\tau$ using $M_T$.

To overcome this limitation, we propose a novel architecture that adaptively learns the threshold dynamically conditioned on the pseudo depth map.
For a given inaccurate pseudo depth map, the stereo confidence map and its associated threshold are inferred in an end-to-end manner.
The confidence map is then thresholded through a differential soft-thresholding operator controlled by the learned threshold.
The proposed threshold learning is capable of dealing with the prediction errors of the confidence map more effectively.
Note that we leverage the soft-thresholding operator to make the network differentiable. %, unlike \cite{cho2019large, tonioni2019unsupervised} based on the hard thresholding.
The thresholded confidence map is then used together with the pseudo depth labels for training the monocular depth estimation network.
Additionally, we propose to enhance the monocular depth map in a probabilistic inference framework.
Unreliable parts of the monocular depth map are identified using the uncertainty map, and these are refined through the pixel-adaptive convolution (PAC) layer \cite{su2019pixel}.
Experimental results validate that the monocular depth accuracy is significantly improved by leveraging the proposed threshold learning and probabilistic depth refinement modules.

Interestingly, the threshold learning can also be beneficial to improve the performance of existing stereo confidence estimation approaches \cite{poggi2016learning, kim2019laf}.
The confidence map obtained from the existing approaches \cite{poggi2016learning, kim2019laf} is refined through the soft-thresholding function controlled by the learned threshold.
As shown in \figref{slope_maskeq}, the soft-thresholding function attenuates low confidence values that are less than the learned threshold $\tau$ to become as close as 0 while amplifying high confidence values to converge to 1. %, encouraging the confidence to have a bimodal distribution consisting of 0 and 1.
We validate through experiments that this process improves the prediction accuracy of the existing confidence estimation approaches. %that uses only a sigmoid function to predict the confidence probability.
To sum up, our contributions are as follows.
\begin{itemize}
    \item We propose a novel framework of monocular depth estimation using pseudo depth labels generated from self-supervised stereo matching methods.%DepthNet, RefineNet, and ThresNet for monocular depth estimation using \emph{self-supervised} pseudo depth labels.
    \item We introduce the threshold network that adaptively learns the threshold of the confidence map for better predicting the reliability of the inaccurate pseudo depth labels.
    \item The monocular depth map is further refined through the probabilistic refinement module based on the PAC layer.
    \item It is shown that the threshold network can also be used to enhance the prediction accuracy of existing confidence estimation approaches.
    % Thorough experiments are provided on KITTI dataset \cite{geiger2013vision} and Cityscapes dataset \cite{cordts2016cityscapes} for the monocular depth estimation, and on Middlebury \cite{scharstein2014high} and KITTI 2015 dataset for confidence estimation.
    %    \item An intensive ablation study provides the quantitative evaluation of the proposed method.
\end{itemize}

\section{Related Work}
\noindent{{\bf Monocular depth estimation.}}
%Although challenging, estimating depth from a monocular image has several benefits, e.g., robustness to occlusions, and thus it has quickly gained popularity in recent years.
Eigen \textit{et al.}~\cite{eigen2014depth} initiated the monocular depth estimation through deep network that regresses a depth map with ground-truth depth information,
inspiring numerous approaches based on multi-scale images~\cite{li2015depth}, up-projection technique~\cite{laina2016deeper}, motion parallax~\cite{ummenhofer2017demon}, ordinal regression~\cite{fu2018deep}, and semantic divide-and-conquer~\cite{wang2020sdc}.
Despite remarkable performance over classical handcrafted approaches, they rely on abundant and high-quality ground-truth depth maps, which is costly to obtain.
%\sr{(ADD all recent ones?)}

To overcome this limitation, self-supervised learning has been introduced by leveraging other forms of supervision from stereo images and video sequences instead of ground truth depth maps.
Garg \textit{et al.}~\cite{garg2016unsupervised} used the stereo photometric reprojection.
Godard \textit{et al.}~\cite{godard2017unsupervised} further used the left-right consistency between stereo images.
Zhou \textit{et al.}~\cite{zhou2017unsupervised} proposed to leverage multi-view synthesis procedure, and this idea was extended using the feature-based warping loss in \cite{zhan2018unsupervised}. %Following this work, Zhan \textit{et al.}~\cite{zhan2018unsupervised} proposed the feature-based warping loss.
%They do not require ground-truth depth, but use stereo images or video sequence relatively easy to obtain.
%They, however, frequently lead to blurry boundaries as \sr{WHY?}.
To take advantages of both supervised and self-supervised learning methods, semi-supervised learning methods have also been presented.
Kuznietsov \textit{et al.}~\cite{kuznietsov2017semi} directly combined supervised and unsupervised loss terms.
Ji \textit{et al.}~\cite{ji2019semi} utilizes an image-depth pair discriminator with a small amount of labeled dataset, alleviating the reliance on supervision.
Recently, Gonzalebello \textit{et al.}~\cite{gonzalezbello2020forget} proposed mirrored exponential disparity (MED) probability volumes to handle occluded areas.% instead of photometric reconstruction loss or learning uncertainty masks. %Since architecture of FAL is complementary to our framework, % leveraging pseudo depth labels as self-supervision, 
%we expect that there will be a good synergy if we adopt the structure of \cite{gonzalezbello2020forget}.
%With high-quality view synthesis network, Luo \textit{et al.}~\cite{luo2018single} formulated a monocular depth estimation problem as a stereo matching problem.
%But, they still inherit the limitations of the existing unsupervised loss terms defined on stereo images or video sequence, as described above.
%In this paper, we introduce other type of self-supervised learning.

The most related to our work is the methods of Guo \textit{et al.}~\cite{guo2018learning}, Cho \textit{et al.}~\cite{cho2019large}, and Tonioni \textit{et al.}~\cite{tonioni2019unsupervised} in which a stereo matching knowledge is distilled to train a monocular depth network.
Since the disparity map estimated by stereo matching inherently contain unreliable ones, they used stereo confidence to build a pseudo-ground-truth disparity map by thresholding the confidence.
Guo \textit{et al.}~\cite{guo2018learning} used a handcrafted occlusion map sensitive to outliers.
Cho \textit{et al.}~\cite{cho2019large} used a fixed threshold empirically, but it is ineffective to use the same threshold for all images.
Unlike this, Tonioni \textit{et al.}~\cite{tonioni2019unsupervised} tried to learn the threshold by using an additional regularization term that allows it to be between 0 and 1, but it is also difficult to learn the appropriate threshold with the implicit constraint by the regularization term.
In our method, effective threshold learning is the main contribution.

\noindent{{\bf Stereo confidence estimation.}}
In parallel with the development of predicting depth from images, stereo confidence estimation has also been actively studied.
%Hu and Mordohai~\cite{hu2012quantitative} presented a comprehensive evaluation of stereo confidence measures, primarily focusing on handcrafted methods~\cite{egnal2004stereo, egnal2002detecting}.
%In early works, single-\cite{} or multi-\cite{} handcrafted confidence features have been used to detect unreliable pixels.
Machine learning approaches \cite{park2015leveraging, spyropoulos2016correctness, kim2017feature} relying on shallow classifier, e.g., random tree~\cite{breiman2001random}, enable one to classify correct and incorrect pixels.
%Further, there have been approaches\cite{} to find the optimal set of handcrafted confidence features using the permutation importance.
%Among them are techniques that utilize the permutation importance to select confidence measures~\cite{park2015leveraging,spyropoulos2016correctness} and leverage a feature augmentation that imposes spatial coherency~\cite{kim2017feature}.
%As all of these techniques use handcrafted measures and shallow classifiers, they lack robustness to challenging circumstances, \sr{e.g., .}
Recently, deep convolutional neural network (CNN)-based approaches have become a mainstream.
Various methods have been proposed that use the single- or bi-modal input, e.g., disparity~\cite{poggi2016learning}, left and right disparities~\cite{seki2016patch}, 3D matching cost~\cite{shaked2017improved}, 3D matching cost and disparity~\cite{kim2018unified}, and disparity and color image~\cite{tosi2018beyond, fu2018learning}.
Kim \textit{et al.}~\cite{kim2019laf} proposed to make full use of the tri-modal input in conjunction with locally adaptive attention and scale networks, achieving state-of-the-art prediction accuracy.
All of these techniques require ground truth depth maps and have been used to refine a depth (or disparity) map with a fixed threshold which is set empirically. Poggi et al.~\cite{poggi2020self} introduced a method for learning self-supervised confidence measure with various criterions.

%As shown in Fig. \ref{Overall_method}, the threshold $\tau$ for the confidence map is adaptively learned from the threshold network. The threshold network is learned using a commonly used cross-entropy loss.
\begin{figure}[t!]
    \centering
    \includegraphics[width=1\columnwidth]{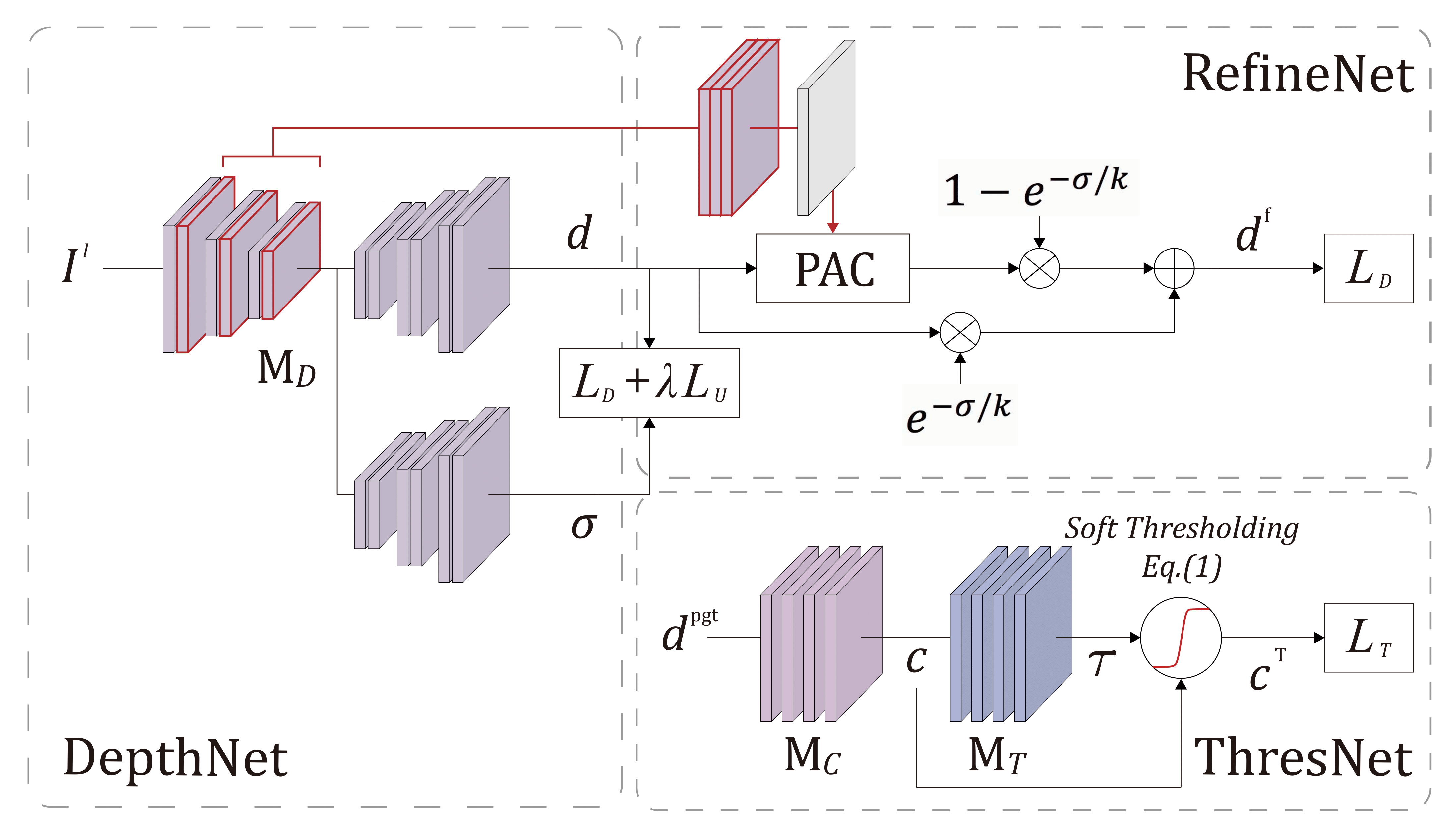}
    \vspace{-0.5cm}
    \caption{The proposed architecture consisting of ThresNet, DepthNet, and RefineNet.
        Given a pair of stereo images, the pseudo ground truth depth map $d^{\mathrm{pgt}}$ is precomputed using a self-supervised stereo matching network.
        The proposed model training begins with $d^{\mathrm{pgt}}$ by computing its confidence map $c$ and the threshold $\tau$ through the ThresNet.
        The thresholded confidence map $c^{\mathrm{T}}$ is obtained using the soft-thresholding function.
        The DepthNet that infers the monocular depth map $d$ and uncertainty map $\sigma$ is trained by minimizing an objective defined using $d^{\mathrm{pgt}}$ filtered out by $c^{\mathrm{T}}$.
        The monocular depth map $d$ is finally refined through the probabilistic refinement module based on the pixel-adaptive convolution (PAC) layer in the RefineNet.
    }
    \label{Overall_method}
    \vspace{-0.3cm}
\end{figure}

% [FU_paper] Z. Fu and M. A. Fard. Learning confidence measures by multi-modal convolutional neural networks. in Proc. IEEE Winter Conf. Applicat. Comput. Vis., pages 1321--1330, 2018.

%%%%%%%%%%%%%%%%%%%%%%%%%%%%%%%%%%%%%%%%%%%%%%%%%%%%%%%%%%%%%%%%%%%%%%%%%%%%%%%%%%%%%%%%%%%%%%%%%%%%%%%%%%%%%%%%%%%%%%%%%%%%%%%%%%%%
%DB: The following is redundant, but may be used for journal extension
%Additionally, there is no consideration for structure-aware information due to the limitations of pixel-wise similarities (e.g., intensity L1 norm and SSIM). %Also, the pixel-wise reconstruction loss does not take into account structure-aware information
%In \cite{ICLR2020}, the semantic structure computed from fixed pretrained semantic segmentation networks, beyond the pixel-level supervision, is directly imposed via pixel-adaptive convolutions based on segmentation features \cite{Pixel_2019}. %, when guiding geometric representation learning.
%This work shows impressive results, but they heavily rely on the semantic features for inferring a monocular depth map. %that address blurry results around object boundaries and occlusions,

%[Pixel_2019] Pixel-adaptive convolutional neural networks, CVPR 2019
%%%%%%%%%%%%%%%%%%%%%%%%%%%%%%%%%%%%%%%%%%%%%%%%%%%%%%%%%%%%%%%%%%%%%%%%%%%%%%%%%%%%%%%%%%%%%%%%%%%%%%%%%%%%%%%%%%%%%%%%%%%%%%%%%%%%

\section{Proposed Method}
%Self-supervised learning for monocular depth estimation has become a prevalent strategy to the problem of insufficient ground truth labels by leveraging other forms of supervision from stereo images and monocular video. %After the pioneering work of \cite{Godard_2017}, numerous approaches have been proposed over the fast few years by exploiting the reconstruction loss in a pair of stereo images or monocular video \cite{}.
%Its reconstruction loss, however, often leads to over-smoothed results across object boundaries and is incapable of handling occluded pixels explicitly.
%We propose a new approach to leverage the \emph{pseudo} depth labels from a pair of stereo images as supervision for monocular depth estimation.

Unlike recent self-supervised monocular depth estimation approaches \cite{garg2016unsupervised, godard2017unsupervised, luo2018single, godard2019digging, poggi2020uncertainty}, we leverage the \emph{pseudo} depth labels from a pair of stereo images as supervision for monocular depth estimation.
\figref{Overall_method} shows the overall procedure of the proposed method consisting of three networks, including DepthNet, RefineNet, and ThresNet.

The proposed model training begins with the pseudo depth labels $d^{\mathrm{pgt}}$ precomputed using the self-supervised stereo matching method \cite{watson2020learning}.
Note that among various options provided in \cite{watson2020learning} for data synthesis, we adopted ‘Monodepth2’ \cite{godard2019digging} which is self-supervised monocular depth network. %, not MiDaS \cite{ranftl2019towards}.
Its confidence map $c$ is estimated by the confidence estimation module $M_C$, aiming at preventing the abuse of erroneous depth values in training the monocular depth network.
To take into account the prediction errors of the confidence map itself, we further learn the threshold $\tau$, truncating the confidence map, adaptively through the threshold module $M_T$. %To make the threshold learning differentiable, we approximate the thresholding operation using $\tau$ with a differential function, called soft-thresholding.
The thresholded confidence map $c^\mathrm{T}$ is obtained via the soft-thresholding by the learned threshold $\tau$. %The confidence map is then modulated (thresholded) by the learned threshold $\tau$.
This operation encourages to trust the pixel with a higher confidence value than a specific $\tau$ value.
The DepthNet is trained by minimizing an objective defined using the pseudo depth labels $d^{\mathrm{pgt}}$ filtered out by the thresholded confidence map $c^\mathrm{T}$.
Finally, our method refines the monocular depth map $d$ through the probabilistic refinement module based on the PAC layer \cite{su2019pixel} in the RefineNet.

%In the following, more details are presented, including the network architectures, loss functions, and training details.

%%%The overall procedure of Fig. \ref{Overall_method} is summarized as follows.
%%%With stereo image pairs and very sparse LiDAR depth maps of 3$\%$ density provided by KITTI benchmark,
%%%the pseudo depth label is first generated by the pretrained stereo matching method.
%%%Its confidence map $c_l$ and threshold $\tau$ are estimated from the confidence network and threshold network, respectively.
%%%The thresholded confidence map $p_l$ manipulated by $\tau$ is then used to compute the depth regression loss function together with the pseudo depth labels.

%The threshold network $M_T$ resembles the encoder of the monocular depth estimation network.
%Convolutional features from the encoder of the monocular depth estimation network $M_D$ are concatenated to the threshold network $M_T$ for reflecting image characteristics.

\subsection{Network Architecture}
\subsubsection{ThresNet}
The ThresNet predicts the confidence map of the inaccurate pseudo depth label and its threshold in an adaptive manner and then generates the thresholded confidence map via the soft-thresholding function.
For the confidence estimation network $M_C$, we adopted the CCNN \cite{poggi2016learning} thanks to its simplicity,
but more sophisticate models \cite{poggi2016learning, kim2019laf, tosi2018beyond} can also be utilized as a backbone.
%We trained CCNN with both self-supervised manner \cite{poggi2020self} and supervised manner using extremely sparse LiDAR depth maps of $3\%$ density \cite{poggi2016learning}.
The threshold network $M_T$ consists of four convolutional layers, followed by global average pooling and $1\times1$ convolution.
%More details on the network architecture will be provided in the supplementary material.

\begin{figure}[t]
	\centering
	\begin{subfigure}[Images with high $\tau$ values]
		{\includegraphics[width=1.0\columnwidth]{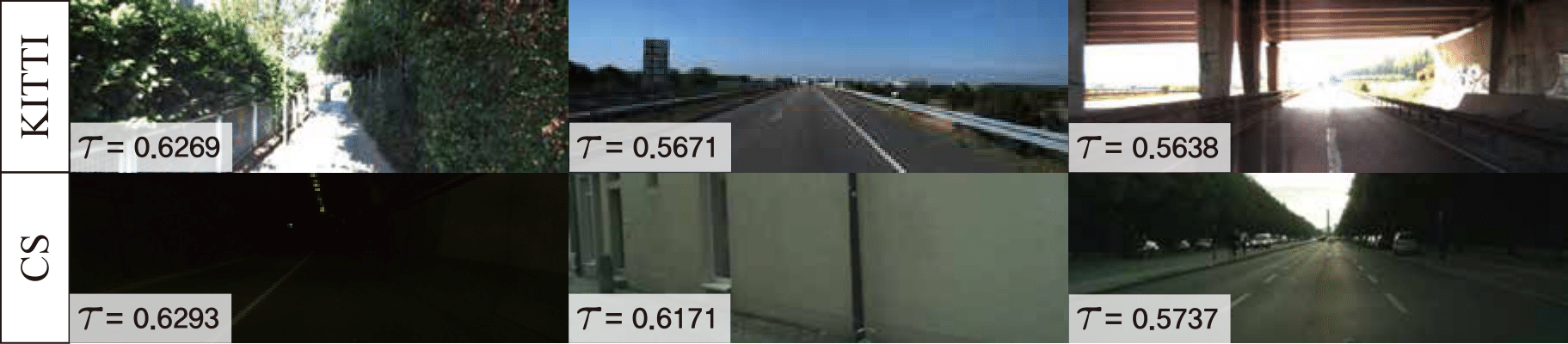}}
	\end{subfigure}
	\begin{subfigure}[Images with low $\tau$ values]
		{\includegraphics[width=1.0\columnwidth]{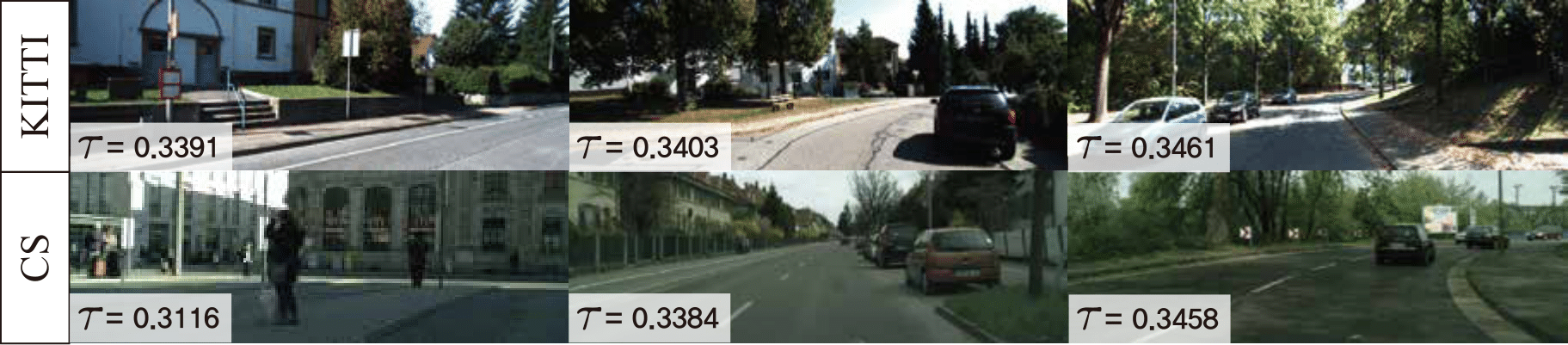}}
	\end{subfigure}
	%\vspace{-0.3cm}
	\caption{Examples of learned threshold $\tau$ by our threshold learning. CS indicates the Cityscapes dataset.}
	\label{fig:tau_images}
	\vspace{-0.35cm}
\end{figure}

The estimated confidence map $c$ is modulated by the threshold $\tau$, such that a depth value with a higher confidence value than a specific $\tau$ value assumes to be trustworthy.
A key issue is how to set accordingly $\tau$ which needs to vary depending on images.
This threshold $\tau$ should be set low in the image where depth inference is easy while being set high in the opposite case (see \figref{fig:tau_images}).
We approximate the thresholding operation with a smooth, differentiable function. %Assuming the threshold can be learned according to image characteristics,
The thresholded confidence map $c^\mathrm{T}$ is computed using the differentiable soft-thresholding function as follows:

\begin{figure}[t!]
    \centering
    \begin{subfigure}[]
        {\includegraphics[width=0.340\columnwidth]{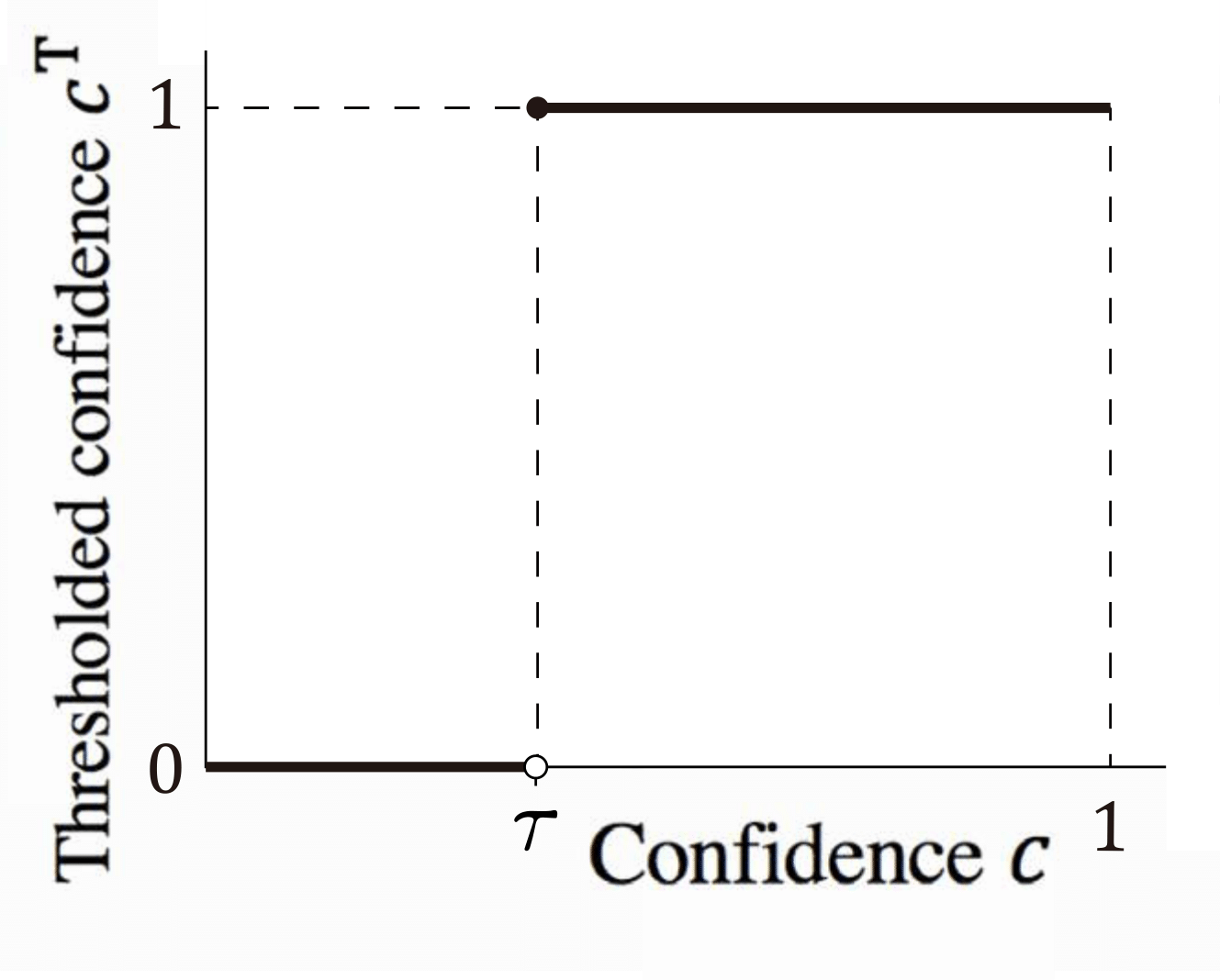}}
    \end{subfigure}
    \hspace{-0.4cm}
    \begin{subfigure}[]
        {\includegraphics[width=0.340\columnwidth]{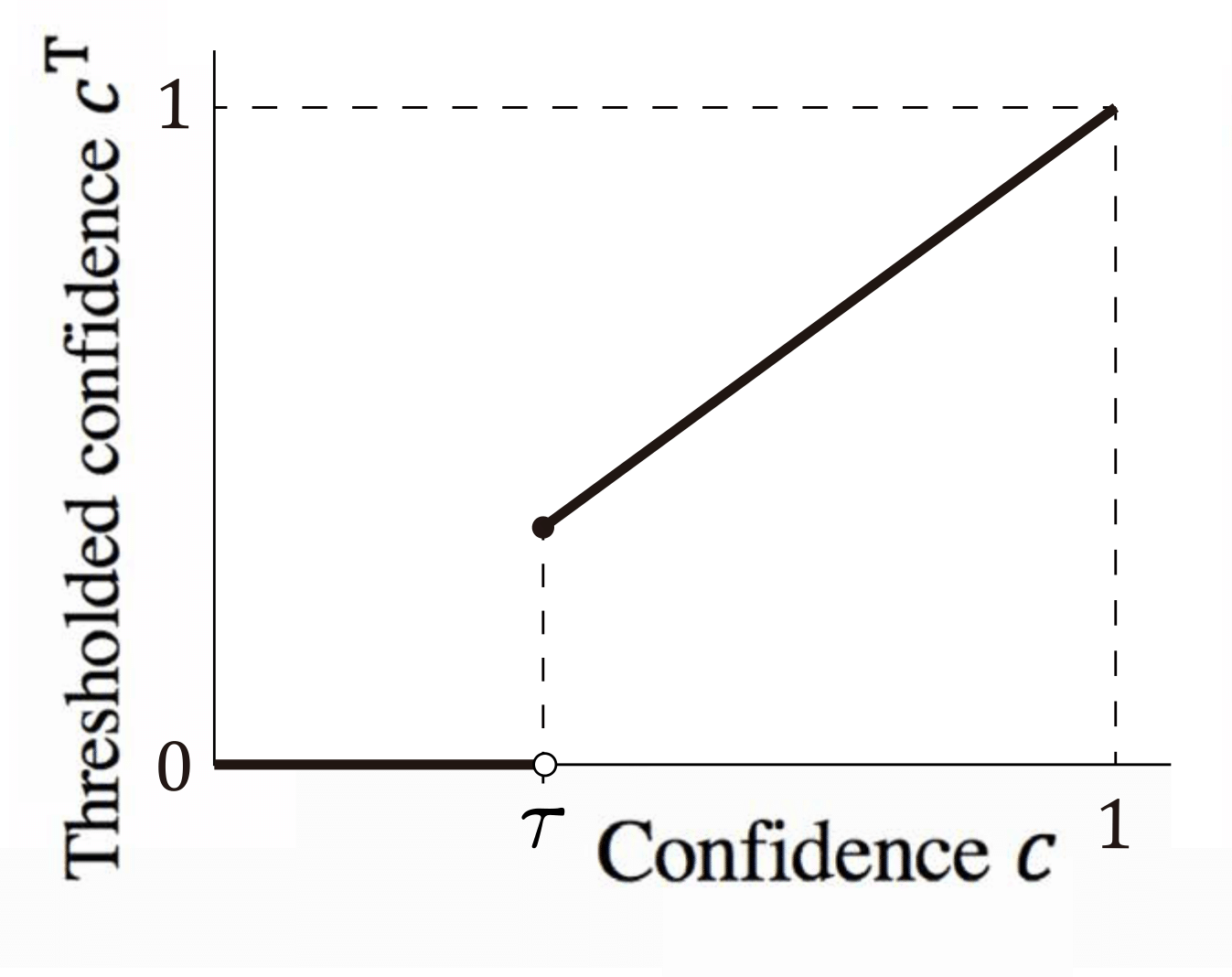}}
    \end{subfigure}
    \hspace{-0.4cm}
    \begin{subfigure}[]
        {\includegraphics[width=0.340\columnwidth]{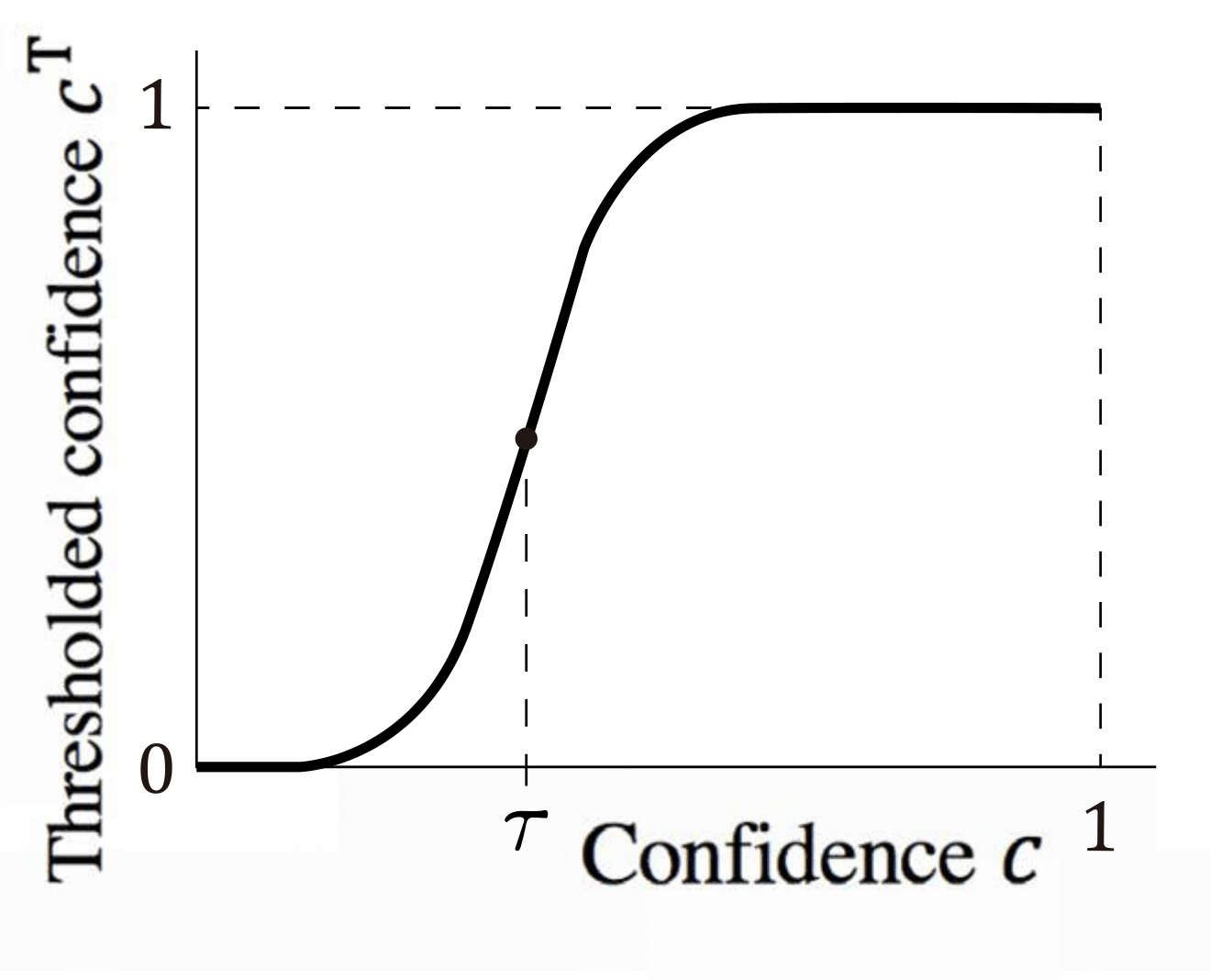}}
    \end{subfigure}
    %\vspace{-0.3cm}
    \caption{Comparison of confidence thresholding operator: (a) hard-thresholding used in \cite{cho2019large},
        (b) hard-thresolding function used in \cite{tonioni2019unsupervised}, and (c) our soft-thresholding function in \eqref{eq:thresholding}.
        The learned threshold is used in (b) and (c), while the threshold is fixed in (a) for all training images.
    }
    \label{slope_maskeq}
    \vspace{-0.3cm}
\end{figure}

\begin{equation}
c^\mathrm{T}_{p}(\tau) = {\frac{1}{1 + e^{-\varepsilon \cdot {(c_p - \tau)}}}},
\label{eq:thresholding}
\end{equation}
%where $c_p$ and $c^\mathrm{T}_p$ indicate confidence and thresholded values at

\noindent where $p$ represents a pixel. The slope of the thresholded confidence map $c^\mathrm{T}$ is adjusted by a hyperparameter $\varepsilon$, which is a positive constant.
Too large $\varepsilon$ changes the soft-thresholding function too rapidly (e.g. $\varepsilon=90$), often making it non-differentiable. We set $\varepsilon=10$ in experiments.
The pixel-varying confidence map is transformed with the per-image threshold $\tau$.
We also investigated a pixel-varying threshold map $\tau_p$, but its performance gain was negligible.
%Since the pixel-varying confidence map is transformed with the threshold, the per-image threshold is enough to guarantee a pixel adaptivity.

\figref{slope_maskeq} compares the confidence thresholding functions.
In \figref{slope_maskeq} (a), the confidence threshold $\tau$ is fixed with a predefined value for all training images without considering image characteristics, often causing inaccurate pseudo depth values to be used during training.
In \figref{slope_maskeq} (b), it is learned using an additional regularization term \cite{tonioni2019unsupervised},
but its performance gain on the monocular depth estimation is rather limited, as reported in the original paper \cite{tonioni2019unsupervised}. %due to the lack of explicit supervision for the threshold learning
The proposed differential soft-thresholding function, controlled by the threshold $\tau$ dynamically conditioned on the pseudo depth map, leads to superior performance on the monocular depth estimation, when the threshold loss $L_T$ is used together.
The ablation study of the confidence thresholding operators is provided in experiments.

%The larger $\varepsilon$, the tighter $c^\mathrm{T}$ values are mapped to 0 or 1.
%This function can also be used to enhance the performance of existing confidence estimation approaches (Section 4).

\figref{fig:tau_images} presents the estimation results of the ThresNet for the KITTI and Cityscape datasets~\cite{cordts2016cityscapes}.
The threshold $\tau$ becomes higher in images where stereo matching do not work well, and vice versa. %varies accordingly depending on the image contents, producing a lower $\tau$ in the image where geometric reasoning is easy and vice versa.
This indicates that the ThresNet is beneficial to improving the monocular depth network by excluding unreliable pseudo depth values effectively.
%The effectiveness of our method will be validated through intensive performance evaluation and ablation study.

%\figref{Toy_example} shows examples of how the monocular depth estimation is gradually improved.
%Some pixels of the pseudo depth maps are inaccurate to be used solely as the supervision for the monocular depth network.
%By leveraging the thresholded confidence map $c^\mathrm{T}$, our method successfully excludes unconfident pixels of inaccurate pseudo ground truth from training, producing better results.

\subsubsection{DepthNet and RefineNet}
The DepthNet and RefineNet infer and refine the monocular depth map by leveraging the pseudo depth labels, masked out by the thresholded confidence map, as supervision.
The DepthNet is based on the encoder-decoder architecture \cite{ronneberger2015u}, in which an encoder takes an image and two decoders estimate the monocular depth map $d$ and its uncertainty map $\sigma$.
The uncertainty map, indicating the variance of the predicted monocular depth map, becomes higher when the prediction is unreliable, and vice versa.
The encoder network consists of the first 13 convolution layers of the VGG network \cite{simonyan2014very}, and the decoder is symmetrical with the encoder. %After every convolution layers, the ReLu layer follows.
The uncertainty map $\sigma$ is used to refine the monocular depth map in the subsequent RefineNet.

We first upsample $L$ feature maps (here $L=4$) from the encoder of the DepthNet to an original resolution and concatenate them.
The concatenated features are then fused by passing through $1\times1$ convolution, generating a guidance feature $g$.
The estimated monocular depth map $d$ is finally fed into the PAC layer \cite{su2019pixel} with the guidance of the feature map $g$.
Unlike the original PAC module that directly infers refined results, we leverage the residual connection that takes into account the uncertainty map $\sigma$ for predicting the refined monocular depth map $d^f$ such that
\begin{equation}
d^f  = e^{-\sigma /k}  \cdot d + (1 - e^{-\sigma /k} )d'
\label{eq:depth_PAC}
\end{equation}
where $d'$ indicates the output of the PAC layer. $k$ is a hyperparameter to control the refinement through the PAC layer, and it was set to 1.

It should be noted that though some monocular depth estimation approaches \cite{poggi2020uncertainty, brickwedde2019mono} have attempted to measure the uncertainty of the monocular depth estimation through deep network,
our method proposes to infer the uncertainty map and use it for a subsequent refinement module.
This framework can also be extended into various pixel-level labeling tasks based on the uncertainty prediction.

%\begin{figure}[t]
%    \centering
%
%    \begin{subfigure}[]
%        {\includegraphics[width=0.245\columnwidth]{img/ex_a.pdf}}
%    \end{subfigure}
%    \hspace{-0.2cm}
%    \begin{subfigure}[]
%        {\includegraphics[width=0.245\columnwidth]{img/ex_c.pdf}}
%    \end{subfigure}
%    \hspace{-0.2cm}
%    \begin{subfigure}[]
%        {\includegraphics[width=0.245\columnwidth]{img/toyex_new.png}}
%    \end{subfigure}
%    \hspace{-0.2cm}
%    \begin{subfigure}[]
%        {\includegraphics[width=0.245\columnwidth]{img/ex_d.pdf}}
%    \end{subfigure}
%    \vspace{-0.3cm}
%    \caption{Examples of gradually improved depth results of (a) input image, from (b) using the monocular depth network with fixed threshold $\tau = 0.3$ \cite{cho2019large}, (c) using the monocular depth network with learned threshold $\tau$ \cite{tonioni2019unsupervised}, and to (d) using the proposed method.}
%    \label{Toy_example}
%    \vspace{-0.25cm}
%\end{figure}

\subsection{Loss Functions}
\label{sec:loss}
%It is crucial to train the threshold network $M_T$ attentively.

\subsubsection{Thresholding loss}
The ThresNet with confidence and threshold networks can be trained in a supervised manner~\cite{poggi2016learning} or a self-supervised manner~\cite{poggi2020self}. For the supervised training, we propose to use the sparse ground truth depth data provided by public benchmarks. For instance, we can leverage extremely sparse LiDAR depth maps of 3$\%$ density provided with a set of stereo image pairs in the KITTI dataset. The ground truth of the thresholded confidence map is generated using the sparse ground truth depth data like existing confidence estimation approaches~\cite{kim2019laf} and this is used to train the ThresNet using a cross-entropy loss $L_T$. More details on the ground truth confidence map are provided in the supplementary material.
Alternatively, the ThresNet can be trained in the self-supervised manner without using the LiDAR depth maps. Following~\cite{poggi2020self}, we generate the pseudo ground truth of the thresholded confidence map according to various criterions (e.g., reprojection error, disparity agreement). The loss $L_T$ for the self-supervised training is defined as a multi-modal binary cross entropy loss of \cite{poggi2020self}. In \tabref{Quantative_KITTI}, we compare the monocular depth accuracy when using the supervised and self-supervised ThresNets, and found the accuracy is almost similar.

In \cite{tonioni2019unsupervised}, the threshold is also learned to exclude depth values with low confidences when training their network.
It was reported that when using the depth regression loss only, the threshold $\tau$ would converge to 1 for masking out all pixels \cite{tonioni2019unsupervised}. %when simply adding the threshold in the depth regression loss,
Thus, they propose to include an additional regularization loss, $-\mathrm{log}(1-\tau)$, that prevents the threshold $\tau$ from approaching 1.
Though this term allows $\tau$ to be between 0 and 1, it does not guarantee to yield accurate prediction results of the threshold $\tau$.
Contrastingly, our method attempts to learn the threshold $\tau$ with the soft-thresholding function and the explicit supervision.
We will verify the effectiveness of our threshold learning approach in the ablation study of \tabref{table:ablation_method1}.
%However, such a regularization strategy with no appropriate supervision does not necessarily yield accurate results.
%However, the threshold learning approach using stereo image pairs only is not a principled solution
%simply encouraging the threshold $\tau$ to be less than 1

\subsubsection{Depth regression loss}
A monocular depth map from the DepthNet is leveraged to compute a confidence-guided depth regression loss $L_D$ assisted by the thresholded confidence map $c^\mathrm{T}$ as follows:
\begin{equation}
L_D = {\frac{1}{Z} \sum\limits_{p\in \Omega} {c^\mathrm{T}_p(\tau) \cdot |d_p - d^{\mathrm{pgt}}_p|}},
\label{eq:depth_loss_LD}
\end{equation}

\noindent where $d$ and $d^{\mathrm{pgt}}$ indicate the predicted depth map and pseudo ground truth depth map, respectively.
$\Omega$ represents a set of all pixels. The loss $L_D$ is normalized with $Z=\sum\limits c^\mathrm{T}_p(\tau)$.

Additionally, we leverage the negative log-likelihood minimization to infer the uncertainty of the network output.
The predictive distribution of the network output $d$ can be modelled as the Laplacian likelihood \cite{kendall2018multi,ilg2018uncertainty,kendall2017uncertainties} as follows:
\begin{equation}
L_U = \frac{1}{{|\Omega |}}\sum\limits_{p \in \Omega } {\left( {\frac{{|d_p  - d_p^\mathrm{pgt} |}}{{\sigma _p }} + \log \sigma _p } \right)},
\label{eq:depth_loss_LU}
\end{equation}

\noindent where the variance $\sigma$ represents the uncertainty map of the predicted depth map.
The logarithmic term $\log \sigma$ prevents $\sigma$ from approaching to infinity \cite{kendall2018multi}.
We combine two losses $L_D$, taking into account the reliability of the pseudo ground truth depth map $d^{\mathrm{pgt}}$, and $L_U$ predicting the uncertainty of the predicted depth map $d$, such that %, and $L_V$ imposing cross-view consistency across left and right images
\begin{equation}
L = L_D + \lambda L_U,
\label{eq:depth_loss}
\end{equation}
where $\lambda$ represents hyperparameter that balances two losses which is experimentally determined to $10^{-3}$.
This enables for modeling the uncertainty of the monocular depth estimation network while considering the confidence of the pseudo depth label.
As shown in \figref{Overall_method}, the DepthNet that infers both the monocular depth map and uncertainty map is trained with $L$ in \eqref{eq:depth_loss},
while the RefineNet leverages $L_D$ in \eqref{eq:depth_loss_LD} as it predicts the final monocular depth map only.

\begin{table*}[]
	\centering
	\small\addtolength{\tabcolsep}{-1pt}
	\caption{Quantitative evaluation for depth estimation with existing methods on KITTI Eigen Split \cite{eigen2014depth} dataset. Numbers in bold and underlined represent $1^{st}$ and $2^{nd}$ ranking, respectively. `Ours$^{\dagger}$’ is obtained using the self-supervised	ThresNet~\cite{poggi2020self}, while `Ours’ indicates the results obtained using the supervised ThresNet.}	
%	The proposed method marked with `${\dagger}$' represents ThresNet trained with self-supervised manner \cite{poggi2020self}.}
	\begin{tabular}{@{}ccccccccccc@{}}
		\toprule
		&               &        &              & \multicolumn{4}{c}{Lower is better}                       & \multicolumn{3}{c}{Accuracy: higher is better}    \\ \midrule
		\multicolumn{1}{c|}{Method} & \multicolumn{1}{c|}{Data} & $\#$p & \multicolumn{1}{c|}{time} & Abs Rel & Sqr Rel & RMSE  & \multicolumn{1}{c|}{RMSE log} & $\delta<1.25$ & $\delta<1.25^2$ & $\delta<1.25^3$ \\ \midrule
		Monodepth \cite{godard2017unsupervised}                   & S    & 56M &9.4ms                              & 0.138   & 1.186   & 5.650 & 0.234                         & 0.813         & 0.930           & 0.969           \\
		%StrAT                       & S                                   & \textit{Self.}           & 0.128   & 1.019   & 5.403 & 0.227                         & 0.827         & 0.935           & 0.971           \\
		%3Net              & S                                 & \textit{Self.}           & 0.126   & 0.961   & 5.205 & 0.220                         & 0.835         & 0.941           & 0.974           \\
		Monodepth2 \cite{godard2019digging}                  & S    & 14M    & 2.9ms                 & 0.108   & 0.842   & 4.891 & 0.207                         & 0.866         & 0.949           & 0.976           \\
		Uncertainty \cite{poggi2020uncertainty}       & S              & 14M    & 3.6ms               & 0.107   & 0.811   & 4.796 & 0.200                         & 0.866         & 0.952           & 0.978           \\
		%Uncertainty$\ddagger$      & S                                  & \textit{Self.}           & 0.107   & 0.806   & 4.798 & 0.199                       & 0.866         & 0.952           & 0.978           \\
		MonoResMatch \cite{tosi2019learning}                & S    & 41M    & 8.3ms                & 0.111   & 0.867   & 4.714 & 0.199                         & 0.864         & 0.954           & 0.979           \\
		DepthHint \cite{watson2019self}         & S     & 33M    & 6.6ms        & 0.102   & 0.762   & 4.602 & 0.189                         & 0.880         & \underline{0.960}           & \underline{0.981}           \\
		PackNet-SfM \cite{guizilini20203d}            & M      & 122M    & 9.5ms     & 0.111   & 0.785   & 4.601 & 0.189                       & 0.878         & \underline{0.960}           & \textbf{0.982}           \\
		Insta-DM \cite{lee2021learning}                & M    & 14M    & 3.0ms                & 0.112   & 0.777   & 4.772 & 0.191                         & 0.872         & 0.959           & \textbf{0.982}           \\
		Ours (D)                 & S     & 28M    & 6.8ms              & \underline{0.099}   & 0.652   & 4.266 &0.187                         & \underline{0.883}         & \underline{0.960}           & \underline{0.981}           \\
		Ours (D+R)                 & S   & 42M    & 8.2ms              & \textbf{0.096}   & \underline{0.627}   & \textbf{4.201} & \underline{0.186}                         & \textbf{0.885}         & \textbf{0.961}           & \textbf{0.982} \\
		Ours$^{\dagger}$ (D)                 & S     & 28M    & 6.8ms              & 0.100   & 0.644   & 4.251 &0.187                         & 0.882         & \underline{0.960}           & \underline{0.981}           \\
		Ours$^{\dagger}$ (D+R)                 & S   & 42M    & 8.2ms              & 0.098   & \textbf{0.621}   & \underline{4.215} & \textbf{0.185}                         & \textbf{0.885}         & \textbf{0.961}           & \textbf{0.982}           \\ \bottomrule
	\end{tabular}
	\label{Quantative_KITTI}
	\vspace{-0.3cm}
\end{table*}

\begin{figure*}[bt!]
	\centering
	\begin{subfigure}[]
		{\includegraphics[width=0.16\textwidth]{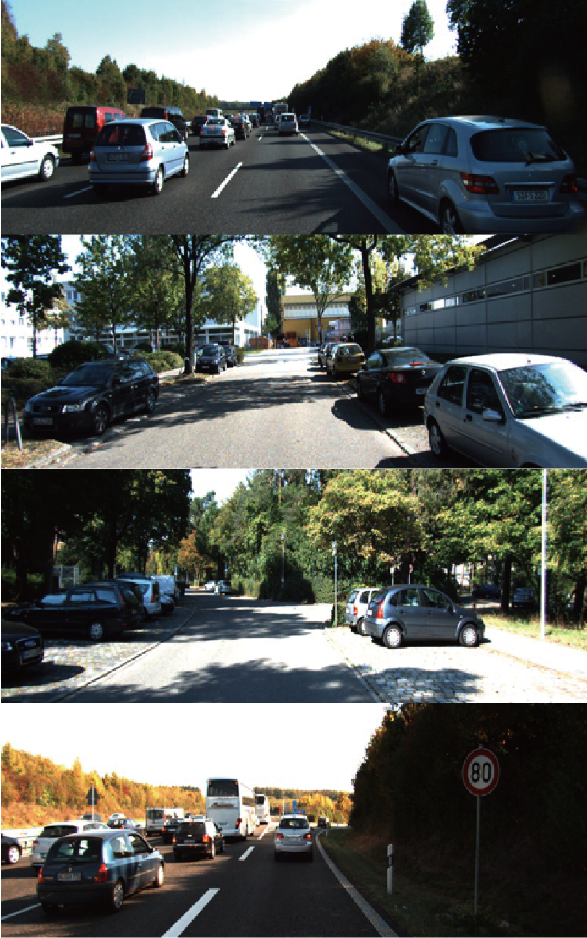}}
	\end{subfigure}
	\hspace{-0.2cm}
	\begin{subfigure}[]
		{\includegraphics[width=0.16\textwidth]{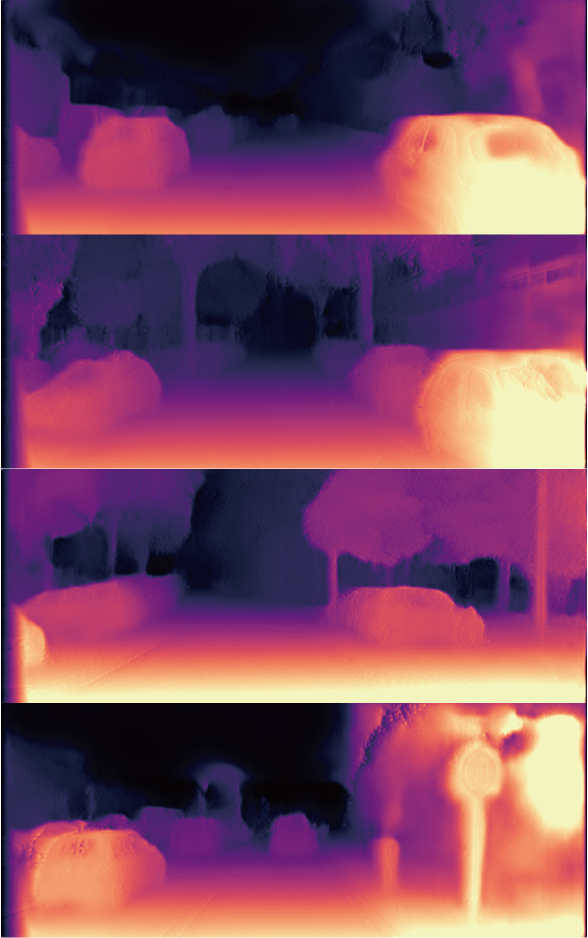}}
	\end{subfigure}
	\hspace{-0.2cm}
	\begin{subfigure}[]
		{\includegraphics[width=0.16\textwidth]{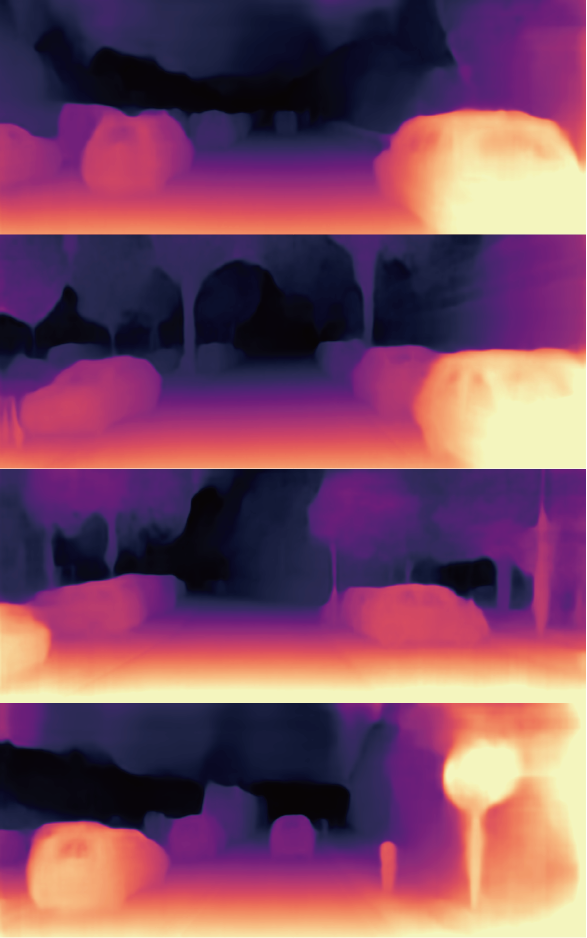}}
	\end{subfigure}
	\hspace{-0.2cm}
	\begin{subfigure}[]
		{\includegraphics[width=0.16\textwidth]{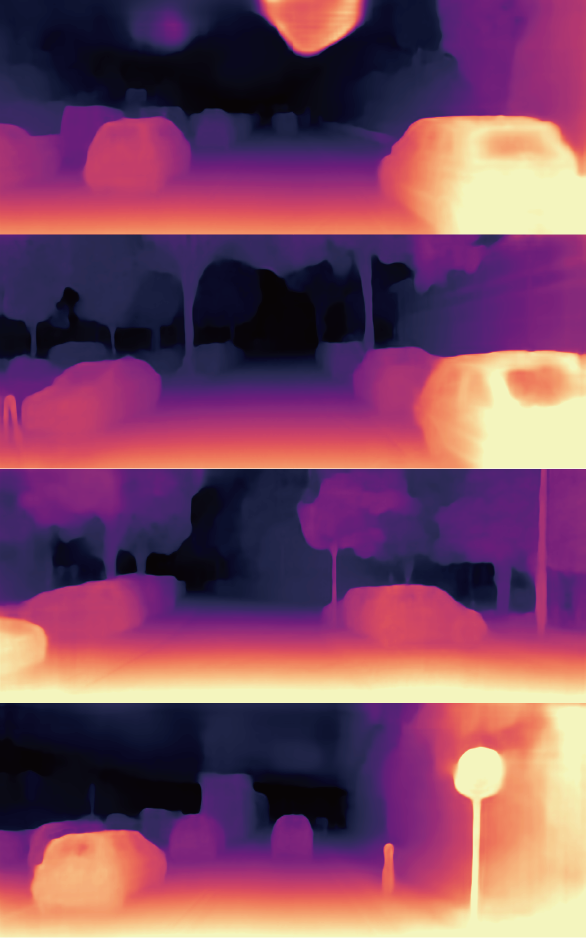}}
	\end{subfigure}
	\hspace{-0.2cm}
	\begin{subfigure}[]
		{\includegraphics[width=0.16\textwidth]{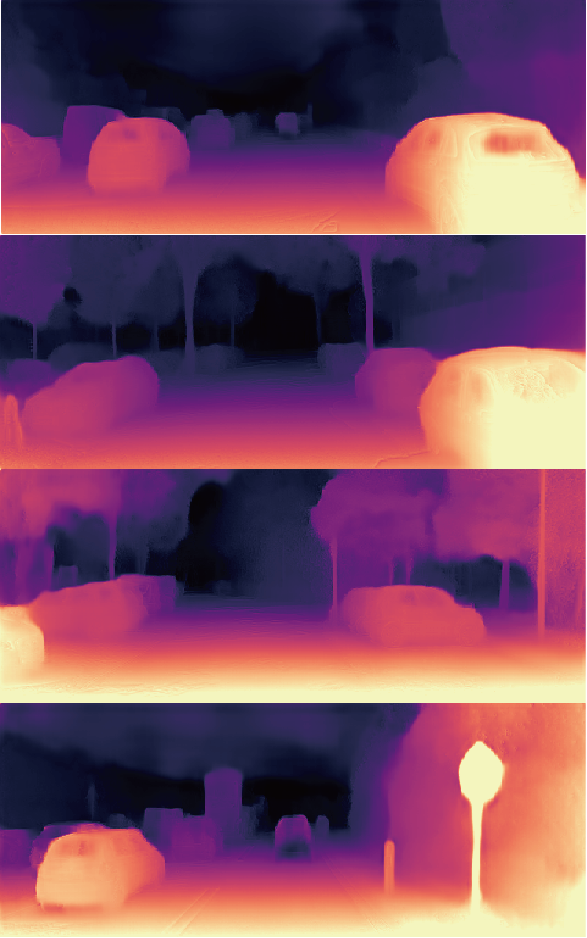}}
	\end{subfigure}
	\hspace{-0.2cm}
	\begin{subfigure}[]
		{\includegraphics[width=0.16\textwidth]{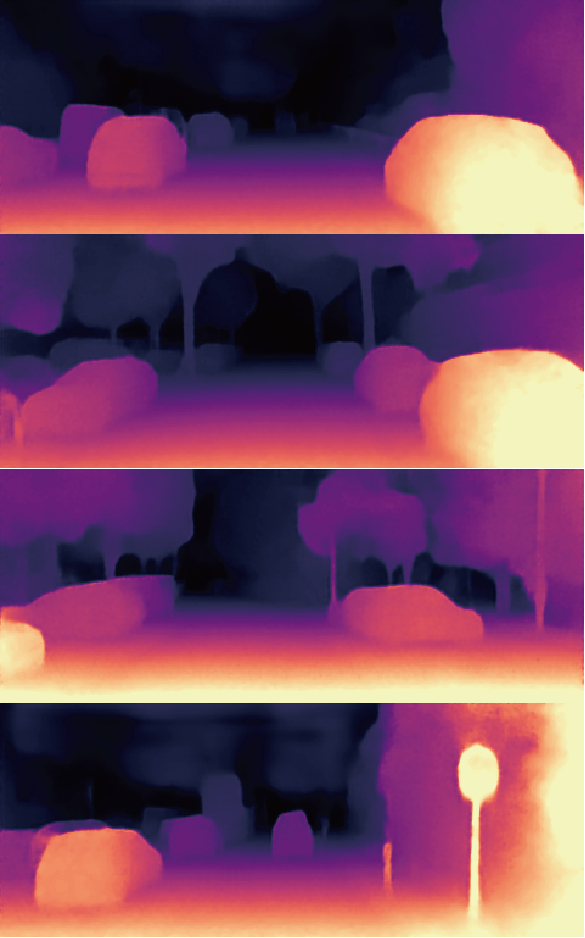}}
	\end{subfigure}
	\caption{Qualitative evaluation with existing monocular depth estimation methods on the Eigen Split \cite{eigen2014depth} of KITTI dataset: (a) input image, (b) Monodepth \cite{godard2017unsupervised}, (c) Monodepth2 \cite{godard2019digging}, (d) DepthHint \cite{watson2019self}, (e) PackNet-SfM \cite{guizilini20203d} and (f) Ours (D+R).}
	\label{Depth_Qual_KITTI}
	\vspace{-0.2cm}
\end{figure*}

%%%%[aaa] Multi-Task Learning Using Uncertainty to Weigh Losses for Scene Geometry and Semantics, CVPR 2018
%%%%[bbb] Uncertainty estimates and multi-hypotheses networks for optical flow, ECCV 2018
%%%%[ccc] What uncertainties do we need in bayesian deep learning for computer vision?, NIPS 2017

%%%Dongbo: Reserver this for journal extension
%Training with only individual view may often lead to the mismatched problem when estimating depth. Therefore, we applied cross-view consistency loss $L_V$ in RefineNet, in order to address the issue.
%\begin{equation}
%L_{V} = {\frac{1}{Z} \sum\limits_{p\in \Omega} {c^\mathrm{T}_p(\tau) \cdot (|d_{p,l} - d_{p,r\rightarrow l}| + |d_{p,r} - d_{p,l\rightarrow r}|)}},
%\label{eq:cross_view_loss_LV}
%\end{equation}
%where $d_l$ and $d_r$ are predicted depth map of left and right image, and $d^{a\rightarrow b}$ represents that depth map $d^a$ is warped onto the different view $b$ with $d^{pgt}_b$.

\subsection{Training Details}
 
%by making use of the uncertainty map $\sigma$ that is inferred together from the DepthNet.

%Similar to existing confidence estimation approaches \cite{poggi2016learning, park2015leveraging, spyropoulos2016correctness, kim2019laf},
%the ThresNet leverages the sparse ground truth depth map for training the confidence and threshold networks by minimizing $L_T$,
%while the DepthNet and RefineNet are trained by minimizing $L$ and $L_D$ defined using the pseudo depth labels precomputed from stereo images.
%The DepthNet and RefineNet are trained by minimizing $L$ and $L_D$ defined using the pseudo depth labels precomputed from stereo images.
In our work, the DepthNet and RefineNet are trained simultaneously by minimizing $L$ and $L_D$, while the ThresNet consisting of confidence and threshold networks is trained solely by minimizing $L_T$, similar to existing confidence estimation approaches \cite{poggi2016learning, park2015leveraging, spyropoulos2016correctness, kim2019laf}.
Though the whole networks can be trained end-to-end, we found through experiments that the performance gain over the separate training is relatively marginal.
%For training the ThresNet, we utilize the extremely sparse depth maps of $3\%$ density, provided from KITTI dataset \cite{geiger2013vision} for supervised manner, and the dense pseudo confidence maps for self-supervised manner.}
%Note that extremely sparse depth maps (e.g., $3\%$ density) are enough to train the ThresNet.
%The thresholded confidence map is leveraged to exclude depth values with low confidences in the pseudo depth labels when training the DepthNet and RefineNet.

It has been reported in literature \cite{poggi2016learning,tosi2018beyond} that the confidence network trained with one dataset exhibits a good generalization capability for another dataset.
In a similar context, our confidence and threshold networks trained with the KITTI dataset show satisfactory generalization capability for different datasets.
%In addition, the confidence and threshold networks play a role of assisting the monocular depth estimation network by identifying reliable depth values from pseudo ground truth depth maps.
Taking these into account, we transfer the knowledge learned from one dataset to another.
To be specific, when only stereo image pairs are available for training (e.g. Cityscape dataset),
the DepthNet and RefineNet are trained via the minimization of $L$ and $L_D$, with the ThresNet being frozen with the parameters trained with the KITTI dataset.
As shown in \figref{fig:tau_images}, the ThresNet trained with the KITTI dataset produces appropriate thresholds for both the KITTI and Cityscape datasets.
%Our code will be publicly available later.

%We found that training the networks with depth maps tends to produce better results than training with disparity maps, and thus all experiments were conducted on the depth domain.
%The conversion between disparity and depth maps is straightforward using the focal length and baseline of a stereo camera.

%\begin{figure}[t!]
%    \centering
%    \includegraphics[width=0.8\columnwidth]{img/ConfMethod.pdf}
%    \caption{The network architecture for confidence estimation boosted by the soft-thresholding.}
%    \label{fig:conf_thres}
%\end{figure}

%Training the sub-networks ($M_D$, $M_C$, and $M_T$) simultaneously from scratch is very challenging due to the imbalanced supervision.

%For the training dataset with no sparse ground truth depth maps (e.g. Cityscape dataset), the confidence and parameter networks trained with the KITTI dataset are used to train the whole network with the Cityscape dataset.
%Let us assume to train the proposed networks using 20,000 stereo images and sparse depth map of $3\%$ density from the KITTI dataset \cite{}.

%The adaptive threshold $\tau$ inferred by the threshold network helps the monocular depth network to use more reliable depth values only.
%This is also beneficial to improving the prediction accuracy of the confidence estimation approach itself.

\section{Extension to Confidence Estimation}
The soft-thresholding attenuates low confidence values that are less than $\tau$ to become as close as 0 while amplifying high confidence values to converge to 1.
It reduces the number of ambiguous pixels to determine the reliability, for which a confidence value is far from 0 or 1.
%This operation improves the prediction accuracy of existing confidence estimation approaches.
We discuss how the soft-thresholding based on the threshold network can improve the prediction accuracy of existing confidence estimation approaches~\cite{poggi2016learning, kim2019laf}.
%Fig. \ref{fig:conf_thres} shows the network architecture for confidence estimation boosted by the soft-thresholding.
In the ThresNet of \figref{Overall_method}, the confidence network can be replaced with the existing confidence estimation approaches.
%and threshold network are connected sequentially, and the encoder of the monocular depth network is used so that the color image features are concatenated to the threshold network.
One difference is that the loss $L_T$ (cross-entropy loss) is measured on the disparity domain, considering that the existing confidence estimation approaches are trained on the disparity domain.
This formulation is model-agnostic, and any kind of existing confidence estimation approaches can be used in a plug-and-play fashion.
%We verify how much the confidence prediction accuracy of existing confidence estimation approaches is improved by using the soft-thresholding based on the threshold network.

\section{Experimental results}

\subsection{Implementation details}
The proposed method was implemented in PyTorch framework and run Titan RTX GPU. We trained the whole networks on the learning rate of $10^{-4}$ and batches of 32 images resized to $192 \times 480$ for 30 epochs.
%In our experiments, $\varepsilon$ for soft-thresholding was set to $10$ that deduces the best result.
We trained the proposed monocular depth estimation network consisting of DepthNet and RefineNet on the standard 20k stereo images provided in the KITTI dataset.
%The sparse LiDAR depth maps of $3\%$ density of the KITTI dataset was used to train the ThresNet.
We evaluate our methods on following five metrics `RMSE', `RMSE log', `Abs Rel', `Sq. Rel', and `Accuracy', proposed in Eigen \textit{et al.}~\cite{eigen2014depth}.

%We trained the depth estimation module on the standard 20000 KITTI training dataset[KITTI .], and trained the confidence estimation module on the 20 KITTI 2012 training dataset. We also trained our model on the Cityscapes dataset to assess how well our methods work on the datasets other than KITTI. For the standard depth evaluation, we utilized the Eigen split [Eigen et al.] with maximum depth setting as 80 meters. We exploited the raw LiDAR data using Gargs crop[Garg et al.] for the test ground truth for depth. In the evaluation for confidence measure module, we tested on the KITTI 2015 dataset using a census transform with a 5x5 local window and MC-CNN[MCCNN et al.], just like the configuration in (Lafnet et al.). Using the structures of CCNN[et al], LAFNet[et al] and ConfNet[et al], we compared the performance of our methods attached to those structures with the performance of the unattached one.
%To evaluate the performance of depth estimation, we leveraged a commonly accepted evaluation method proposed in [Depth map prediction from a single image using a multi-scale deep network] with 5 evaluation indicators : RMSE, RMSE log, Abs Rel, Sq Rel, and Accuracies.  To evaluate the performance of confidence estimation, we used the sparsification curve and its area under curve (AUC) which is used in [laf paper 8, 39, 27, 37, 21, lafnet]. The AUC is used to evaluate the ability of the confidence measure to differentiate the accurate disparity from the inaccurate with regard to the optimal answer.

\subsection{Evaluation on monocular depth estimation}
\subsubsection{KITTI}
In \tabref{Quantative_KITTI}, we evaluated the monocular depth estimation performance quantitatively on the KITTI Eigen Split \cite{eigen2014depth} dataset with setting maximum depth to 80 meters with Gargs crop \cite{garg2016unsupervised}.
A comprehensive evaluation was conducted with Monodepth~\cite{godard2017unsupervised}, Uncertainty~\cite{poggi2020uncertainty}, MonoResMatch~\cite{tosi2019learning}, Monodepth2~\cite{godard2019digging}, DepthHint~\cite{watson2019self}, PackNet-SfM~\cite{guizilini20203d}, and Insta-DM~\cite{lee2021learning}.
For the training data, `S' indicates using stereo images for self-supervised monocular depth estimation.
%`LL' refers to a left image and LiDAR ground truth depth map, while `SL' refers to stereo images and LiDAR ground truth depth map.
`M' represents a monocular video sequence.
%`LP' refers to a left image and pseudo depth map used in our method.
The evaluation of the proposed method is twofold; `Ours (D)' trained with only the DepthNet using $L_D$ in \eqref{eq:depth_loss_LD} without refining the depth map, and `Ours (D+R)' trained with the DepthNet and RefineNet.

%As reported in \tabref{Quantative_KITTI}, the superior performance of `Ours (D)' over existing methods demonstrates the effectiveness of the proposed threshold learning approach.
As reported in \tabref{Quantative_KITTI}, although `Ours (D)' leverages a rather simple encoder-decoder architecture, it achieves the superior performance over existing methods, demonstrating the effectiveness of the proposed threshold learning approach.
In `Ours (D+R)', the monocular depth accuracy was further improved by making use of the probabilistic refinement module based on the uncertainty map and the PAC layer in the RefineNet. We also evaluated the number of parameters used and an inference time, noted as `\#p' and `time', respectively. Our method uses relatively smaller or similar number of parameters compared to other methods. % More details about the parameters used and the inference time are provided in supplementary materials.
`Ours$^{\dagger}$' is obtained using the self-supervised ThresNet~\cite{poggi2020self}, while `Ours' indicates the results obtained using the supervised ThresNet. We found that their monocular depth accuracy is almost similar. The following results including ablation study were conducted with the supervised ThresNet.
%We noticed that whether ThresNet is trained with self-supervised~\cite{poggi2020self} or supervised manner~\cite{poggi2016learning} make only marginal difference.} %there is no significant difference between the training of ThresNet using self-supervised method \cite{poggi2020self} and the supervised manner \cite{poggi2016learning}.
\figref{Depth_Qual_KITTI} shows the qualitative comparison with existing methods on the KITTI Eigen Split \cite{eigen2014depth} dataset.
It was shown that the proposed method recovers complete instances better while preserving fine object boundaries.

%\subsubsection{Cityscapes}
%We also evaluated the performance of the proposed method on the Cityscapes dataset. The Cityscapes dataset provides only stereo images without the sparse LiDAR depth maps, and thus the ThresNet trained with the KITTI dataset was used to infer the threshold.
%\tabref{Quantitative_city} provides quantitative evaluations on the Cityscapes validation dataset,
%setting maximum depth to 80 meters with the per-image median scaling approach \cite{zhou2017unsupervised}.
%We used the SGM depth~\cite{hirschmuller2005accurate} as ground truth for the evaluation.
%The performance evaluation includes Monodepth \cite{godard2017unsupervised}, Kuznietsov \textit{et al.}~\cite{kuznietsov2017semi}, MonoResMatch \cite{tosi2019learning}, Monodepth2 \cite{godard2019digging}, DepthHint \cite{watson2019self}, PackNet-SfM \cite{guizilini20203d}.
%\figref{Qualitative_city} shows the qualitative comparison with various methods.
%Though the pretrained models were used for the confidence and threshold networks, our method still outperforms state-of-the-arts approaches.
%\hspace{-0.4cm}

\subsection{Cityscapes}
We also evaluated the performance of the proposed method on the Cityscapes dataset. The Cityscapes dataset provides only stereo images without the ground truth, and thus the ThresNet trained with the KITTI dataset was used to infer the threshold.
\tabref{Quantitative_city} shows the quantitative evaluation on Cityscapes dataset \cite{cordts2016cityscapes} with the DepthNet and RefineNet fine-tuned on the Cityscapes dataset, while the ThresNet is freezed. We compared our results with Monodepth2 \cite{godard2019digging}, DepthHint \cite{watson2019self} and PackNet-SfM \cite{guizilini20203d}.
We set maximum depth to 80 meters with the per-image median scaling approach \cite{zhou2017unsupervised}. We used the SGM depth~\cite{hirschmuller2005accurate} as ground truth for the evaluation.
%For the training data, `M' indicates using a monocular video sequence, and `S' indicates using stereo images.
%`LP' refers to a left image and pseudo depth map used in our method.
The outstanding performance of our method supports the claim that the ThresNet trained with the KITTI dataset shows a satisfactory generalization capability for different datasets.

\begin{table*}[]
	\centering
	\small\addtolength{\tabcolsep}{-1pt}
	\caption{Quantitative evaluation for monocular depth estimation results on Cityscapes validation dataset with fine-tuning on Cityscapes training dataset. Numbers in bold and underlined represent $1^{st}$ and $2^{nd}$ ranking, respectively.}
	\begin{tabular}{@{}ccccccccccc@{}}
		\toprule
		&                                    & \multicolumn{4}{c}{Lower is better}                       & \multicolumn{3}{c}{Accuracy: higher is better}    \\ \midrule
		\multicolumn{1}{c|}{Method} & \multicolumn{1}{c|}{Data} & Abs Rel & Sqr Rel & RMSE  & \multicolumn{1}{c|}{RMSE log} & $\delta<1.25$ & $\delta<1.25^2$ & $\delta<1.25^3$ \\ \midrule
		%PackNet-SfM \cite{guizilini20203d}      & M    & 0   & 0   & 0 & 0   & 0    & 0   & 0  \\
		Monodepth2 \cite{godard2019digging}     & S    & 0.124   & 1.287   & 7.293 & 0.223   & 0.785    & 0.947   & 0.981  \\
		Struct2Depth \cite{casser2019unsupervised} & M     & 0.145   & 1.737   & 7.280   & 0.205  & 0.813  & 0.942   & 0.978  \\
		DepthHint \cite{watson2019self}         & S    & 0.128   & 1.268   & 7.156 & 0.218   & 0.812    & 0.949   & 0.982  \\
		Gordon \cite{gordon2019depth} & M     & 0.127   & 1.330   & 6.960   & \textbf{0.195}  & 0.830  & 0.947   & 0.981  \\
		Ours (D)                & S     & \underline{0.123}   & \underline{1.141}   & \underline{6.735}   & \underline{0.204}  & \underline{0.844}  & \underline{0.962}   & \underline{0.985}  \\
		Ours (D+R)                & S     & \textbf{0.115}   & \textbf{1.125}   & \textbf{6.584}   & \textbf{0.195}  & \textbf{0.857}  & \textbf{0.963}   & \textbf{0.986}  \\ \bottomrule
	\end{tabular}
	\label{Quantitative_city}
	\vspace{-0.3cm}
\end{table*}
\begin{figure*}[bt!]
	\centering
	\begin{subfigure}[]
		{\includegraphics[width=0.16\textwidth]{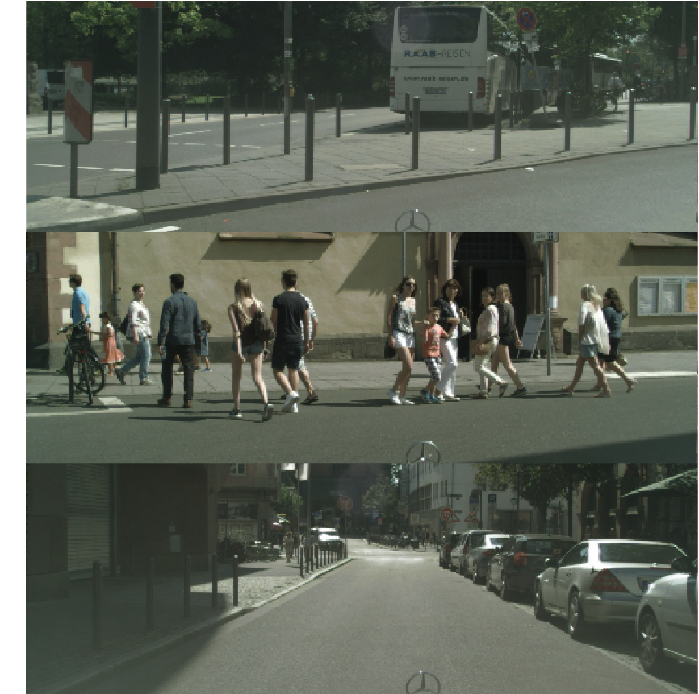}}
	\end{subfigure}
	\hspace{-0.3cm}
	\begin{subfigure}[]
		{\includegraphics[width=0.16\textwidth]{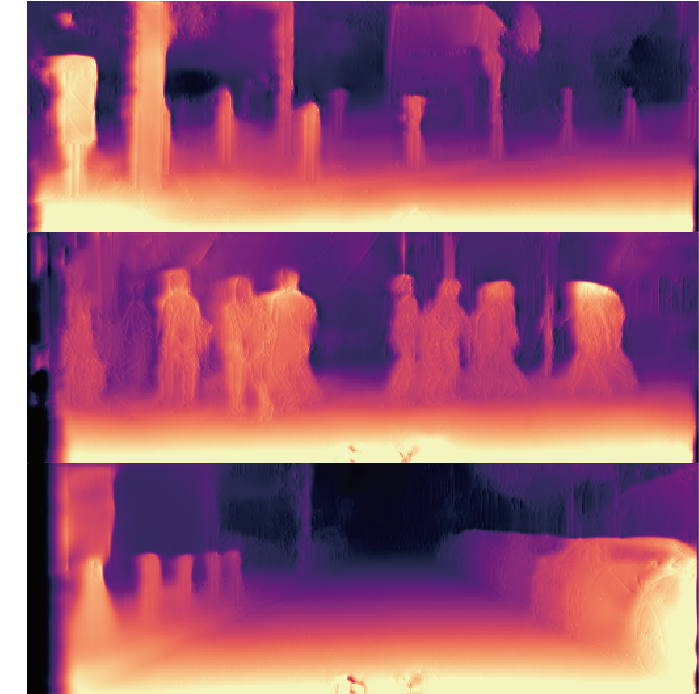}}
	\end{subfigure}
	\hspace{-0.3cm}
	\begin{subfigure}[]
		{\includegraphics[width=0.16\textwidth]{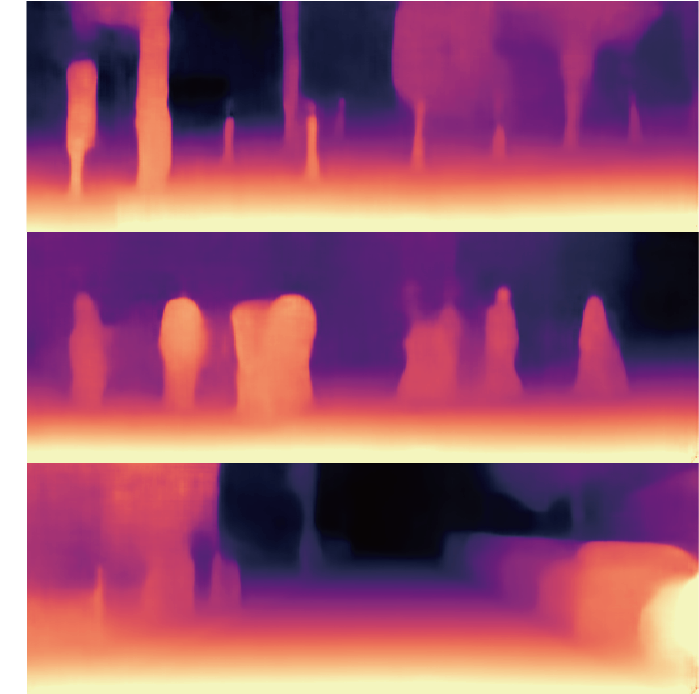}}
	\end{subfigure}
	\hspace{-0.3cm}
	\begin{subfigure}[]
		{\includegraphics[width=0.16\textwidth]{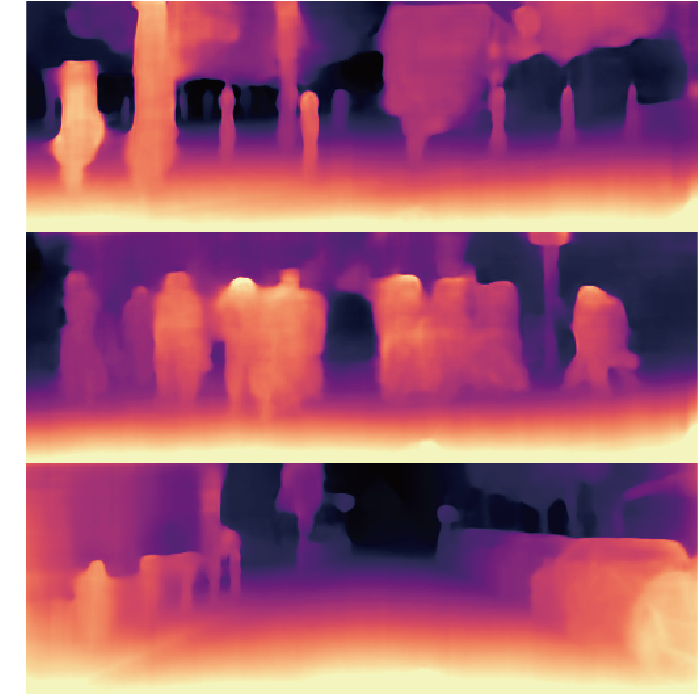}}
	\end{subfigure}
	\hspace{-0.2cm}
	\begin{subfigure}[]
		{\includegraphics[width=0.1543\textwidth]{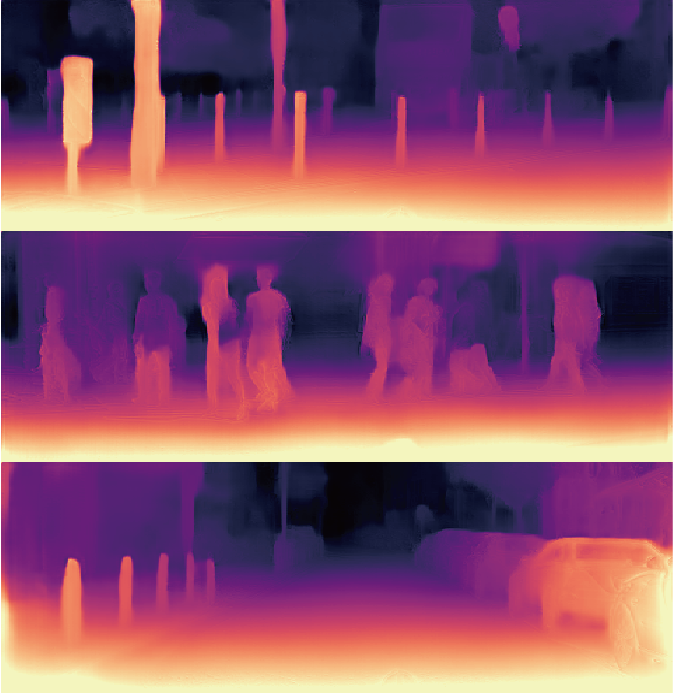}}
	\end{subfigure}
	\hspace{-0.3cm}
	\begin{subfigure}[]
		{\includegraphics[width=0.16\textwidth]{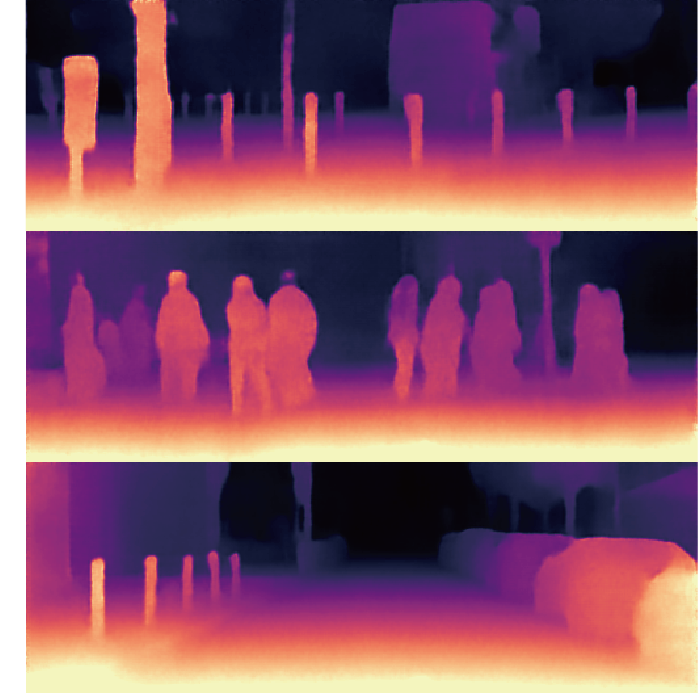}}
	\end{subfigure}
	\caption{Qualitative evaluation for depth estimation with existing methods on Cityscapes validation dataset: (a) Input image, (b) Monodepth \cite{godard2017unsupervised}, (c) MonoResMatch \cite{tosi2019learning}, (d) DepthHint \cite{watson2019self}, (e) PackNet-SfM \cite{guizilini20203d} (f) Ours (D+R).}
	\label{Qualitative_city}
	\vspace{-0.3cm}
\end{figure*}

\subsection{Evaluation on uncertainty estimation}
To evaluate the performance of the uncertainty measure, we use sparsification plots used in \cite{ilg2018uncertainty}. {\color{red}`AUSE'} denotes the Area Under the Sparsification Error which quantifies how close the estimate is to the oracle uncertainty, which is {\color{red}lower} the better.
{\color{blue}`AURG'} denotes the Area Under the Random Gain, which indicates how better it is compared to the case without modelling, which is {\color{blue}higher} the better. In \tabref{Uncertainty_measure}, the uncertainty measure estimated by the proposed method was compared with `Monodepth2-Log' of Poggi \textit{et al.}~\cite{poggi2020uncertainty}, trained under the same setup as our experiments.

\subsection{Ablation study}
\noindent {\bf{Threshold learning}}
In \tabref{table:ablation_method1}, we conducted the ablation study to validate the performance improvement by the proposed threshold learning over existing threholding approaches \cite{cho2019large, tonioni2019unsupervised}.
For a fair comparison, we obtained the results using the monocular depth network trained with only the DepthNet (without the uncertainty decoder), when varing thresholding functions.
`Baseline' represents the results obtained using the confidence map without thresholding.
The results of \cite{cho2019large} were obtained using the hard thresholding of \figref{slope_maskeq} (a) with $\tau=0.3$, following the setup of \cite{cho2019large}.
The performance of \cite{cho2019large,tonioni2019unsupervised} was almost similar, though the method in \cite{tonioni2019unsupervised} learned the threshold $\tau$ with the thresholding function of \figref{slope_maskeq} (b).
We found that the regularization loss $-log(1-\tau)$ \cite{tonioni2019unsupervised}, used to prevent the threshold $\tau$ from approaching 1, does not generate a meaningful variant for the learned threshold due to the lack of explicit supervision for the threshold learning. %, thus producing similar results to the method \cite{cho2019large} of using a fixed threshold.
`Tonioni et al. \cite{tonioni2019unsupervised} + $L_T$' were obtained using the thresholding function of \figref{slope_maskeq} (b) and our loss $L_T$.
The performance gain over `Tonioni et al. \cite{tonioni2019unsupervised}' demonstrates the effectiveness of $L_T$.
`Ours (D)' achieves a substantial performance gain, demonstrating the effectiveness of the proposed threshold learning with $L_T$.

% \usepackage{booktabs}
\begin{table}[t]
    \centering
    \small\addtolength{\tabcolsep}{-3.5pt}
    \caption{Quantitative evaluation for uncertainty estimation with the state-of-the-art method on KITTI Eigen Split \cite{eigen2014depth} dataset. Numbers in bold indicate the better performance.}
    \begin{tabular}{@{}c|cc|cc|cc@{}}
        \toprule
        & \multicolumn{2}{c|}{Abs Rel}                                                                     & \multicolumn{2}{c|}{RMSE}                                                                        & \multicolumn{2}{c}{$\delta \geq 1.25$}                                                           \\ \midrule
        Method      & \multicolumn{1}{c|}{{\color{red}AUSE}} & {\color{blue}AURG} & \multicolumn{1}{c|}{{\color{red}AUSE}} & {\color{blue}AURG} & \multicolumn{1}{c|}{{\color{red}AUSE}} & {\color{blue}AURG} \\ \midrule
        Uncertainty~\cite{poggi2020uncertainty} & 0.022                                                     & 0.036                                & 0.938                                                     & 2.402                                & \textbf{0.018}                               & 0.061                                \\
        Ours        & \textbf{0.021}                                                     & \textbf{0.048}                                & \textbf{0.765}                                                     & \textbf{2.881}                                & 0.025                               & \textbf{0.080}                                \\ \bottomrule
    \end{tabular}
    \label{Uncertainty_measure}
    \vspace{-0.3cm}
\end{table}

\noindent{\bf{Adaptability}}
We also validated the effectiveness of our method when applied to different network architectures, e.g., PackNet~\cite{guizilini20203d}.
\tabref{table:ablation_method2} shows quantitative evaluation results when using our confidence threshold learning and probabilistic refinement on the PackNet architecture.
`PackNet (D)' represents the results obtained using the DepthNet only, whereas `PackNet (D+R)' is the results using both DepthNet and RefineNet.
We observed that our framework also improves the monocular depth accuracy for the PackNet architecture.

\noindent{\bf{Uncertainty}}
To evaluate the importance of using the estimated uncertainty in the RefineNet, we compared the results obtained using the proposed depth refinement of \eqref{eq:depth_PAC} and the simple depth refinement ($d^f=d+d'$) without $\sigma$ in \tabref{table:ablation_method3},
demonstrating the effectiveness of the depth refinement based on the uncertainty map.

\noindent{\bf{Pseudo ground truth depth labels}}
So far, all experiments were conducted with the self-supervised pseudo depth maps obtained using \cite{watson2020learning}.
To validate the adaptability of our framework with respect to the pseudo depth labels, we performed additional experiments with the pseudo ground truth depth maps generated by \cite{tonioni2019learning}, which are trained with synthetic data and fine-tuned with an self-supervised reconstruction loss with meta-learning framework.
\tabref{table:ablation_method4} shows that the monocular depth accuracy is still superior to state-of-the-arts monocular depth estimation approaches. %when using the pseudo depth labels of \cite{l2a}
%Also, recently, it was shown that several domain-invariant stereo matching approaches \cite{tonioni2019learning,zhang2020domain,cai2020matching}, trained with synthetic dataset only, achieve comparable performance on real scenes.
%Our framework is capable of effectively leveraging stereo knowledge distilled by self-supervised stereo model \cite{watson2020learning} or domain-invariant stereo model \cite{tonioni2019learning,zhang2020domain,cai2020matching}, meaning that advances in these stereo models lead to the improvement of the monocular depth accuracy.
%different from conventional monocular depth estimation approaches \cite{godard2017unsupervised,godard2019digging,guizilini20203d}.

%[aa1] Domain-invariant Stereo Matching Networks, ECCV 2020
%[aa2] Matching-space Stereo Networks for Cross-domain Generalization, 3DV 2020

%Since our framework is effective at excluding unconfident pixels from training, it always works well regardless of which pseudo ground truth is used.
%Currently, our method uses \cite{watson2020learning} as the pseudo ground truth for all the experiments. However, since our method is effective at excluding unconfident pixels from training, it always works well regardless of which pseudo ground truth is used.
%\tabref{table:ablation_method4} shows the quantitative evaluation of the results obtained by the proposed method using L2A~\cite{l2a} as a pseudo ground truth on the KITTI Eigen Split~\cite{eigen2014depth} dataset.

\begin{table}[t]
    \centering
    \small\addtolength{\tabcolsep}{-4pt}
    \caption{Comparison with other thresholding methods on the KITTI Eigen Split~\cite{eigen2014depth} dataset. We evaluated the performance with the supervised ThresNet, and $L_T$ is a cross-entropy loss.}% All settings are trained with the pseudo ground truth labels in our proposed method, while the methods of masking the confidence map are different, which can be easily understood by the Figure \ref{slope_maskeq}. The reason for the slight difference in performance between Cho\cite{cho2019large} and Tonioni\cite{tonioni2019unsupervised} is due to the $\tau$ regularization loss used in Tonioni\textit{et al.}~\cite{tonioni2019unsupervised}. The regularization loss used in Tonioni\textit{et al.}~\cite{tonioni2019unsupervised}, which makes $\tau$ not converge to 1, does not cause a big variant over the value of $\tau$ between images, which makes it similar to the method of fixed $\tau$.} %abs and rms represent 'Abs Rel', and 'RMSE', respectively. in depth estimation performance according to our proposed methods}
    \begin{tabular}{@{}c|c|cccc@{}}
        \toprule
        & {$\tau$} & {Abs} & {RMSE} & {$\delta<1.25$}\\ \midrule
        Baseline & $\times$ & 0.108        & 4.552        & 0.869                  \\
        Cho et al. \cite{cho2019large} &fixed & 0.102        & 4.441        & 0.874                  \\
        Tonioni et al. \cite{tonioni2019unsupervised} &learned & 0.101        & 4.453        & 0.878                  \\
        Tonioni et al. \cite{tonioni2019unsupervised} + $L_T$ &learned & 0.100        & 4.390        & 0.879                 \\
        Ours (D) &learned & 0.099        & 4.266        & 0.883                  \\ \bottomrule
    \end{tabular}
    \label{table:ablation_method1}
    \vspace{-0.2cm}
\end{table}

\begin{table}[t]
    \centering
    \small\addtolength{\tabcolsep}{-4pt}
    \caption{Quantitative evaluation of the results obtained by applying our threshold learning and probabilistic refinement to the PackNet-SfM architecture \cite{guizilini20203d} on the KITTI Eigen Split \cite{eigen2014depth} dataset.}
    %abs and rms represent 'Abs Rel', and 'RMSE', respectively. in depth estimation performance according to our proposed methods}
    \begin{tabular}{@{}c|cc|ccc@{}}
        \toprule
        & {Abs} & {RMSE} & {$\delta<1.25$} \\ \midrule
        PackNet-SfM \cite{guizilini20203d}   & 0.111        & 4.601        & 0.878\\
        PackNet-SfM (D)   & 0.105        & 4.258        & 0.880\\
        PackNet-SfM (D+R) & 0.100        & 4.225        & 0.883\\  \bottomrule
    \end{tabular}
    \label{table:ablation_method2}
    \vspace{-0.2cm}
\end{table}

\begin{table}[t]
    \centering
    \small\addtolength{\tabcolsep}{-4pt}
    \caption{Ablation study of the uncertainty map.}
    \begin{tabular}{@{}c|cc|ccc@{}}
        \toprule
        & {Abs} & {Sqr} & {RMSE} & {RMSE log} & {$\delta<1.25$} \\ \midrule
        \eqref{eq:depth_PAC} w/o $\sigma$  & 0.099   & 0.661     & 4.298    & 0.188    & 0.881\\
        \eqref{eq:depth_PAC}  & 0.096   & 0.627     & 4.201    & 0.186    & 0.885\\  \bottomrule
    \end{tabular}
    \label{table:ablation_method3}
    \vspace{-0.2cm}
\end{table}

\begin{table}[t]
    \centering
    \small\addtolength{\tabcolsep}{-4pt}
    \caption{Quantitative evaluation when using pseudo depth labels generated by \cite{tonioni2019learning} on the KITTI Eigen Split~\cite{eigen2014depth} dataset.}
    %abs and rms represent 'Abs Rel', and 'RMSE', respectively. in depth estimation performance according to our proposed methods}
    \begin{tabular}{@{}c|cc|ccc@{}}
        \toprule
        & {Abs} & {Sqr} & {RMSE} & {RMSE log} & {$\delta<1.25$} \\ \midrule
        Ours (D)   & 0.102        & 0.728        & 4.281         & 0.189        & 0.880\\
        Ours (D+R) & 0.100        & 0.711        & 4.230         & 0.187        & 0.883\\  \bottomrule
    \end{tabular}
    \label{table:ablation_method4}
    \vspace{-0.2cm}
\end{table}

\subsection{Confidence evaluation}
We validated the effectiveness of the proposed threshold learning in terms of confidence prediction accuracy by applying it to two confidence estimation approaches, CCNN~\cite{poggi2016learning} and LAFNet~\cite{kim2019laf}.
We trained the two confidence estimation methods with 20 out of 194 images provided in the KITTI 2012 training dataset~\cite{geiger2013vision}. Note that the confidence estimation approaches~\cite{poggi2016learning,kim2019laf} are evaluated by training them in a supervised manner.
The area under the curve (AUC) \cite{hu2012quantitative}, which is a common metric for confidence estimation approaches, was used for an objective evaluation.
%The lower AUC value implies that the confidence map is more capable of distinguishing correct matches from incorrect ones.
%Due to page limits, some results are provided in supplementary materials.
Refer to the supplementary material for details on measuring AUC and optimal AUC and more results.
%We trained the two confidence estimation methods with 20 out of 194 images provided in KITTI 2012 training dataset \cite{geiger2013vision}.
Following confidence estimation literatures, input disparity maps used for predicting the confidence maps were obtained using two popular stereo algorithms, `Census-SGM' \cite{hirschmuller2005accurate} and `MC-CNN' \cite{vzbontar2016stereo}.

\begin{figure}[t!]
    \centering
    \begin{subfigure}[color image]
        {\includegraphics[width=0.319\columnwidth]{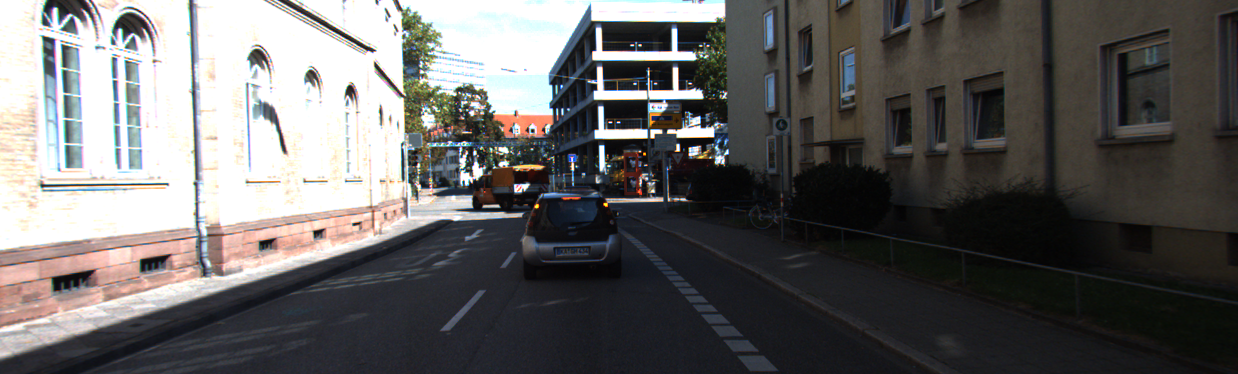}}
    \end{subfigure}
    \begin{subfigure}[CCNN]
        {\includegraphics[width=0.319\columnwidth]{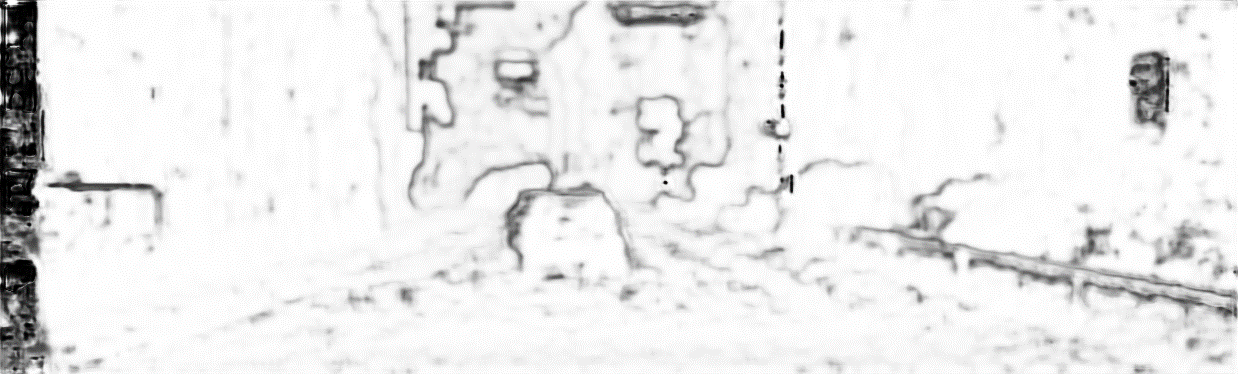}}
    \end{subfigure}
    \begin{subfigure}[LAFNet]
        {\includegraphics[width=0.319\columnwidth]{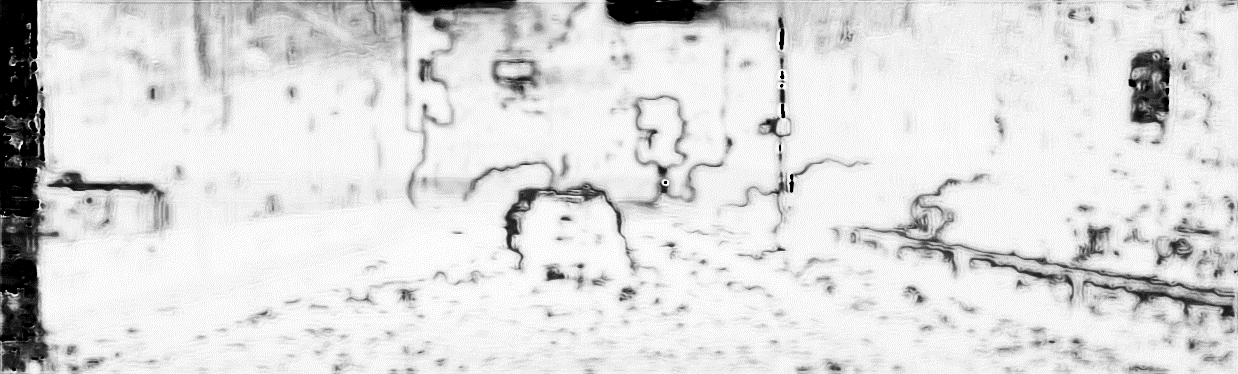}}
    \end{subfigure}
    \begin{subfigure}[input disparity]
        {\includegraphics[width=0.319\columnwidth]{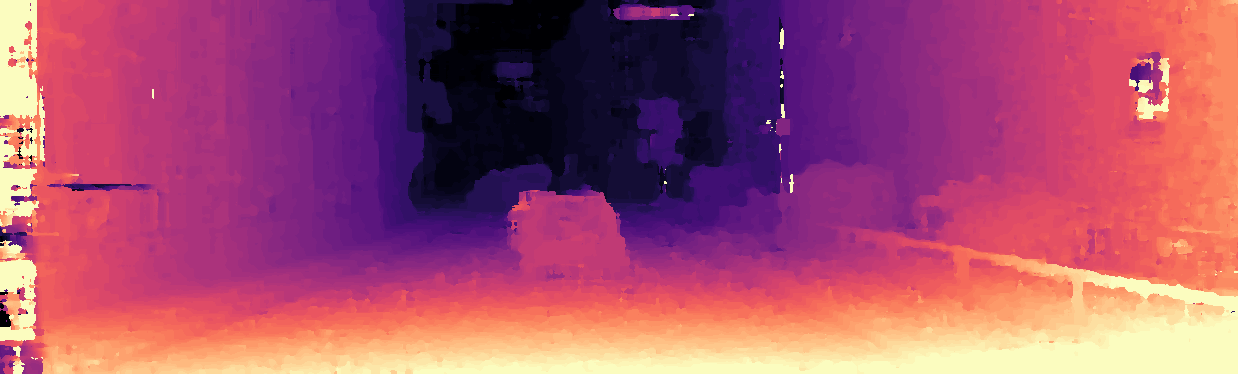}}
    \end{subfigure}
    \begin{subfigure}[CCNN w/ $\tau$]
        {\includegraphics[width=0.319\columnwidth]{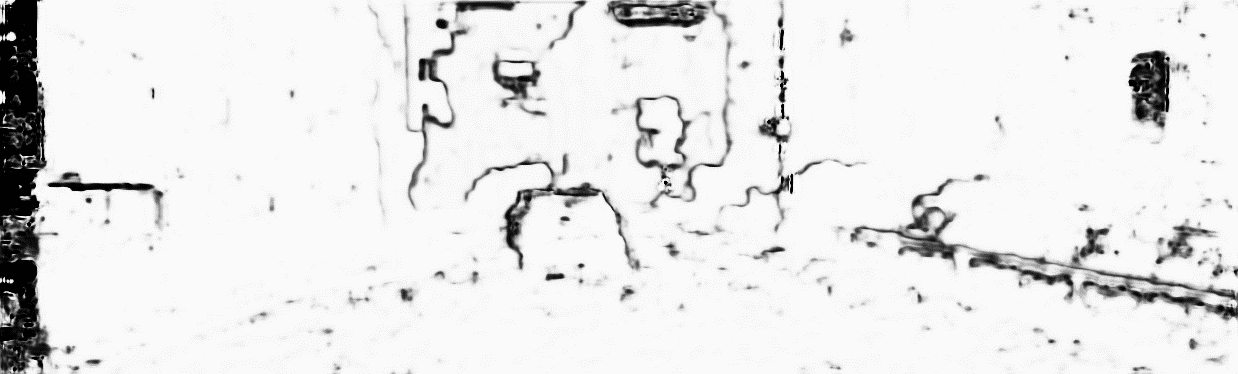}}
    \end{subfigure}
    \begin{subfigure}[LAFNet w/ $\tau$]
        {\includegraphics[width=0.319\columnwidth]{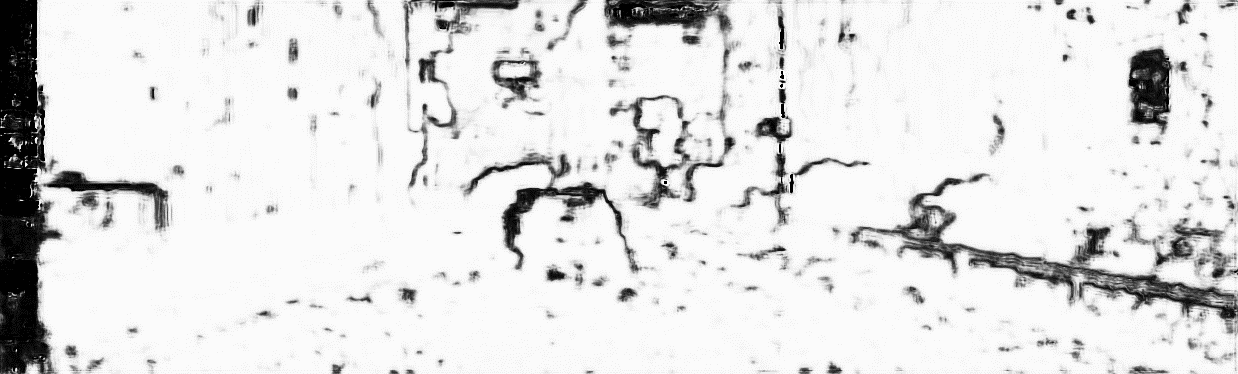}}
    \end{subfigure}
    \caption{Qualitative results of confidence map on KITTI 2015 dataset using census-SGM.}
    \label{conf_qualitative}
    \vspace{-0.4cm}
\end{figure}

%Our networks in Figure \ref{fig:conf_thres} were fine-tuned with the threshold network $\texttt{M}_{T}$ and the encoder of the monocular depth network $\texttt{M}_{D}$.
\tabref{table_conf} shows objective evaluation results for 200 images of KITTI 2015 dataset \cite{menze2015object} and 15 images of Middlebury v3 dataset \cite{scharstein2014high}.
`w/$\tau$' denotes our results using the soft-thresholding technique.
LAFNet* denotes the LAFNet \cite{kim2019laf} in which 3D cost volume is not used as an input.
Our approach consistently outperforms the original confidence estimation methods, demonstrating the effectiveness of the proposed threshold learning.
\figref{conf_qualitative} compares the confidence maps visually.
While the original confidence maps contain ambiguous values for which it is difficult to determine whether the depth label is correct, our thresholded confidence map yields more distinct values that are close to 0 or 1.
Such a binarization enables the estimated confidence to have similar distribution to ground truth confidence, thus improving a discriminative power.

%%%%%%%%%%%%%%%%%%%%%%%%%%%%%%%%%%%%%%%%%%%%%%%%%%%%%%%%%%%%%%%%%%%%%%%%%%%%%%%%%%%%%%%%%%%%%%%%%%%%%%%%%%%%%%%%%%%%%%%%%%%%%%%%%%%%%%%%%%%%%%%%%%%%%%%%%%%%%%%%%%%%%%%%%%%%%%%%%%%%%%%%%%%%%%%%%%%
%DB: Move this to supp.
%To measure this, ROC curve is figured out at initial by sorting disparity pixels following decreasing order of confidence and iteratively substituting the low confident disparity pixel with the ground-truth disparity %sampling subset of them
%for computing the error rate indicating the percentage of pixels with a difference greater than $\rho$ from ground-truth disparity.
%The optimal AUC is computed according to the fact that the error rate $\varepsilon$ is ideally 0 when sampling the first $(1 - \varepsilon)$ pixels and is equal to
%\begin{equation}
%    AUC_{opt}=\int_{1-\varepsilon}^{1}{\frac{x-(1-\varepsilon)}{x}dx}=\varepsilon+(1-\varepsilon)\ln{1-\varepsilon}.
%\end{equation}
%Also, $\rho$ is set to 3 for KITTI and 1 for Middlebury.
%%%%%%%%%%%%%%%%%%%%%%%%%%%%%%%%%%%%%%%%%%%%%%%%%%%%%%%%%%%%%%%%%%%%%%%%%%%%%%%%%%%%%%%%%%%%%%%%%%%%%%%%%%%%%%%%%%%%%%%%%%%%%%%%%%%%%%%%%%%%%%%%%%%%%%%%%%%%%%%%%%%%%%%%%%%%%%%%%%%%%%%%%%%%%%%%%%%

\begin{table}[]
    \centering
    \small\addtolength{\tabcolsep}{-4pt}
    \caption{Performance evaluation of confidence estimation for KITTI 2015 and Middlebury v3 datasets with two popular stereo matching methods C-SGM (Census-SGM) \cite{hirschmuller2005accurate} and MC-CNN \cite{vzbontar2016stereo}.
    AUC values are reported and the lower is the better.}
    \begin{tabular}{@{}lcc@{}}
        \toprule
        \multicolumn{1}{c}{\multirow{2}{*}{}} & KITTI 2015    & MID 2014      \\ \cmidrule(l){2-3}
        \multicolumn{1}{c}{}                  & C-SGM / MC-CNN & C-SGM / MC-CNN \\ \toprule
        \bottomrule
        CCNN                                  & 1.868 / 3.190 & 9.486 / 9.787 \\
        CCNN w/$\tau$                        & 1.720 / 3.525 & 8.314 / 9.497 \\
        LAFNet*                               & 1.797 / 3.051 & 8.895 / 9.660 \\
        LAFNet* w/$\tau$                     & 1.687 / 3.037 & 8.988 / 9.456 \\
        LAFNet                                & 1.680 / 2.903 & 8.884 / 9.305 \\
        LAFNet w/$\tau$                      & \textbf{1.587} / \textbf{2.885} & \textbf{8.680} / \textbf{8.622} \\ \toprule
        \midrule
        optimal                               & 0.737 / 2.761 & 3.887 / 4.985 \\ \bottomrule
    \end{tabular}
    \label{table_conf}
    \vspace{-0.3cm}
\end{table}

\section{Conclusion}
In this work, we have proposed a novel framework for monocular depth estimation based on pseudo depth labels generated by self-supervised stereo matching methods.
%Instead of relying on the self-supervision using the reconstruction loss across stereo images, we proposed to use pseudo depth labels .
The confidence map is used to exclude erroneous depth values within the pseudo depth labels.
The prediction errors in the confidence map are further suppressed by making use of the soft-thresholding based on threshold learning.
Furthermore, the probabilistic refinement module enables improving the monocular depth accuracy with the help of the uncertainty map.
The proposed framework has shown impressive performances over state-of-the-arts on several popular datasets.
It was also shown that threshold learning can also boost the prediction accuracy of existing confidence approaches.
%As future work, we will investigate the possibility of applying the soft-thresholding to various tasks based on binary prediction.
% \usepackage{booktabs}
%\begin{table}[h]
%    \centering
%    \caption{{Monocular depth accuracy according to $\varepsilon$}}
%    \begin{tabular}{@{}c|cc|ccc@{}}
%        \toprule
%        $\varepsilon$  & \textbf{abs}   & \textbf{rms}   & \textbf{$\delta<1.25$} & \textbf{$\delta<1.25^2$} & \textbf{$\delta<1.25^3$} \\ \midrule
%        \textbf{10} & \textbf{0.098} & \textbf{4.253} & 0.884                  & \textbf{0.960}           & \textbf{0.981}                    \\
%        \textbf{30} & 0.101          & 4.283          & 0.882                  & 0.959                    & 0.980                    \\
%        \textbf{50} & 0.100          & 4.265          & 0.882                  & 0.959                    & 0.980                    \\
%        \textbf{70} & 0.100          & 4.272          & \textbf{0.885}         & \textbf{0.960}           & 0.980                    \\
%        \textbf{90} & 0.100          & 4.255          & 0.884                  & \textbf{0.960}           & \textbf{0.981}                    \\ %\bottomrule
%    \end{tabular}
%    \label{table:ablation_eps}
%\end{table}

\bigskip

{\small
\bibliographystyle{ieee_fullname}
\bibliography{egbib}
}

\end{document}

	%%%%%%%%% TITLE
	\renewcommand\footnotemark{}
	\title{Adaptive confidence thresholding for monocular depth estimation\\- Supplementary material}
	
	\author{
		%Authors
		% All authors must be in the same font size and format.
		Hyesong Choi\textsuperscript{\rm 1}$^{*}$\thanks{$^{*}$ Equal contribution. $^{\dagger}$ Corresponding author.}, Hunsang Lee\textsuperscript{\rm 2}$^{*}$, Sunkyung Kim\textsuperscript{\rm 1}, Sunok Kim\textsuperscript{\rm 3},\\Seungryong Kim\textsuperscript{\rm 4}, Kwanghoon Sohn\textsuperscript{\rm 2}, Dongbo Min\textsuperscript{\rm 1}$^{\dagger}$\\
		\textsuperscript{\rm 1}Ewha W. University,
		\textsuperscript{\rm 2}Yonsei University,
		\textsuperscript{\rm 3}Korea Aerospace University,
		\textsuperscript{\rm 4}Korea University\\
		%\tt\small $\{$hyesongchoi2010,skk0951324$\}$@gmail.com,
		%\tt\small $\{$hslee91,khsohn$\}$@yonsei.ac.kr,\\
		%\tt\small sunok.kim@kau.ac.kr,
		%\tt\small seungryong$\_$kim@korea.ac.kr,
		%\tt\small dbmin@ewha.ac.kr
		\\
	}
	
	\maketitle
	\ificcvfinal\thispagestyle{empty}\fi

\hspace{-0.1cm}
%%%%%%%%% BODY TEXT
In this document, we provide more comprehensive results not provided in the original manuscript due to the page limit as below.
The code to reproduce our results will be publicly available soon. Note that all experiments were conducted with the supervised ThresNet.
\begin{itemize}
    \item Histogram of the learned threshold $\tau$ and qualitative evaluation of depth results computed using different thresholding methods (Section 1)
    \item Qualitative evaluation for monocular depth estimation with state-of-the-arts on KITTI and Cityscapes datasets (Section 2.2 and 2.3)
    \item Quantitative evaluation for monocular depth estimation on Cityscape dataset without fine-tuning (Section 2.4)   
    \item Quantitative evaluation for monocular depth estimation using improved ground truth depth maps \cite{uhrig2017sparsity} on KITTI dataset (Section 2.5)
    \item Performance analysis according to a hyperparameter $\varepsilon$ used in the soft-thresholding function (Section 2.6)
    \item Quantitative evaluation of the proposed method according to the use of DepthNet and RefineNet (Section 2.7)
    %\item More ablation study on the threshold learning (Section 1.8)%Quantitative evaluation of the confidence masking method (Section 1.8)
    \item Ground truth confidence map and evaluation metric used in the confidence estimation (Section 3.1 and 3.2)
    \item Qualitative result for confidence estimation with state-of-the-arts on KITTI dataset (Section 3.3)
    \item Evaluation metric used in the uncertainty estimation (Section 4.1)
    \item Qualitative result for uncertainty maps on KITTI dataset (Section 4.2)
\end{itemize}

\begin{figure}[t!]
    \centering
    \begin{subfigure}[]
        {\includegraphics[width=0.340\columnwidth]{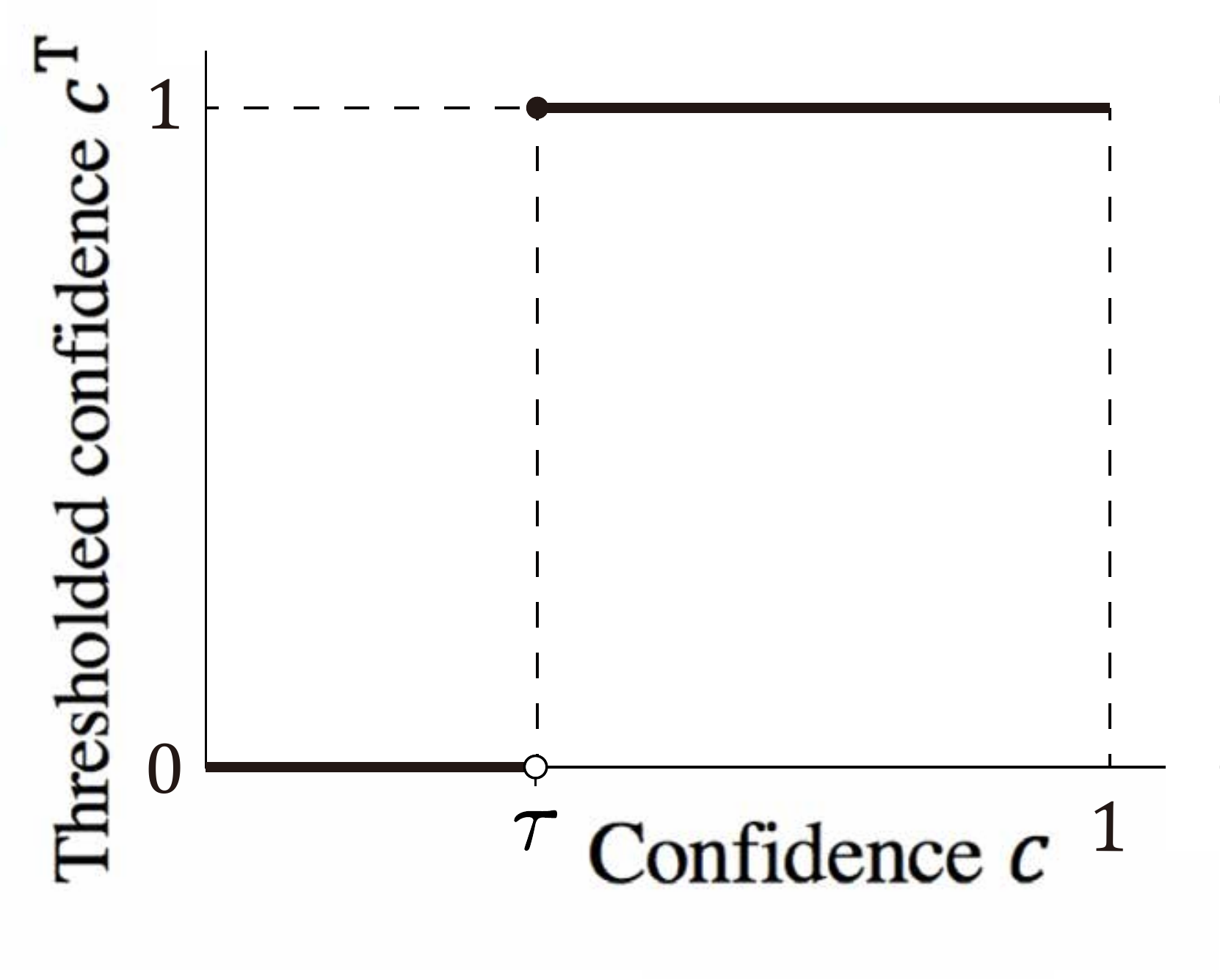}}
    \end{subfigure}
    \hspace{-0.4cm}
    \begin{subfigure}[]
        {\includegraphics[width=0.340\columnwidth]{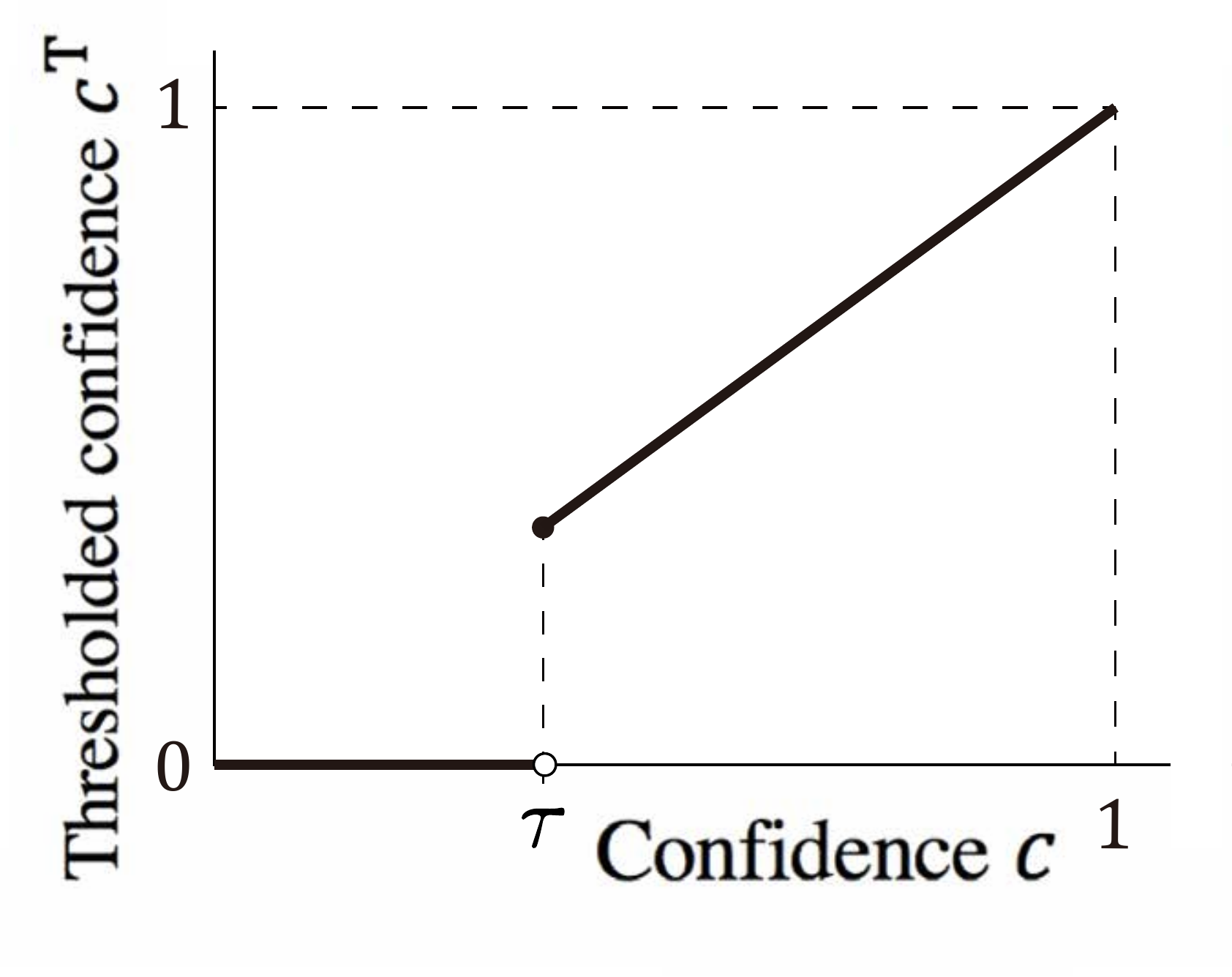}}
    \end{subfigure}
    \hspace{-0.4cm}
    \begin{subfigure}[]
        {\includegraphics[width=0.340\columnwidth]{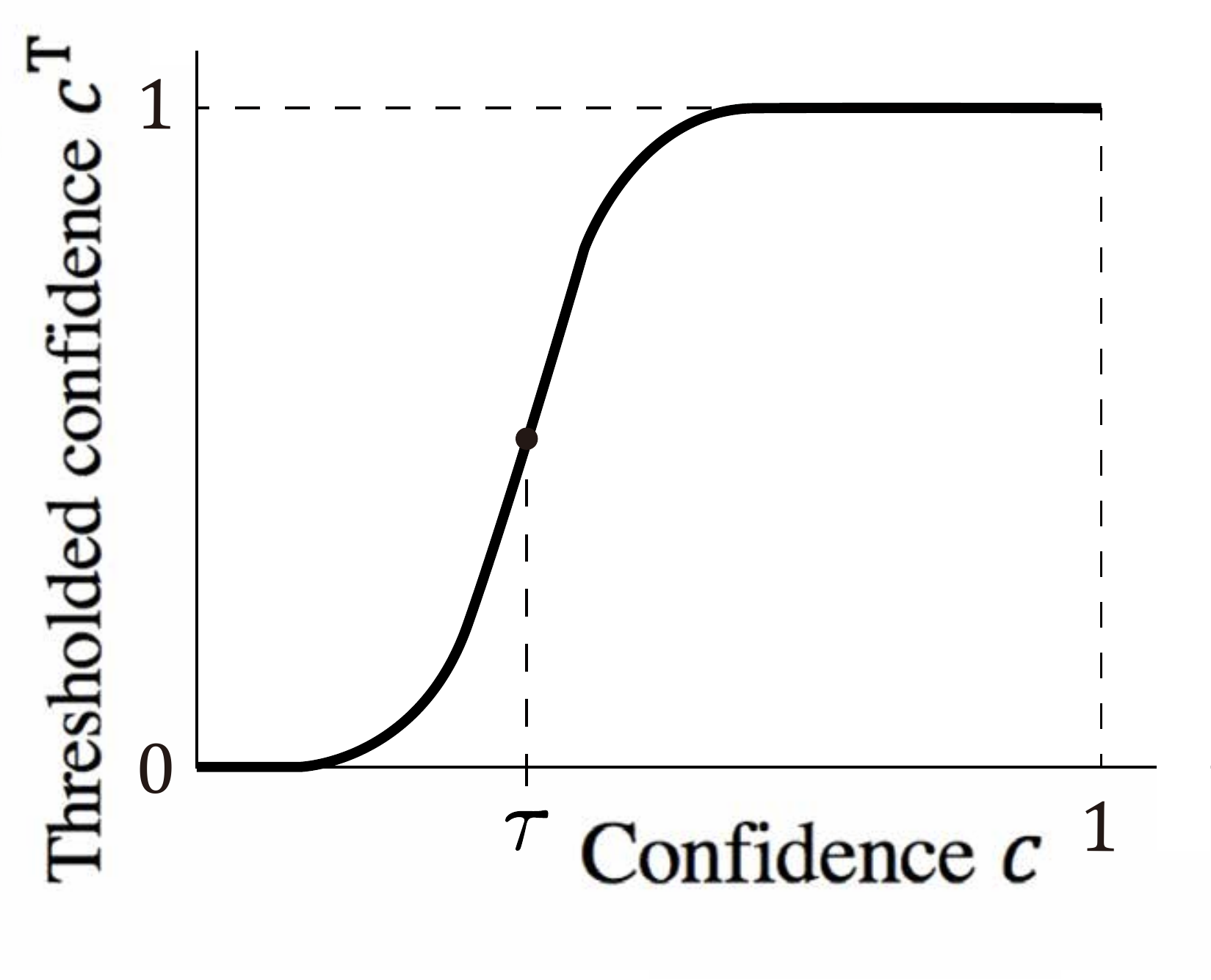}}
    \end{subfigure}
    %\vspace{-0.3cm}
    \caption{Comparison of confidence thresholding operator: (a) hard-thresholding used in \cite{cho2019large},
        (b) hard-thresolding function used in \cite{tonioni2019unsupervised}, and (c) our soft-thresholding function.
        The learned threshold is used in (b) and (c), while the threshold is fixed in (a) for all training images.
    }
    \label{slope_maskeq}
    \vspace{-0.3cm}
\end{figure}

\begin{figure}[t!]
    \centering
    \begin{subfigure}[]
        {\includegraphics[width=0.340\columnwidth]{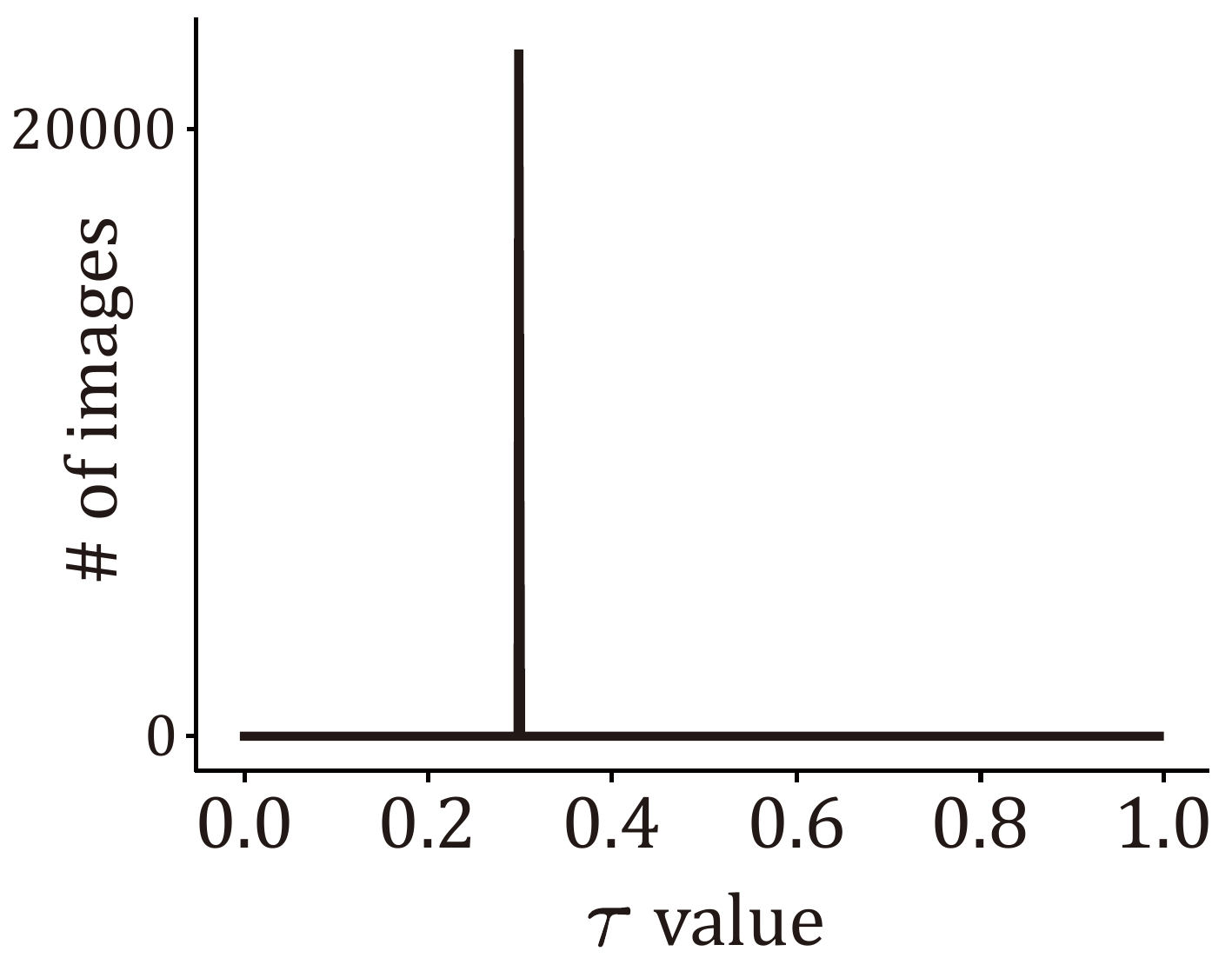}}
    \end{subfigure}
    \hspace{-0.4cm}
    \begin{subfigure}[]
        {\includegraphics[width=0.340\columnwidth]{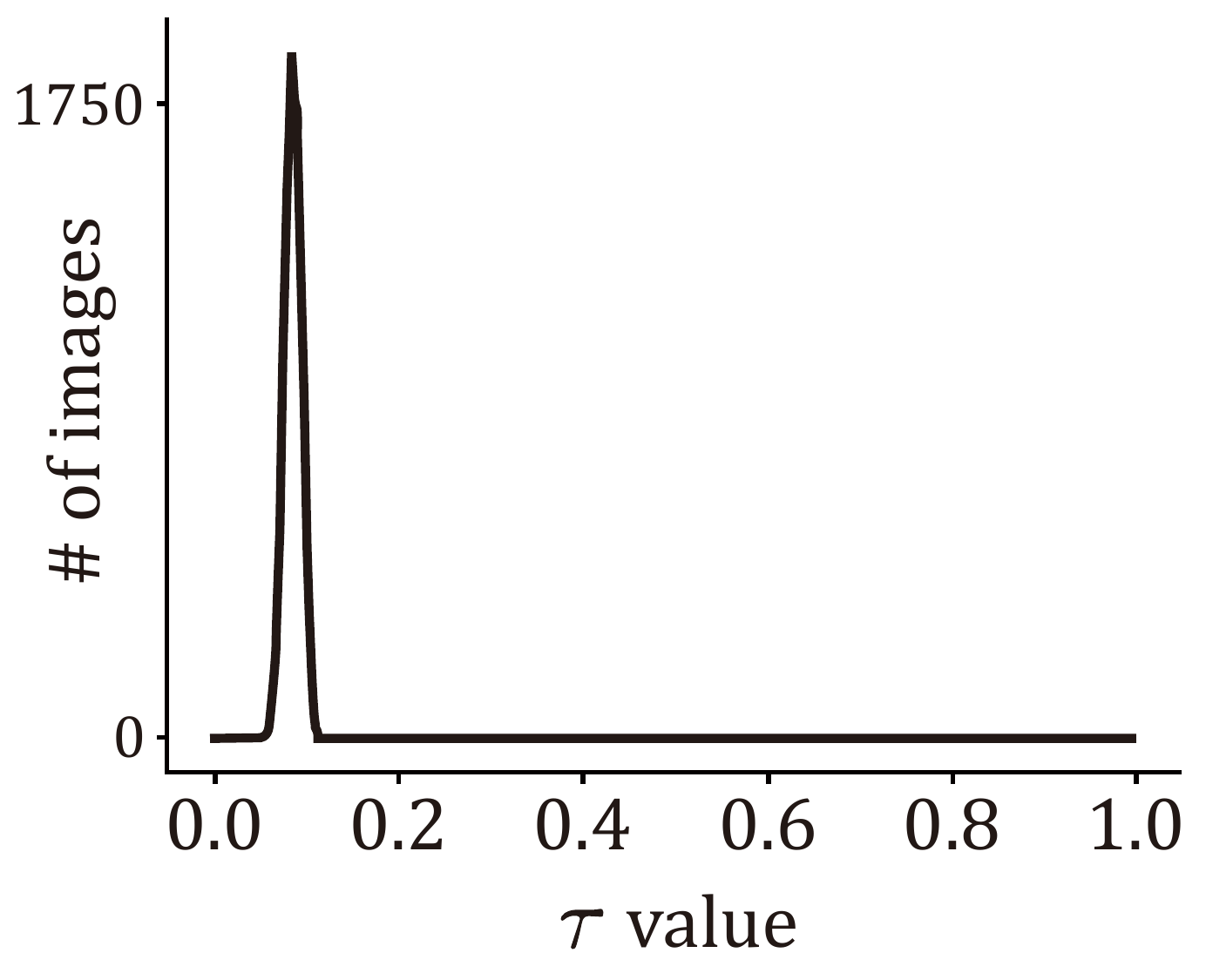}}
    \end{subfigure}
    \hspace{-0.4cm}
    \begin{subfigure}[]
        {\includegraphics[width=0.340\columnwidth]{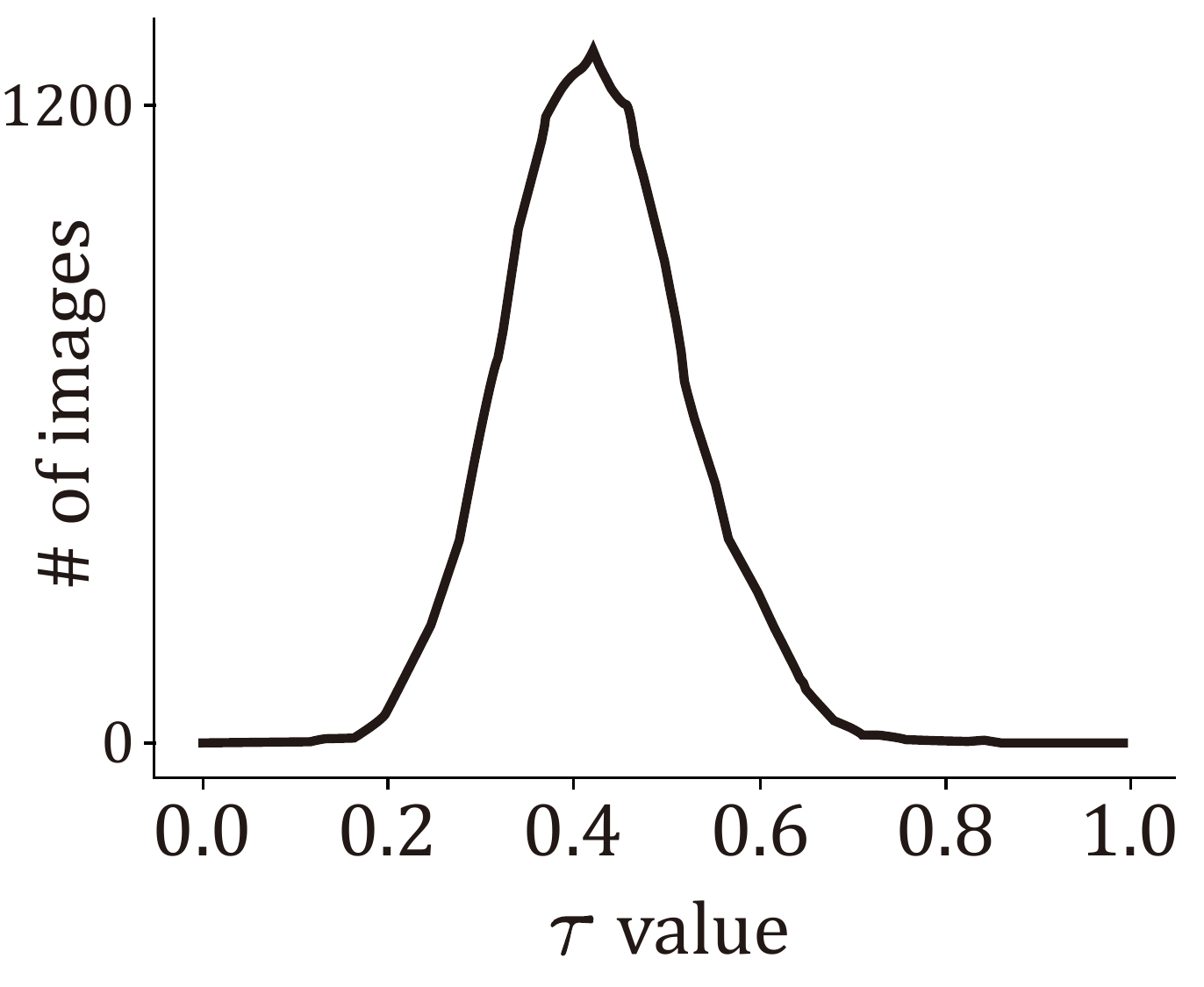}}
    \end{subfigure}
    %\vspace{-0.3cm}
    \caption{Histogram of the learned threshold $\tau$ on three different thresholding methods: (a) hard-thresholding used in \cite{cho2019large}, (b) hard-thresolding function used in \cite{tonioni2019unsupervised}, and (c) our soft-thresholding function.}
    \label{tau_distribution}
    \vspace{-0.3cm}
\end{figure}

\begin{figure}[t]
	\centering
	
	\begin{subfigure}[]
		{\includegraphics[width=0.245\columnwidth]{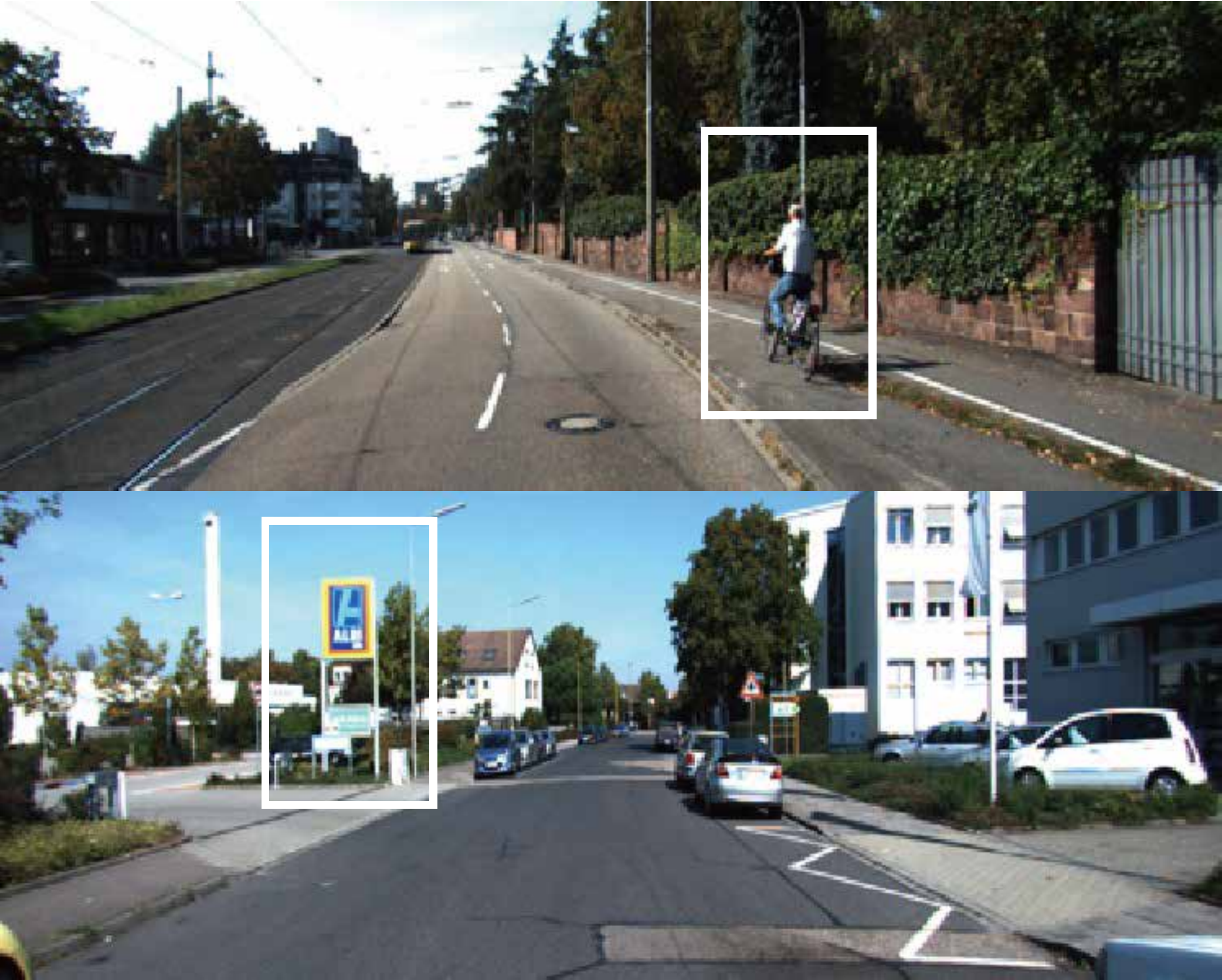}}
	\end{subfigure}
	\hspace{-0.2cm}
	\begin{subfigure}[]
		{\includegraphics[width=0.245\columnwidth]{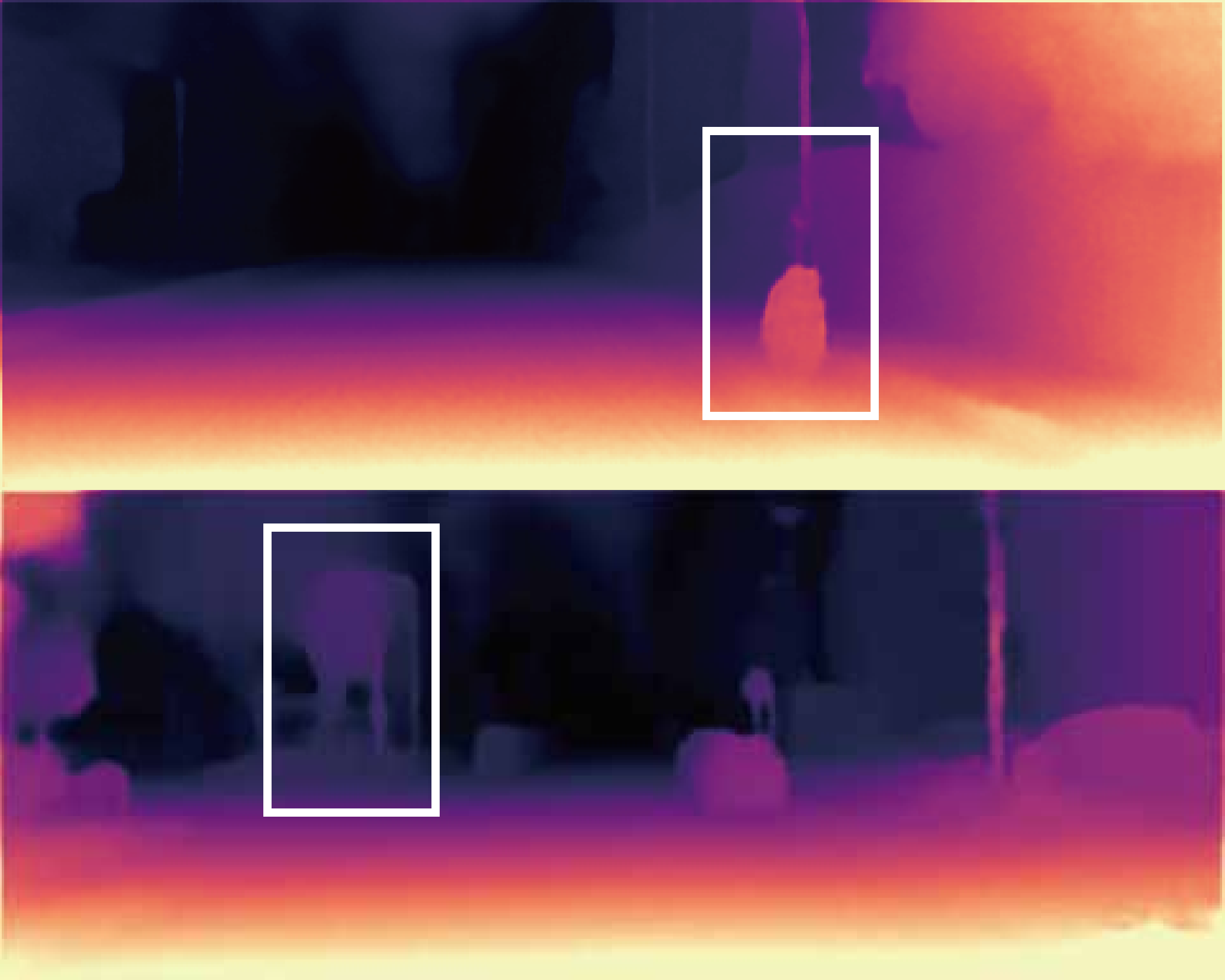}}
	\end{subfigure}
	\hspace{-0.2cm}
	\begin{subfigure}[]
		{\includegraphics[width=0.245\columnwidth]{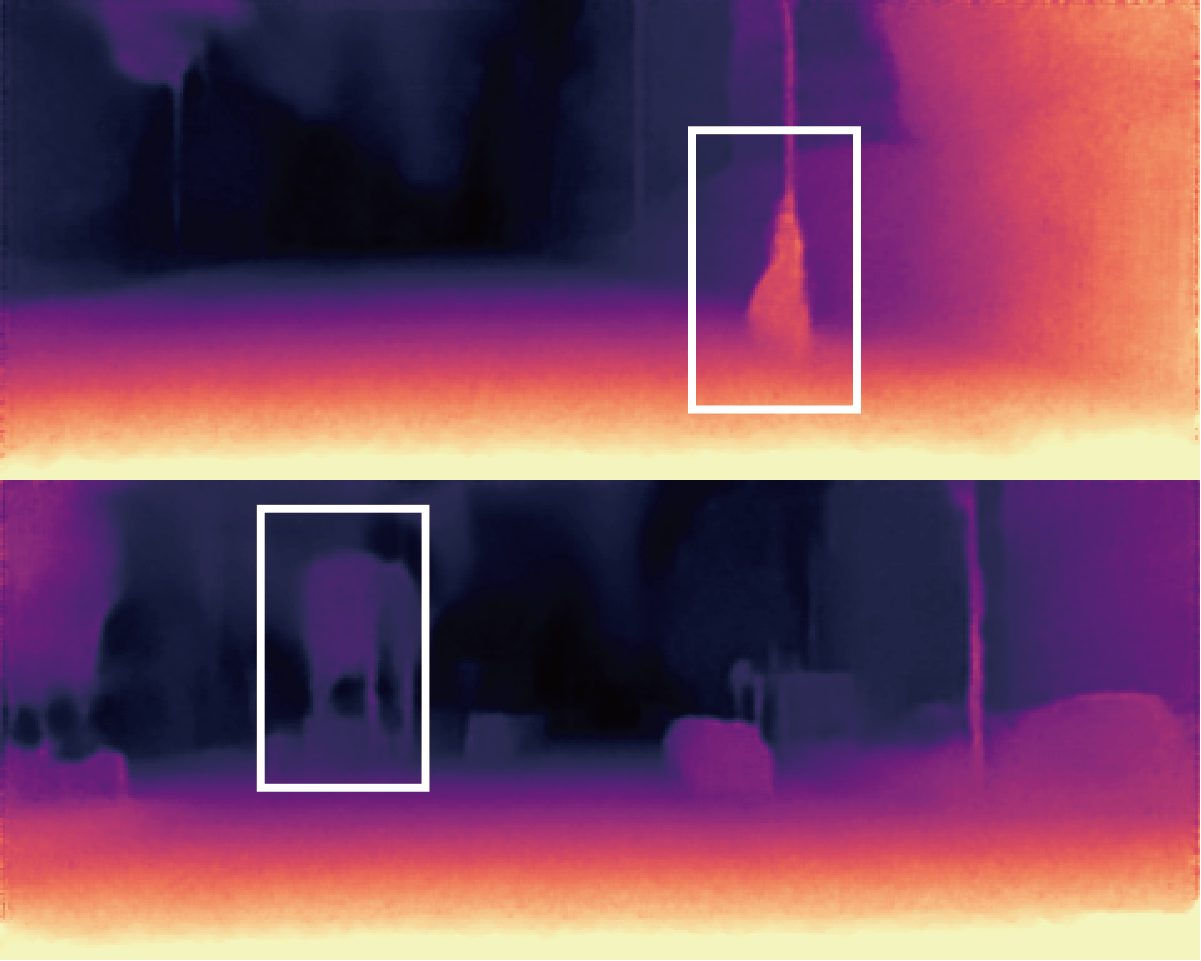}}
	\end{subfigure}
	\hspace{-0.2cm}
	\begin{subfigure}[]
		{\includegraphics[width=0.245\columnwidth]{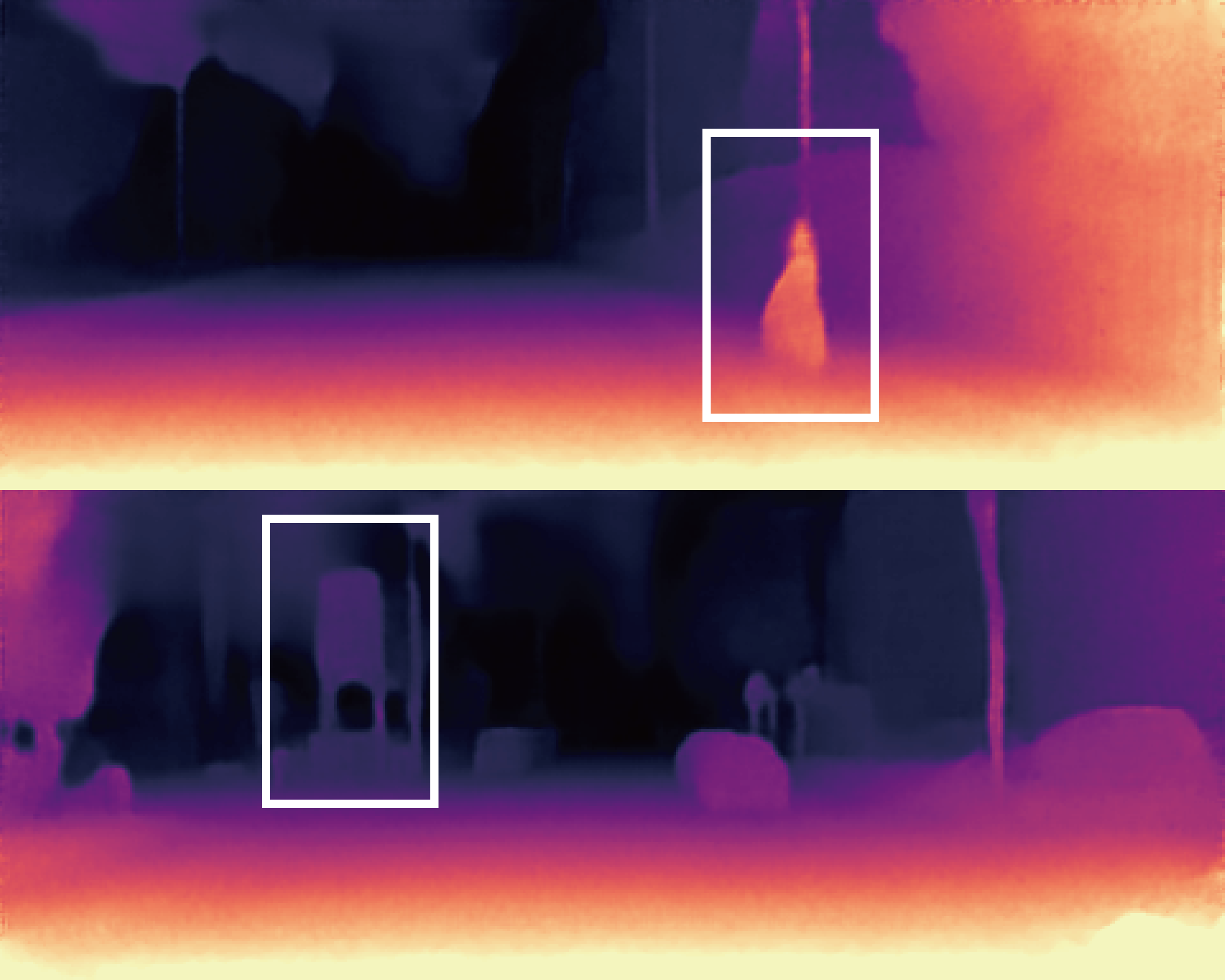}}
	\end{subfigure}
	%\vspace{-0.3cm}
	\caption{Examples of gradually improved depth results of (a) input image, from (b) using the monocular depth network with fixed threshold $\tau = 0.3$ \cite{cho2019large}, (c) using the monocular depth network with learned threshold $\tau$ \cite{tonioni2019unsupervised}, and to (d) using the proposed method.}
	\label{Toy_example}
	\vspace{-0.25cm}
\end{figure}

\begin{figure*}[]
	\centering
	\begin{subfigure}[]
		{\includegraphics[width=0.16\textwidth]{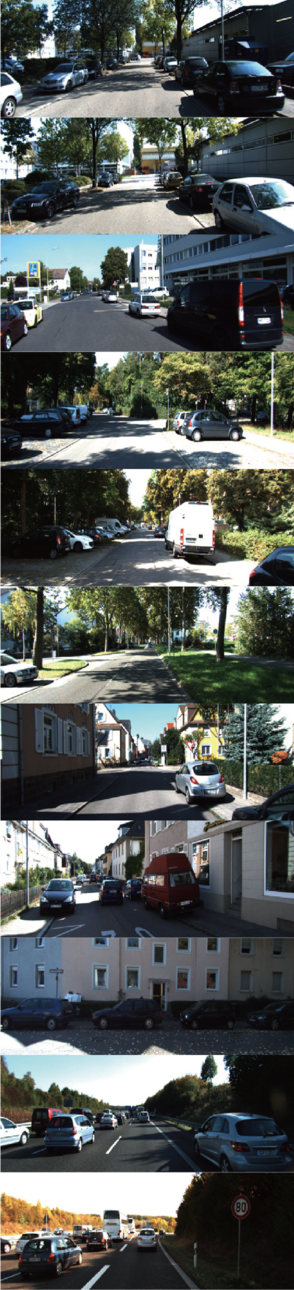}}
	\end{subfigure}
	\hspace{-0.21cm}
	\begin{subfigure}[]
		{\includegraphics[width=0.16\textwidth]{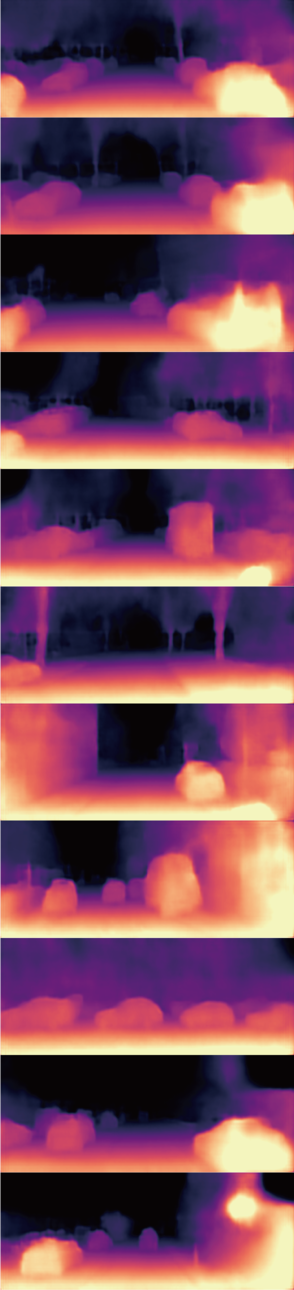}}
	\end{subfigure}
	\hspace{-0.21cm}
	\begin{subfigure}[]
		{\includegraphics[width=0.16\textwidth]{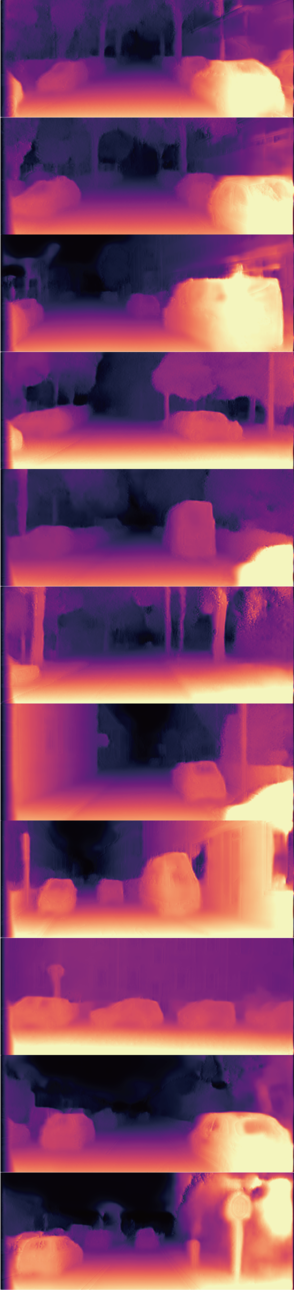}}
	\end{subfigure}
	\hspace{-0.21cm}
	\begin{subfigure}[]
		{\includegraphics[width=0.16\textwidth]{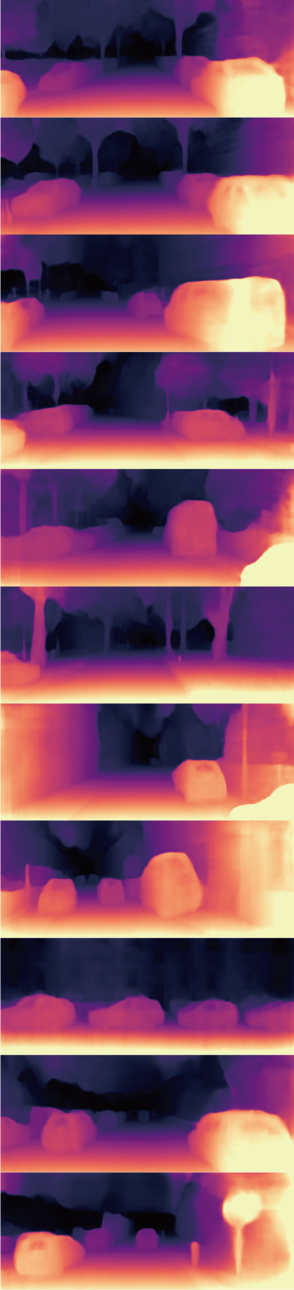}}
	\end{subfigure}
	\hspace{-0.21cm}
	\begin{subfigure}[]
		{\includegraphics[width=0.16\textwidth]{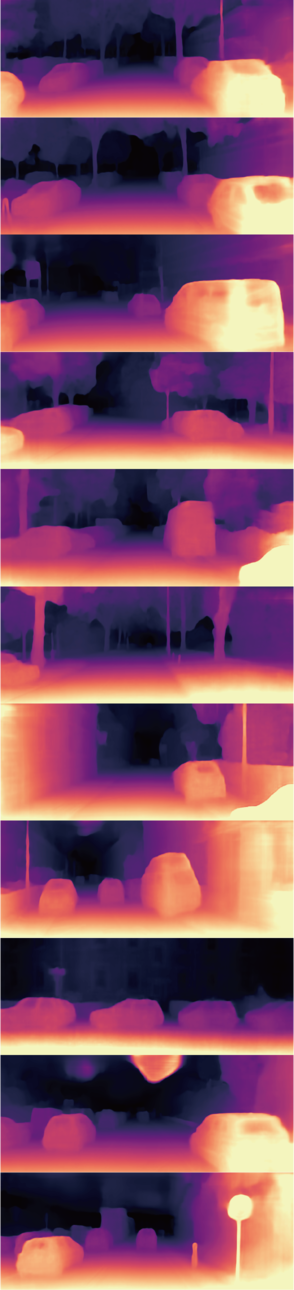}}
	\end{subfigure}
	\hspace{-0.21cm}
	\begin{subfigure}[]
		{\includegraphics[width=0.16\textwidth]{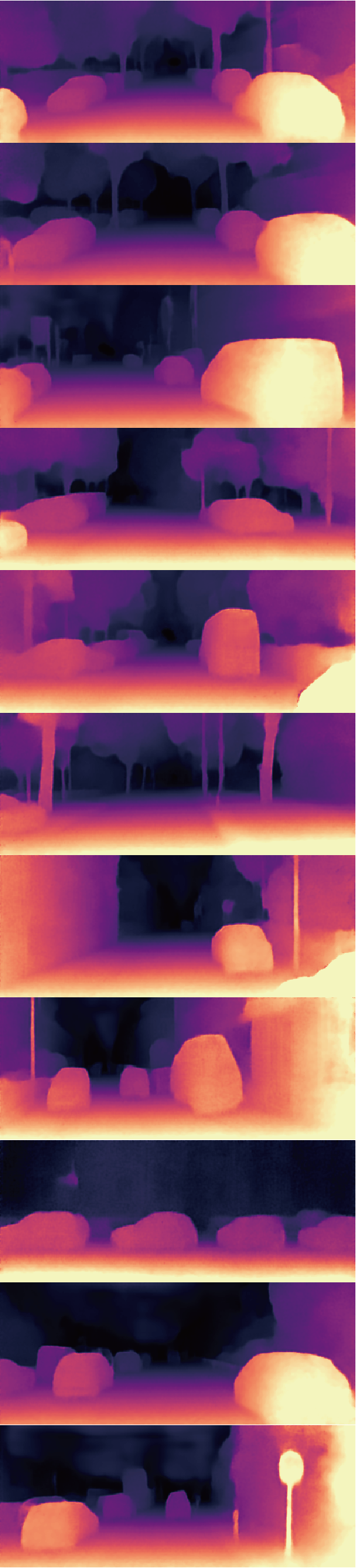}}
	\end{subfigure}
	\caption{Qualitative evaluation with existing monocular depth estimation methods on the Eigen split \cite{eigen2014depth} of KITTI dataset: (a) input image, (b) Kuznietsov \textit{et al.}~\cite{kuznietsov2017semi}, (c) Monodepth \cite{godard2017unsupervised}, (d) Monodepth2 \cite{godard2019digging}, (e) DepthHint \cite{watson2019self}, and (f) Ours (D+R) in the submitted manuscript. Compared to other results, our method predicts instances very well without holes or distortions while recovering fine object boundaries.
		Additionally, our method is capable of predicting thin instances precisely.}
	\label{Supp_Qual_KITTI}
\end{figure*}

\begin{figure*}[]
	\centering
	\begin{subfigure}[]
		{\includegraphics[width=0.203\textwidth]{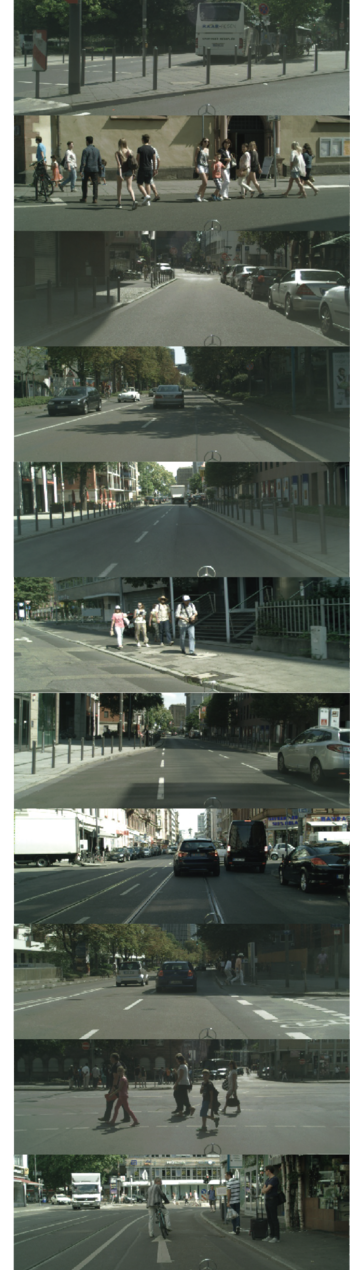}}
	\end{subfigure}
	\hspace{-0.35cm}
	\begin{subfigure}[]
		{\includegraphics[width=0.203\textwidth]{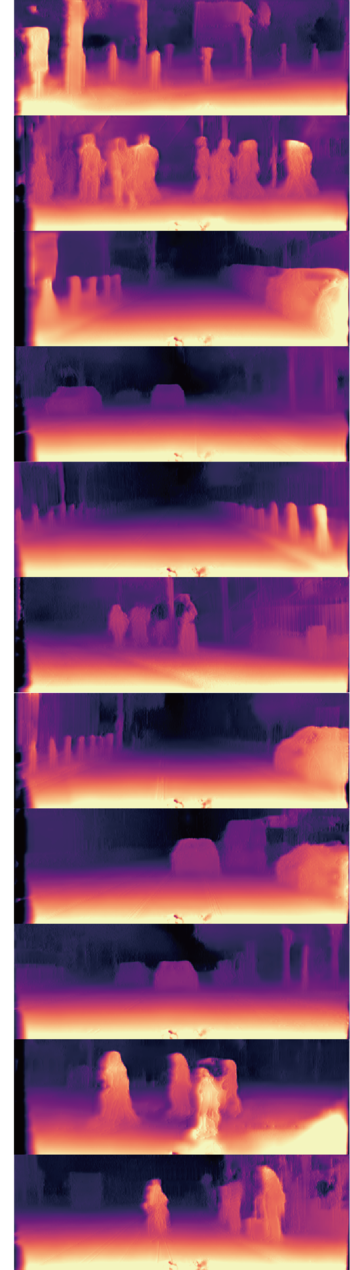}}
	\end{subfigure}
	\hspace{-0.35cm}
	\begin{subfigure}[]
		{\includegraphics[width=0.203\textwidth]{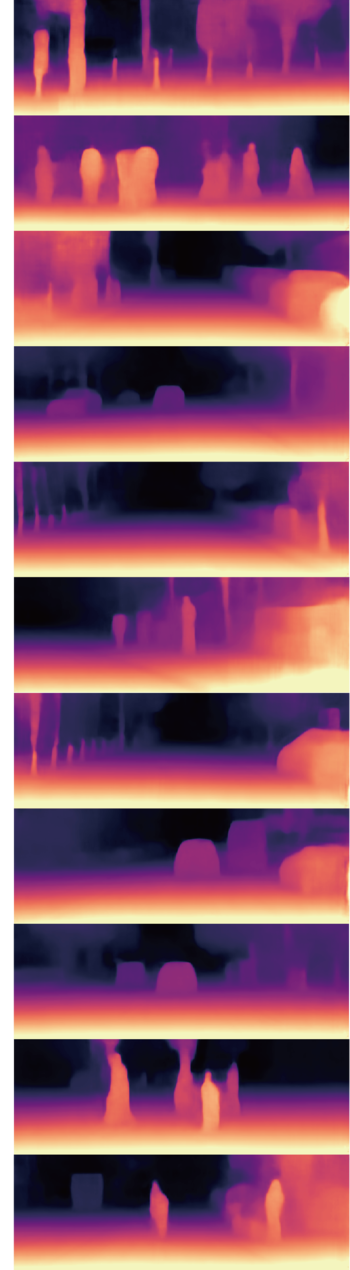}}
	\end{subfigure}
	\hspace{-0.35cm}
	\begin{subfigure}[]
		{\includegraphics[width=0.203\textwidth]{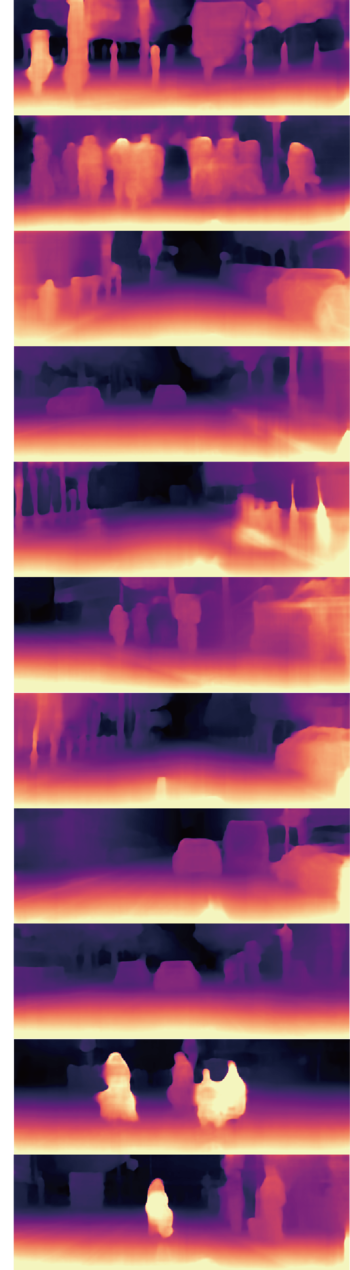}}
	\end{subfigure}
	\hspace{-0.35cm}
	\begin{subfigure}[]
		{\includegraphics[width=0.203\textwidth]{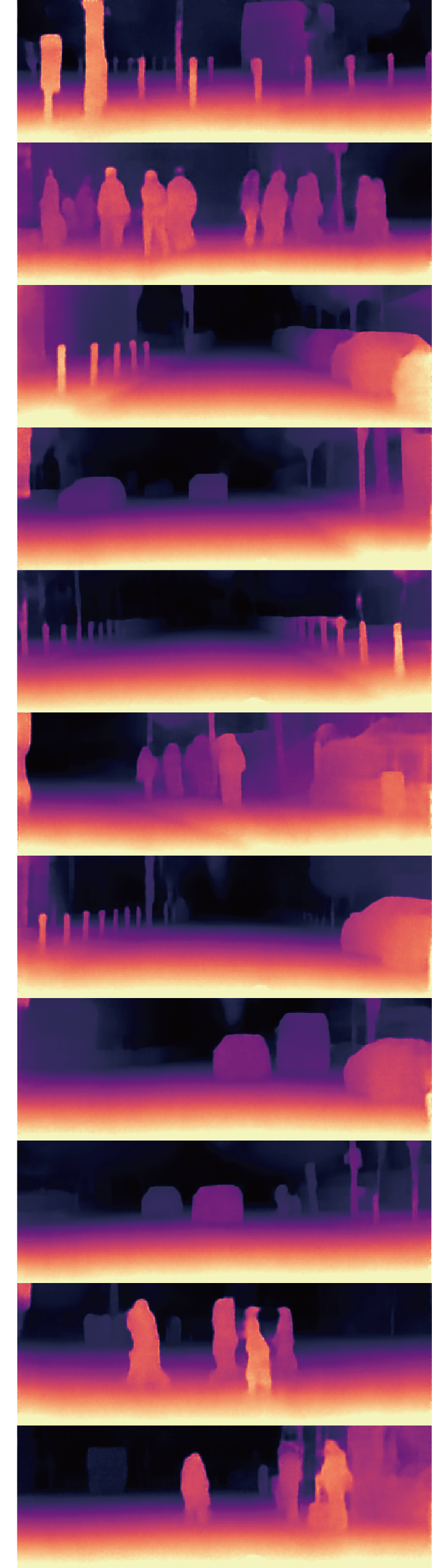}}
	\end{subfigure}
	\caption{Qualitative evaluation for depth estimation with existing methods on Cityscapes validation dataset: (a) input image, (b) \cite{godard2017unsupervised}, (c) \cite{tosi2019learning}, (d) \cite{watson2019self} and (e) Ours (D+R) of the original manuscript. Similar to KITTI results, our method is remarkable at predicting fine details with no distortions at all instances and recovering thin objects that appear frequently in the Cityscapes dataset, whereas other methods often fail to predict accurate depth values at these regions.}
	\label{Supp_Qual_City}
\end{figure*}

\section{Comparison with thresholding methods}
This section further highlights the effectiveness of our thresholding method by analyzing the distribution of learned threshold $\tau$. %In our original manuscript, we compared two hard thresholding methods \cite{cho2019large, tonioni2019unsupervised} with the proposed method. However, only the proposed method was able to select the threshold for each image adatively, and exclude unreliable pseudo depth values effectively. The reason will be explained in detail.
\figref{slope_maskeq} recaps the thresholding operators (Fig. 2 of the paper) for the completeness of supplementary document. %Therefore, the proposed method is the sole soft-thresholding which is implemented to be differentiable.
To analyze the distribution of the learned threshold $\tau$, we plotted the histogram of $\tau$ values learned using 20k images of KITTI training dataset \cite{geiger2013vision} in \figref{tau_distribution}.
For a fair comparison, all experiments were conducted under the same environment, including network architecture, loss function, and training data.
All results were obtained without the probabilistic refinement of the RefineNet.

Cho \textit{et al.}~\cite{cho2019large} fixed the threshold to $0.3$ for all training images.
Tonioni \textit{et al.}~\cite{tonioni2019unsupervised} attempted to learn the threshold adaptively for each image by applying the regularization loss $-\log(1-\tau)$,
but it simply prevents $\tau$ values from converging to 0 or 1 and does not take into account image characteristics that enable $\tau$ to be learned adaptively.
As shown in \figref{tau_distribution} (b), $\tau$ values predicted by \cite{tonioni2019unsupervised} are concentrated around specific values (0.1) with very small variance, meaning that almost similar threshold $\tau$ is used for all training images.
This is the reason why the performance gain of Tonioni \textit{et al.}~\cite{tonioni2019unsupervised} over Cho \textit{et al.}~\cite{cho2019large} is relatively marginal, as reported in Tab. 4 of the original manuscript.
%Our method uses a set of stereo image pairs and extremely sparse LiDAR depth maps of 3$\%$ density as a explicit supervision.
Contrarily, our method learns image-adaptive $\tau$ values as plotted in \figref{tau_distribution} (c).
Fig. 3 of the original manuscript also reports that the proposed method learned the threshold $\tau$ accordingly.
The threshold $\tau$ was set low in the images where depth inference is easy, while being set high in the opposite case including saturation, low-light, and textureless region.

%Consequentially, our thresholding method that learns the threshold $\tau$ with considering image characteristics led to performance improvement.
\figref{Toy_example} shows the examples of gradually improved depth results according to different thresholding methods.
The proposed method yields qualitatively better results, where complete instances are recovered and fine object boundaries are well preserved, than other hard thresholding methods.
Also, in Tab. 4 of the original manuscript, it can be seen that the proposed method outperforms the two methods quantitatively.

\section{Monocular depth estimation results}
This section provides more results for comparative study with state-of-the-art methods in terms of monocular depth accuracy.

\subsection{Evaluation metrics}
In order to evaluate the depth estimation performance, same as the original manuscript, five commonly-used evaluation metrics proposed in \cite{eigen2014depth} were adopted as follows:

\begin{itemize}
    \item Abs Rel = ${\frac{1}{\left | \Omega \right |}\sum_{p \in \Omega}{\frac{\left |  d_p - d_p^{\mathrm{gt}}\right |}{d_p^{gt}}}}$
    \item Sq Rel = ${\frac{1}{\left | \Omega \right |}\sum_{p \in \Omega}{\frac{(d_p - d_p^{\mathrm{gt}})^2}{d_p^{\mathrm{gt}}}}}$
    \item RMSE = $\sqrt{{\frac{1}{\left | \Omega \right |}\sum_{p \in \Omega}{(d_p - d_p^{\mathrm{gt}})^2}}}$
    \item RMSE log = $\sqrt{{\frac{1}{\left | \Omega \right |}\sum_{p \in \Omega}{(log(d_p) - log(d_p^{\mathrm{gt}}))^2}}}$
    \item $\delta < 1.25^{n} = \%$ of $d_p$  s.t.  $\delta = \max(\frac{d_p}{d_p^{\mathrm{gt}}}, \frac{d_p^{\mathrm{gt}}}{d_p}) < 1.25^n \\ {\mathrm{for}}\;\; n=1,2,3$,
    % Thorough experiments are provided on KITTI dataset \cite{geiger2013vision} and Cityscapes dataset \cite{cordts2016cityscapes} for the monocular depth estimation, and on Middlebury \cite{scharstein2014high} and KITTI 2015 dataset for confidence estimation.
    %    \item An intensive ablation study provides the quantitative evaluation of the proposed method.
\end{itemize}

\noindent where $d_p$ and $d_p^{\mathrm{gt}}$ indicate the estimated depth map and ground truth depth map at a pixel $p$, respectively. $\Omega$ represents a set of valid pixels.
%    \db{The equation are normalized using $Z=\sum\limits c^\mathrm{T}_p(\tau)$.????? USE different notation not Z here.}

%\begin{table*}[]
%   \centering
%   %\small\addtolength{\tabcolsep}{-2pt}
%   \caption{Quantitative evaluation for depth estimation with existing methods on NYU-v2 test dataset with fine-tuning on NYU-v2 training dataset. Numbers in bold and underlined represent $1^{st}$ and $2^{nd}$ ranking, respectively.}
%   \begin{tabular}{@{}ccccccccccc@{}}
%       \toprule
%       &                                    & \multicolumn{4}{c}{Lower is better}                       & \multicolumn{3}{c}{Accuracy: higher is better}    \\ \midrule
%       \multicolumn{1}{c|}{Method} & \multicolumn{1}{c|}{Data} & rms & rel  & \multicolumn{1}{c|}{log10} & $\delta<1.25$ & $\delta<1.25^2$ & $\delta<1.25^3$ \\ \midrule
%       MovingIndoor \cite{zhou2019moving} & M  & 0.712   & 0.208   & 0.086  & 0.674  & 0.900   & 0.968  \\
%       JointDepth-Pose \cite{zhao2020towards}  & M    & 0.686   & 0.189   & 0.079  & 0.701  & 0.912 & 0.978  \\
%       Monodepth2 \cite{godard2019digging}     & M    & 0.617   & 0.170 & 0.072 & 0.748 & 0.942 & 0.986  \\
%       Ours (D)                & LP     & 0.584   & 0.155   & 0.072  & 0.759  & 0.943   & 0.986  \\
%       Ours (D+R)                & LP     & 0.571 & 0.152 & 0.069 & 0.776 & 0.944   & 0.987  \\ \bottomrule
%   \end{tabular}
%   \label{Supp_Quan_nyu}
%   %\vspace{-0.3cm}
%\end{table*}

\begin{table*}[h]
	\centering
	\small\addtolength{\tabcolsep}{-2pt}
	\caption{Quantitative evaluation for depth estimation with existing methods on Cityscapes validation dataset without fine-tuning on Cityscapes training dataset. Numbers in bold and underlined represent $1^{st}$ and $2^{nd}$ ranking, respectively.}
	\begin{tabular}{@{}ccccccccccc@{}}
		\toprule
		&                                    & \multicolumn{4}{c}{Lower is better}                       & \multicolumn{3}{c}{Accuracy: higher is better}    \\ \midrule
		\multicolumn{1}{c|}{Method} & \multicolumn{1}{c|}{Data} & Abs Rel & Sqr Rel & RMSE  & \multicolumn{1}{c|}{RMSE log} & $\delta<1.25$ & $\delta<1.25^2$ & $\delta<1.25^3$ \\ \midrule
		Monodepth \cite{godard2017unsupervised}                   & S              & 0.631   & 10.257   & 13.424 & 0.525                         & 0.281         & 0.546           & 0.749           \\
		MonoResMatch \cite{tosi2019learning}                & S    & 0.241   & 2.149   & 9.064 & 0.296                         & 0.570         & 0.891           & 0.966           \\
		PackNet-SfM \cite{guizilini20203d}            & M     & 0.245   & 2.240   & 8.920 & 0.298                       & 0.557         & 0.892           & 0.967           \\
		Monodepth2 \cite{godard2019digging}                  & S    & 0.242   & 2.308   & 8.563 & 0.290                         & 0.591         & 0.904           & 0.971           \\
		DepthHint \cite{watson2019self}         & S     & \textbf{0.220}   & 2.008   & 8.363 & \textbf{0.273}                         & 0.613         & 0.922           & \underline{0.975}           \\
		Ours (D)                & S     & 0.238   & \underline{1.983}   & \underline{8.176}   &0.282  & \underline{0.629}  & \underline{0.923}   &\textbf{0.976}  \\
		Ours (D+R)                & S     & \underline{0.225}   & \textbf{1.962}   & \textbf{8.010}   &\underline{0.276}  & \textbf{0.631}  & \textbf{0.924}   &\textbf{0.976}  \\ \bottomrule
	\end{tabular}
	\label{Supp_Quan_city}
	%\vspace{-0.3cm}
\end{table*}

\begin{table*}[]
    \centering
    \caption{Quantitative evaluation for monocular depth estimation with existing methods on KITTI Eigen split dataset \cite{eigen2014depth} with improved ground truth depth maps \cite{uhrig2017sparsity}. Numbers in bold and underlined represent $1^{st}$ and $2^{nd}$ ranking, respectively.}
    \begin{tabular}{@{}ccccccccc@{}}
        \toprule
        \multicolumn{1}{l}{}        & \multicolumn{1}{l}{}      & \multicolumn{4}{c}{Lower is better}                                                                                     & \multicolumn{3}{c}{Accuracy: higher is better}                                              \\ \midrule
        \multicolumn{1}{c|}{Method} & \multicolumn{1}{c|}{Data} & \multicolumn{1}{c|}{Abs Rel} & \multicolumn{1}{c|}{Sqr Rel} & \multicolumn{1}{c|}{RMSE} & \multicolumn{1}{c|}{RMSE log} & \multicolumn{1}{c|}{$\delta<1.25$} & \multicolumn{1}{c|}{$\delta<1.25^2$} & $\delta<1.25^3$ \\ \midrule
        SfMLeaner \cite{zhou2017unsupervised}                   & M                         & 0.176                        & 1.532                        & 6.129                     & 0.244                         & 0.758                              & 0.921                                & 0.971           \\
        Vid2Depth \cite{mahjourian2018unsupervised}                   & M                         & 0.134                        & 0.983                        & 5.501                     & 0.203                         & 0.827                              & 0.944                                & 0.981           \\
        DDVO \cite{wang2018learning}                        & M                         & 0.126                        & 0.866                        & 4.932                     & 0.185                         & 0.851                              & 0.958                                & 0.986           \\
        EPC++ \cite{luo2019every}                       & M                         & 0.120                        & 0.789                        & 4.755                     & 0.177                         & 0.856                              & 0.961                                & 0.987           \\
        Monodepth2 \cite{godard2019digging}                  & S                         & 0.090                        & 0.545                        & 3.942                     & 0.137                         & 0.914                              & 0.983                                & \underline{0.995}           \\
        Uncertainty (Boot+Log) \cite{poggi2020uncertainty}      & S                         & 0.085                        & 0.511                        & 3.777                     & 0.137                         & 0.913                              & 0.980                                & 0.994           \\
        Uncertainty (Boot+Self) \cite{poggi2020uncertainty}     & S                         & 0.085                        & 0.510                        & 3.792                     & 0.135                         & 0.914                              & 0.981                                & 0.994           \\
        Uncertainty (Snap+Log) \cite{poggi2020uncertainty}      & S                         & 0.084                        & 0.529                        & 3.833                     & 0.138                         & 0.914                              & 0.980                                & 0.994           \\
        Uncertainty (Snap+Self) \cite{poggi2020uncertainty}     & S                         & 0.086                        & 0.532                        & 3.858                     & 0.138                         & 0.912                              & 0.980                                & 0.994           \\
        UnRectDepthNet \cite{kumar2020unrectdepthnet}              & M                         & 0.081                        & 0.414                        & 3.412                     & \textbf{0.117}                & 0.926                              & \textbf{0.987}                          & \textbf{0.996}     \\
        PackNet-SfM \cite{guizilini20203d}                 & M                         & \underline{0.078}               & 0.420                        & 3.485                     & 0.121                         & \textbf{0.931}                     & \underline{0.986}                                & \textbf{0.996}     \\
        Ours (D)            & S                     & \underline{0.078}               & \underline{0.361}               & \underline{3.223}            & 0.120                   & \underline{0.930}                        & \textbf{0.987}                          & \textbf{0.996}     \\
        Ours (D+R)            & S                     & \textbf{0.076}               & \textbf{0.340}              & \textbf{3.171}           & \underline{0.119}                   & \textbf{0.931}                        & \textbf{0.987}                          & \textbf{0.996}     \\ \bottomrule
    \end{tabular}
    \label{Supp_Quan_Improved}
\end{table*}

\begin{table*}[]
    \centering
    \caption{Quantitative depth estimation results according to $\varepsilon$ value evaluated on KITTI Eigen Split \cite{eigen2014depth} raw dataset.}
    \begin{tabular}{@{}c|ccccccc@{}}
        \toprule
        $\varepsilon$  & \multicolumn{1}{c|}{Abs Rel} & \multicolumn{1}{c|}{Sqr Rel} & \multicolumn{1}{c|}{RMSE} & \multicolumn{1}{c|}{RMSE log} & \multicolumn{1}{c|}{$\delta<1.25$} & \multicolumn{1}{c|}{$\delta<1.25^2$} & $\delta<1.25^3$ \\ \midrule
        \textbf{10} & 0.099                        & 0.652                        & 4.266                     & 0.187                         & 0.883                              & 0.960                                & 0.981           \\
        \textbf{30} & 0.102                        & 0.657                        & 4.290                     & 0.189                         & 0.881                              & 0.959                                & 0.980           \\
        \textbf{50} & 0.100                        & 0.649                        & 4.272                     & 0.188                         & 0.881                              & 0.959                                & 0.979           \\ \bottomrule
    \end{tabular}
    \label{Qual_epsilon}
\end{table*}

\begin{table*}[]
    \centering
    \caption{Quantitative depth estimation results for three cases of the proposed method on KITTI Eigen Split \cite{eigen2014depth} raw dataset.}
    \begin{tabular}{@{}c|ccccccc@{}}
        \toprule
        Method  & \multicolumn{1}{c|}{Abs Rel} & \multicolumn{1}{c|}{Sqr Rel} & \multicolumn{1}{c|}{RMSE} & \multicolumn{1}{c|}{RMSE log} & \multicolumn{1}{c|}{$\delta<1.25$} & \multicolumn{1}{c|}{$\delta<1.25^2$} & $\delta<1.25^3$ \\ \midrule
        Ours (D) & 0.099                        & 0.652                        & 4.266                     & 0.187                         & 0.883                              & 0.960                                & 0.981           \\
        Ours (R) & 0.096                        & 0.646                        & 4.280                     & 0.189                         & 0.882                              & 0.959                                & 0.980           \\
        Ours (D+R) & 0.096                        & 0.629                        & 4.187                     & 0.185                         & 0.887                              & 0.963                                & 0.983           \\ \bottomrule
    \end{tabular}
    \label{Supp_Quan_method}
\end{table*}

\subsection{Qualitative evaluation on KITTI}
\figref{Supp_Qual_KITTI} shows more results on the Eigen Split \cite{eigen2014depth} of KITTI dataset. %the qualitative evaluation with existing monocular depth estimation methods on
We compared our results with (b) Kuznietsov \textit{et al.}~\cite{kuznietsov2017semi}, (c) Monodepth \cite{godard2017unsupervised}, (d) Monodepth2 \cite{godard2019digging}, (e) DepthHint \cite{watson2019self}, and (f) Ours (D+R).
Compared to other results, our method predicts instances very well without holes or distortions while recovering fine object boundaries.
Additionally, our method is capable of predicting thin objects precisely.

\subsection{Qualitative evaluation on Cityscapes dataset}
\figref{Supp_Qual_City} shows more qualitative results on Cityscapes dataset \cite{cordts2016cityscapes} of Fig. 5 in original manuscript. Note that it is fine-tuned on Cityscapes dataset.
We compared our results with three existing methods: (b) \cite{godard2017unsupervised}, (c) \cite{tosi2019learning}, (d) \cite{watson2019self} and (e) Ours (D+R) in the original manuscript.
Similar to KITTI results, our method is remarkable at predicting fine details with no distortions at all instances and recovering thin objects that appear frequently in the Cityscapes dataset, whereas other methods often fail to predict accurate depth values at these regions.
%has a bunch of thin instances, where our method outperforms others at.

\subsection{Quantitative evaluation on Cityscapes dataset without fine-tuning}
We also evaluated the performance of the proposed method on the Cityscapes dataset without fine-tuning.
\tabref{Supp_Quan_city} provides the quantitative evaluation on the Cityscapes validation dataset \cite{cordts2016cityscapes}, setting maximum depth to 80 meters.
%Note that the Cityscapes results provided in the original manucript were fine-tuned on Cityscapes dataset.
The performance evaluation includes Monodepth \cite{godard2017unsupervised}, MonoResMatch \cite{tosi2019learning}, Monodepth2 \cite{godard2019digging}, DepthHint \cite{watson2019self}, PackNet-SfM \cite{guizilini20203d}.
Even without fine-tuning on the Cityscapes dataset, our method still outperforms state-of-the-arts approaches, and it shows that our model trained on KITTI dataset generalizes well on other dataset without bias.

%\subsection{Quantitative evaluation on NYU-v2 dataset with fine-tuning}
%To ensure our method's capability of generalization on the datasets that does not provide a stereo image pairs, we also evaluated our method on NYU-v2 dataset with depth network fine-tuned on NYU-v2 dataset, confidence network and threshold network freezed. In this case, our method chose raw kinect depth as the pseudo ground truth, and we tested our method on the 1449 dense labeled pairs of aligned depth maps. \figref{} shows the quantitative evaluation on NYU-v2 dataset with depth network fine-tuned  We compared our results with Monodepth2 \cite{godard2019digging}, DepthHint \cite{watson2019self} and PackNet-SfM \cite{guizilini20203d}. Relatively high performance of our method supports the claim that our confidence and threshold networks trained with the KITTI dataset show satisfactory generalization capability for different datasets.

\subsection{Quantitative evaluation on KITTI improved ground truth depth maps}
To strengthen credibility to quantitative evaluation, we also measured the monocular depth accuracy by using test frames with the improved ground truth depth maps made available in \cite{uhrig2017sparsity} for KITTI Eigen split dataset \cite{eigen2014depth}.
The improved ground truth maps are high quality depth maps generated by accumulating LiDAR point clouds from 5 consecutive frames.
\tabref{Supp_Quan_Improved} shows the quantitative evaluation with existing methods on the KITTI Eigen split dataset using the improved ground truth depth maps \cite{uhrig2017sparsity}.
We compared our results with `SfMLeaner' \cite{zhou2017unsupervised}, `Vid2Depth' \cite{mahjourian2018unsupervised}, `DDVO' \cite{wang2018learning}, `EPC++' \cite{luo2019every}, `Monodepth2' \cite{godard2019digging}, `Uncertainty' \cite{poggi2020uncertainty}, `PackNet-SfM' \cite{guizilini20203d}, and `UnRectDepthNet' \cite{kumar2020unrectdepthnet}.
For the training data, `M' and `S' indicate using a monocular video sequence and stereo images, respectively.
%These methods employed the monocular video as supervision except for `Uncertainty' using stereo image pairs.
Our method produces superior performance compared to other methods.
%Despite the fact that our method uses monocular image for training <- This cannot be weakness, if suitable evidence is not provided.

\subsection{Choice of $\varepsilon$ value}
We set $\varepsilon=10$ for the differentiable soft-thresholding function in (1) of the original manuscript.
\tabref{Qual_epsilon} shows the quantitative results according to $\varepsilon$ on the KITTI Eigen split \cite{eigen2014depth} raw dataset.
Though the best accuracy was achieved with $\varepsilon=10$, no significant change was observed depending on varying $\varepsilon$.

\subsection{Ablation study of DepthNet and RefineNet}
The evaluation of the proposed method was conducted for three cases; `Ours (D)' trained with only the DepthNet using $L_D$ without refining the depth map, `Ours (R)' trained with the DepthNet and RefineNet using $L_U$ only, and `Ours (D+R)' trained with the DepthNet and RefineNet using all losses.
\tabref{Supp_Quan_method} shows the quantitative results of the above three cases on KITTI Eigen Split \cite{eigen2014depth} dataset.
The performance gain of `Ours (D+R)' over `Ours (R)' supports the effectiveness of the proposed confidence learning.

\begin{figure*}[]
    \centering
    \begin{tabular}{@{}cccc@{}}
        \includegraphics[width=0.235\textwidth]{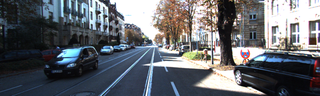} & \includegraphics[width=0.235\textwidth]{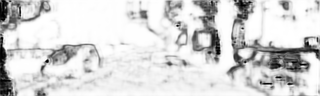} & \includegraphics[width=0.235\textwidth]{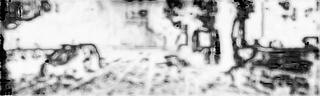} & \includegraphics[width=0.235\textwidth]{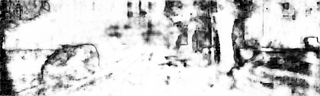} \\
        \includegraphics[width=0.235\textwidth]{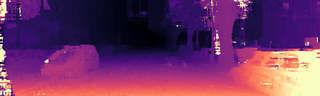} & \includegraphics[width=0.235\textwidth]{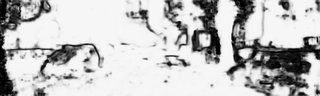} & \includegraphics[width=0.235\textwidth]{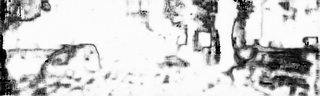} & \includegraphics[width=0.235\textwidth]{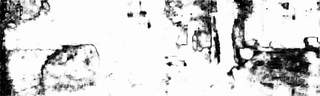} \\ \toprule

        \includegraphics[width=0.235\textwidth]{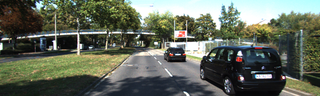} & \includegraphics[width=0.235\textwidth]{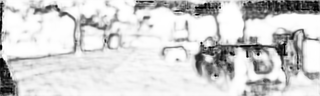} & \includegraphics[width=0.235\textwidth]{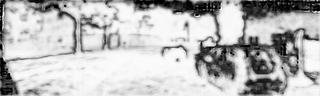} & \includegraphics[width=0.235\textwidth]{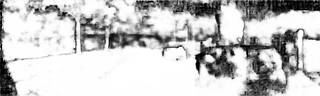} \\
        \includegraphics[width=0.235\textwidth]{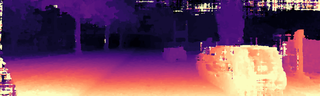} & \includegraphics[width=0.235\textwidth]{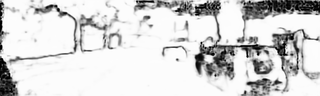} & \includegraphics[width=0.235\textwidth]{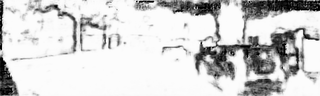} & \includegraphics[width=0.235\textwidth]{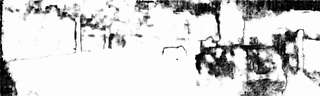} \\ \toprule

        \includegraphics[width=0.235\textwidth]{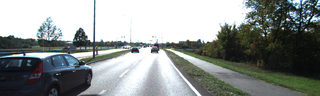} & \includegraphics[width=0.235\textwidth]{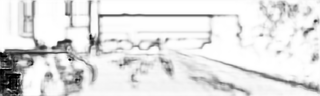} & \includegraphics[width=0.235\textwidth]{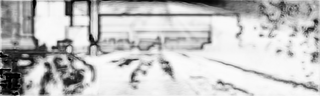} & \includegraphics[width=0.235\textwidth]{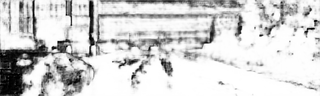} \\
        \includegraphics[width=0.235\textwidth]{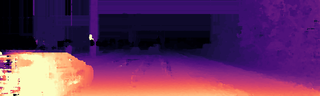} & \includegraphics[width=0.235\textwidth]{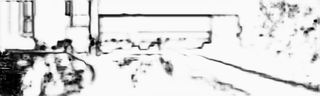} & \includegraphics[width=0.235\textwidth]{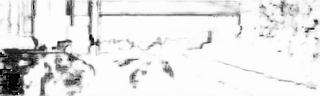} & \includegraphics[width=0.235\textwidth]{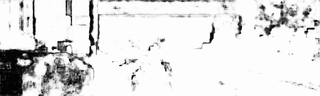} \\ \toprule

        \includegraphics[width=0.235\textwidth]{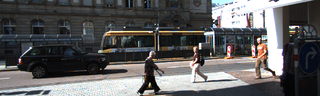} & \includegraphics[width=0.235\textwidth]{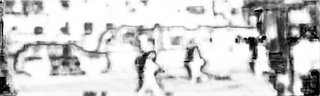} & \includegraphics[width=0.235\textwidth]{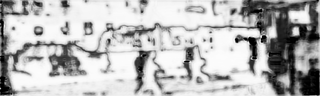} & \includegraphics[width=0.235\textwidth]{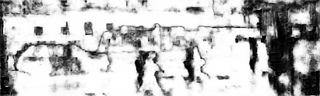} \\
        \includegraphics[width=0.235\textwidth]{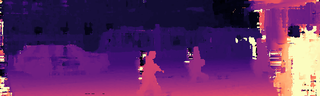} & \includegraphics[width=0.235\textwidth]{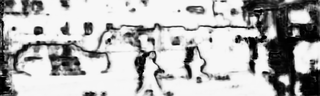} & \includegraphics[width=0.235\textwidth]{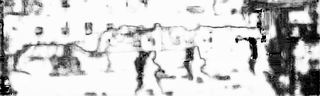} & \includegraphics[width=0.235\textwidth]{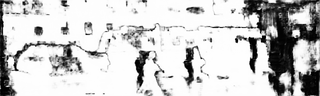} \\ \toprule

        \includegraphics[width=0.235\textwidth]{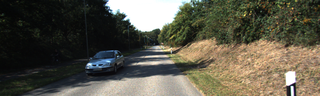} & \includegraphics[width=0.235\textwidth]{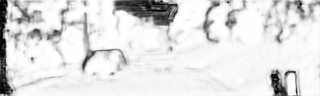} & \includegraphics[width=0.235\textwidth]{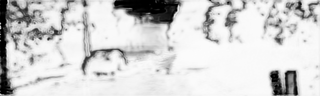} & \includegraphics[width=0.235\textwidth]{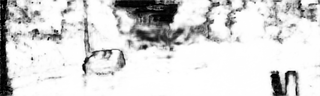} \\
        \includegraphics[width=0.235\textwidth]{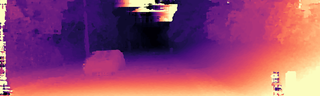} & \includegraphics[width=0.235\textwidth]{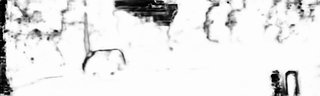} & \includegraphics[width=0.235\textwidth]{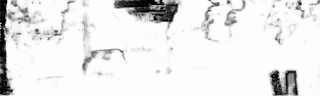} & \includegraphics[width=0.235\textwidth]{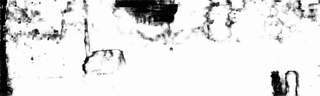} \\ \toprule

        \includegraphics[width=0.235\textwidth]{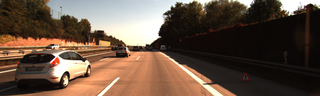} & \includegraphics[width=0.235\textwidth]{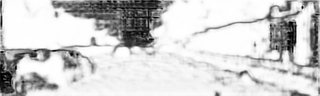} & \includegraphics[width=0.235\textwidth]{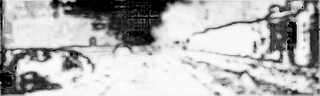} & \includegraphics[width=0.235\textwidth]{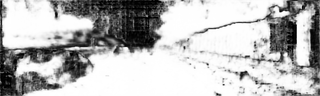} \\
        \includegraphics[width=0.235\textwidth]{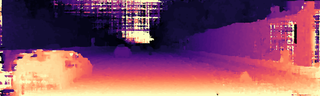} & \includegraphics[width=0.235\textwidth]{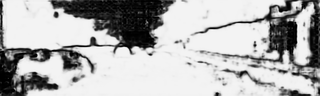} & \includegraphics[width=0.235\textwidth]{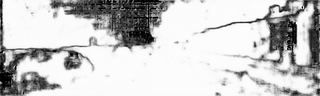} & \includegraphics[width=0.235\textwidth]{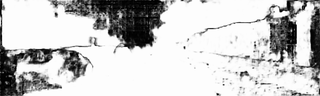} \\ \toprule

        \includegraphics[width=0.235\textwidth]{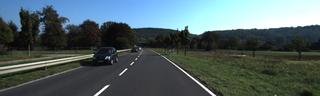} & \includegraphics[width=0.235\textwidth]{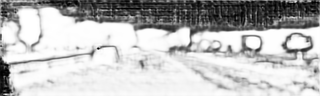} & \includegraphics[width=0.235\textwidth]{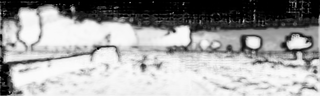} & \includegraphics[width=0.235\textwidth]{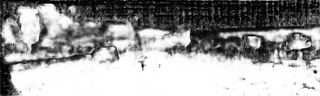} \\
        \includegraphics[width=0.235\textwidth]{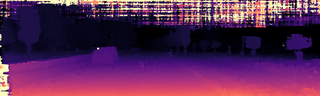} & \includegraphics[width=0.235\textwidth]{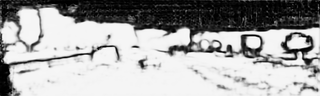} & \includegraphics[width=0.235\textwidth]{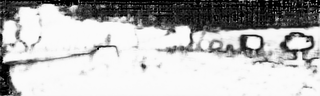} & \includegraphics[width=0.235\textwidth]{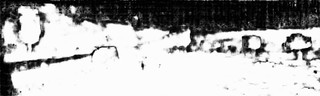} \\

        (a) & (b) & (c) & (d)
    \end{tabular}
    \caption{Qualitative evaluation for confidence estimation on KITTI 2015 dataset \cite{menze2015object}:
        Input disparity maps used for confidence estimation were obtained by Census-SGM \cite{hirschmuller2005accurate}.
        The estimated confidence maps for each input disparity map are displayed every two rows. The top and bottom of two rows indicate: (a) color image and input disparity image, (b) CCNN \cite{poggi2016learning} and CCNN w/$\tau$, (c) LAFNet* \cite{kim2019laf} and LAFNet* w/$\tau$ and (d) LAFNet and LAFNet w/$\tau$.
        `w/$\tau$' denotes the proposed network using the soft-thresholding. LAFNet* denotes the LAFNet  \cite{kim2019laf} in which 3D cost volume is not used as an input.}
    \label{fig_conf}
\end{figure*}

\section{Confidence estimation results}

\subsection{Generating ground-truth confidence map}
To train the confidence network, the ground truth confidence map is required as supervision.
Following existing confidence estimation approaches \cite{poggi2016learning, kim2019laf}, the ground truth confidence map $c^{\mathrm{gt}}$ was computed by using an absolute difference between the ground truth disparity map and the input disparity map (the pseudo ground truth disparity map in our work).

\begin{equation}
c_p^{\mathrm{gt}}=\begin{cases}
1, & \text{if $|d_p - d^{\mathrm{pgt}}_p|\le \rho$.}\\
0, & \text{otherwise}.
\end{cases}
\end{equation}
The threshold value $\rho$ is set to 3 for KITTI \cite{menze2015object} and 1 for Middlebury \cite{scharstein2014high}.

\subsection{Evaluation metric}
The area under the curve (AUC) \cite{hu2012quantitative} was used for evaluating the performance of estimated confidence maps.
The receiver operating characteristic (ROC) curve is first computed by sorting disparity pixels in a decreasing order of confidence and sequentially sampling high confidence disparity pixels. %removing low confident disparity pixels %with the ground-truth disparity %sampling subset of them
It computes the error rate indicating the percentage of pixels with a difference larger than $\rho$ from ground truth disparity. Then, AUC is computed by integral of the ROC curve.
The optimal AUC is computed according to the fact that the error rate $\zeta$ is ideally 0 when sampling the first $(1 - \zeta)$ pixels \cite{hu2012quantitative}, which is equal to
\begin{equation}
AUC_{opt}=\int_{1-\zeta}^{1}{\frac{x-(1-\zeta)}{x}dx}=\zeta+(1-\zeta)\ln{1-\zeta}.
\end{equation}

\begin{figure*}[]
    \centering
    \includegraphics[width=2\columnwidth]{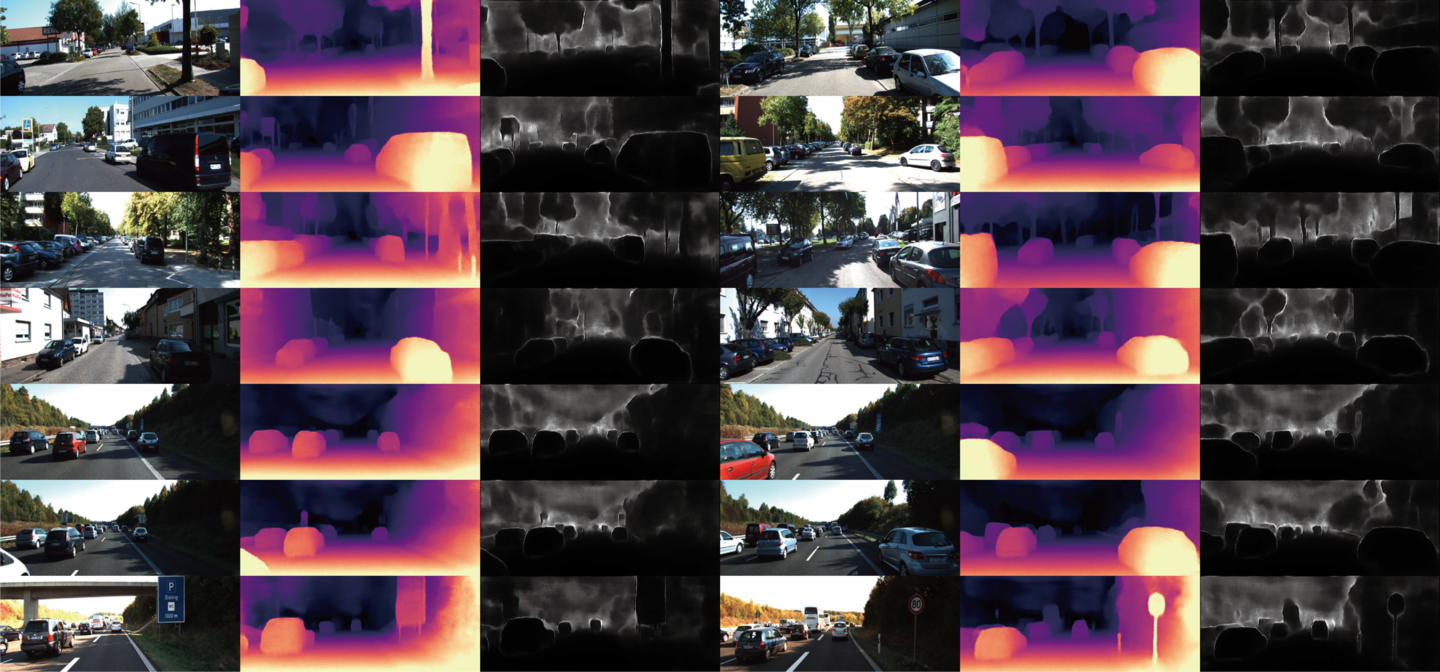}
    \caption{Qualitative result of uncertainty measure on KITTI Eigen Split \cite{eigen2014depth} test dataset. From left to right, input image, estimated depth map, and uncertainty map of the estimated depth are displayed.
    }
    \label{Qual_uncertainty}
\end{figure*}

\subsection{Qualitative evaluation on KITTI 2015 dataset}
\figref{fig_conf} shows more qualitative results of confidence map evaluated on KITTI 2015 dataset \cite{menze2015object}.
Input disparity maps used for confidence estimation were obtained by Census-SGM \cite{hirschmuller2005accurate}.
The estimated confidence maps for each input disparity map are displayed every two rows.
The top and bottom of two rows indicate: (a) color image and input disparity image, (b) CCNN \cite{poggi2016learning} and CCNN w/$\tau$, (c) LAFNet* \cite{kim2019laf} and LAFNet* w/$\tau$ and (d) LAFNet and LAFNet w/$\tau$.
`w/$\tau$' denotes the thresholded confidence map obtained using the soft-thresholding. LAFNet* denotes the LAFNet \cite{kim2019laf} in which 3D cost volume is not used as an input.
As shown in \figref{fig_conf}, the proposed thresholded confidence maps contain fewer ambiguous values than the original confidence maps.

\section{Uncertainty Estimation Details}
\subsection{Evaluation metric}
We evaluated the performance of the uncertainty estimation used in the proposed model using the sparsification error \cite{ilg2018uncertainty}. Similar to the confidence evaluation, we first sorted disparity pixels following decreasing order of uncertainty, and iteratively extracted high uncertain disparities and provided them as inputs for computing error metrics. The ideal error ranked by the true error to the ground truth is referred to as oracle. With a sparsicifation error, we computed the Area Under the Sparsification Error curve (AUSE) and the Area Under the Random Gain (AURG) to evaluate the quality of the uncertainty map. While the AUSE is measured as the difference between the sparsification and its oracle, the AURG is obtained as subtracting the estimated sparsification curve from flat curve with a random uncertainty which is modeled as a constant.
%To evaluate quality of the uncertainty of the proposed model, sparsification error \cite{ilg2018uncertainty} is used. Similar to evaluation of confidence maps, we first sort disparity pixels following decreasing order of uncertainty. Then, high uncertain disparities are iteratively extracted and provided as input for computing error metrics. The ideal error that ranked by the true error to the ground truth is refered as oracle. With sparsicifation error, we measured the Area Under the Sparsification Error curve (AUSE) and the Area Under the Random Gain (AURG) to evaluate the quality of the uncertainty map. The AUSE is measured as the difference between the sparsification and its oracle. On the other hand, the AURG is obatined as subracting estimated sparsification curve from random one.

\subsection{Qualitative evaluation on KITTI dataset}
\figref{Qual_uncertainty} shows the qualitative results of uncertainty map evaluated on KITTI Eigen Split \cite{eigen2014depth} test dataset.\
%From the left, input image, estimated depth map, and the uncertainty map of the estimated depth.
The qualitative result indicates that uncertain areas of the estimated depth map are usually located around object boundaries and sky.

\bigskip

{\small
\bibliographystyle{ieee_fullname}
\bibliography{egbib}
}